%% file: main.tex
\documentclass{article} 
\usepackage{iclr2027_conference,times}

\input{math_commands.tex}

\usepackage{amsmath,amssymb,amsthm}
\usepackage{mathrsfs}
\usepackage{booktabs,multirow}
\usepackage{array,colortbl}

\usepackage{graphicx}
\usepackage{subcaption}
\usepackage[percent]{overpic}
\usepackage{xcolor}
\usepackage{url}
\usepackage[hidelinks,hyperfootnotes=false]{hyperref}
\usepackage{cleveref}
\crefname{algorithm}{Algorithm}{Algorithms}
\Crefname{algorithm}{Algorithm}{Algorithms}
\crefname{appendix}{Appendix}{Appendices}
\Crefname{appendix}{Appendix}{Appendices}
\crefname{proposition}{Proposition}{Propositions}
\Crefname{proposition}{Proposition}{Propositions}
\crefname{assumption}{Assumption}{Assumptions}
\Crefname{assumption}{Assumption}{Assumptions}
\usepackage{algorithm}
\usepackage{algorithmic}
\usepackage{placeins}
\usepackage{microtype}

\newcommand{\ie}{\emph{i.e.}}

\renewcommand{\topfraction}{0.90}
\renewcommand{\textfraction}{0.06}
\renewcommand{\floatpagefraction}{0.75}
\newtheorem{theorem}{Theorem}
\newtheorem{proposition}{Proposition}

\newtheorem{assumption}{Assumption}

\newcommand{\cH}{\mathcal H}
\newcommand{\cB}{\mathcal B}
\newcommand{\cS}{\mathcal S}
\newcommand{\methodname}{FUSE}
\definecolor{scalecolumn}{gray}{0.95}

\title{FUSE: Fairness-aware Update Screening against Byzantine Exploitation
in $q$-Fair Federated Learning}

\newif\ifarxivversion
\arxivversiontrue

\ifarxivversion
  \iclrfinalcopy
  \input{authors.tex}
\else
  \author{Anonymous authors}
\fi

\begin{document}
\maketitle

\ifarxivversion
  \lhead{}

  \begingroup
  \renewcommand{\thefootnote}{\fnsymbol{footnote}}

  \footnotetext[1]{
    Corresponding author: xiaojinzhang@hust.edu.cn.
  }

  \footnotetext[2]{
    This work was completed while Tiantong Sun was
    an undergraduate student at
    Huazhong University of Science and Technology.
  }
  \endgroup
\fi

\begin{abstract}
Client-level fair federated learning, in the form of $q$-fair federated
learning ($q$-FFL), agnostic federated learning, or tilted empirical risk
minimization, has emerged as an attractive solution to equalize utility across
heterogeneous clients. This paradigm grants high-loss clients greater relative
influence, while a central server aggregates their loss-weighted updates into a
single shared model. Yet, the Byzantine robustness of $q$-fair federated
learning has been largely neglected, leaving the server unable to tell
legitimate fairness-induced amplification from malicious update scaling. To
address this critical gap, we formalize the problem of Byzantine-robust
$q$-fair federated learning under an \emph{update-only interface}, in which the
server observes nothing but the fairness-scaled client updates. Our analysis
under shared class-conditional label skew shows that fairness scaling preserves
the nonnegative class-mixture structure of honest updates, while replacing
their unit-sum constraint by a client-dependent radial scale. Moving beyond
magnitude-based screening, we show that this warped geometry supplies two
complementary tests, one radial and one structural, that a single norm cannot
provide. We then introduce \methodname{}, a structure-aware robust aggregator
that uses a small trusted labeled dataset for coefficient recovery, scale and
cone-consistency screening, and post-screening step-size control. Across four
benchmarks and seven Byzantine attacks, \methodname{} improves seed-wise
worst-attack accuracy over the strongest baseline on each benchmark by
3.9--30.1 percentage points, while largely preserving the influence of the
high-loss honest clients that fairness is meant to protect.
\end{abstract}

\section{Introduction}

Client-level fair federated learning aims to improve the utility of disadvantaged
clients through objectives such as $q$-fair federated learning ($q$-FFL),
agnostic federated learning, and tilted empirical risk minimization
\citep{li2020fair,mohri2019agnostic,li2021term}.
Recent multi-objective, proportional, and coalitional formulations extend this
principle to broader notions of client welfare
\citep{hu2022fedmgda,zhang2023propfair,chaudhury2022core,pan2024fedlf}.
A common implementation is to incorporate the loss-dependent weight locally:
$q$-FedAvg forms loss-weighted updates at the client, PropFair transforms the
client-side objective, and H-nobs directly transmits the fairness-aware gradient
$\nabla H_i = F_i^q\nabla F_i$
\citep{li2020fair,zhang2023propfair,zhou2023hnobs}.
This motivates an \emph{update-only} interface in which the server observes the
composite fairness-scaled update but not its loss and descent direction
separately; avoiding an additional loss report also removes a data-dependent
communication channel that can incur extra privacy cost
\citep{ling2026fedfdp,zhu2019leakage}.

Robustness under this interface, however, remains poorly understood:
$q$-FedAvg and PropFair do not model Byzantine clients, FedFDP focuses on
fairness and privacy, while H-nobs and its extension primarily screen
fairness-aware updates through their magnitudes
\citep{li2020fair,zhang2023propfair,ling2026fedfdp,zhou2023hnobs,basharat2025trustworthy}.
Such magnitude-based screening is intrinsically ambiguous in $q$-FFL, because
large updates may arise either from legitimate amplification of high-loss
clients or from adversarial scaling
\citep{zhou2023hnobs,basharat2025trustworthy}.
Meanwhile, structure-aware robust aggregation exploits class-mixture geometry,
but existing simplex-based formulations do not account for the radial distortion
introduced by fairness scaling
\citep{bao2024boba}.
We therefore study Byzantine-robust $q$-FFL under the update-only interface and
ask what geometric structure remains observable after the loss weight and
descent direction are fused into a single message
\citep{zhou2023hnobs,bao2024boba}.

\begin{figure}[!b]
\centering
\includegraphics[width=0.78\linewidth]{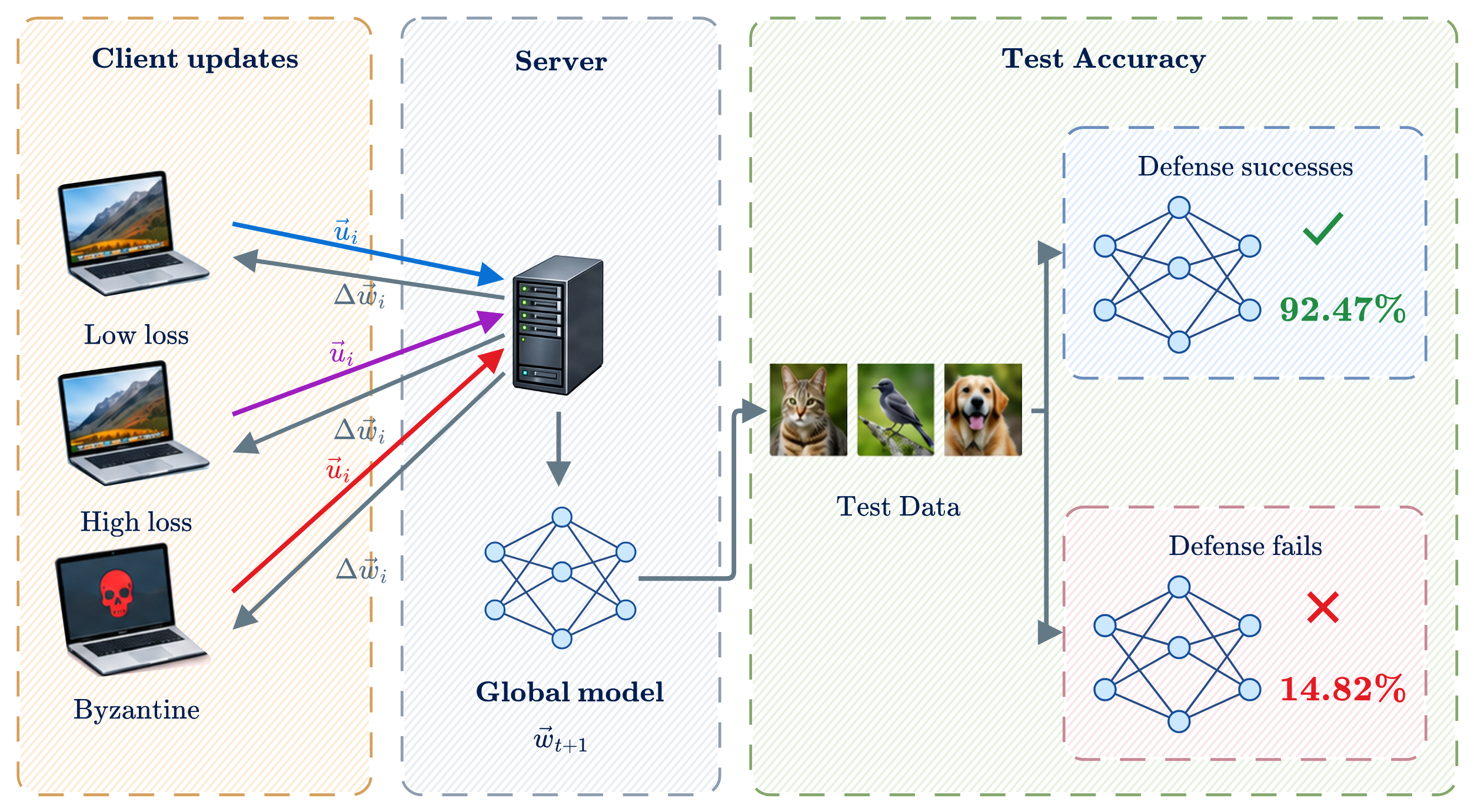}
\caption{\textbf{Update-only Byzantine-robust $q$-fair federated learning.}
Honest clients submit loss-scaled updates $u_i=F_i^q\Delta_i$, Byzantine
clients may submit arbitrary vectors, and the server aggregates them into
$w_{t+1}$ with the help of a small trusted labeled dataset. Because $q$-fair
learning legitimately enlarges the updates of high-loss honest clients, update
magnitude alone cannot distinguish fairness-induced amplification from
malicious scaling.}
\label{fig:setting}
\end{figure}

\subsection{Main Contributions}

\textbf{Problem formulation and fairness-warped geometry.} We formalize
Byzantine-robust $q$-fair federated learning under an update-only interface,
with up to $f<m/2$ of the $m$ submissions arbitrarily corrupted at each round
and a server that observes only the fairness-scaled updates. Under shared
class-conditional label skew, we show that an honest submission stays a
nonnegative combination of class-conditioned reference directions, but that its
coefficients no longer sum to one: fairness scaling replaces the unit-sum
simplex of prior structural methods~\citep{bao2024boba} by a nonnegative cone
whose coefficient mass is exactly the fairness scale $F_i^q$. A magnitude test
alone cannot separate that legitimate radial growth from adversarial scaling.

\textbf{Structure-aware robust aggregation.} Building on this geometry, we
propose \methodname{}, which recovers each submission's latent coefficients
from a small trusted labeled set by nonnegative least squares. The coefficients
supply two complementary statistics: their mass measures radial scale, and the
residual outside the trusted cone measures structural incompatibility.
\methodname{} spends its rejection budget on the two in sequence, and, since
the normalized coefficients also reconstruct the client loss, rescales the
retained mean to restore the step size that fairness weighting would inflate.

\textbf{Empirical validation.} We evaluate \methodname{} on four benchmarks
(CIFAR-10, MNIST, EMNIST-Balanced, and AGNews) covering both vision and
language modalities, against seven Byzantine attacks and seven baselines, over
five registered seeds. Our approach improves seed-wise worst-attack accuracy
over the strongest baseline on each benchmark by 3.9--30.1 percentage points,
while rejecting $0-9.7\%$ of the high-loss honest submissions that fairness is
meant to protect. Controlled geometry experiments, a full $2^3$ ablation, and
an adaptive cone-preserving attack isolate the mechanism and its boundary.

\section{Problem of Robust $q$-Fair Federated Learning}
\label{sec:setting}

We consider a synchronous federated learning system comprising $m$ clients
and a global model $w_t\in\R^d$. Each client $i$ holds a local dataset
$\mathcal D_i$ and has empirical loss
\[
F_i(w)=\frac{1}{|\mathcal D_i|}
\sum_{\xi\in\mathcal D_i}\ell(w;\xi).
\]
In this context, the goal of $q$-fair learning is to optimize one shared model
while assigning greater relative influence to high-loss clients. Specifically,
we seek a model minimizing
\[
\mathcal J_q(w)
=\frac{1}{m}\sum_{i=1}^{m}\frac{F_i(w)^{q+1}}{q+1},
\qquad q\ge0.
\]

\textbf{Update-only $q$-fair FL.}
A typical realization of this objective weights each client descent direction
by $F_i(w_t)^q$. Specifically, at round $t$, an honest client computes a
one-step update and transmits only the resulting weighted message:
\begin{equation}
\Delta_i^t=-\eta\nabla F_i(w_t),
\qquad
u_i^t=F_i(w_t)^q\Delta_i^t.
\label{eq:fair_submission}
\end{equation}
The server therefore observes the product $u_i^t$, but not its loss and update
factors separately. We refer to this setting as the \emph{update-only
interface}, and omit $t$ when unambiguous.

\textbf{Robust $q$-fair federated learning.}
We are interested in solving robust $q$-fair federated learning, wherein a
fraction of client updates may be corrupted prior to aggregation.
Specifically, Byzantine clients may coordinate and submit arbitrary vectors.
Our goal is to design an aggregation rule that remains effective despite such
corruptions while preserving the intended emphasis on high-loss clients.

\begin{assumption}[Byzantine and observation model]
\label{assump:byzantine_model}
At each round, $\cS_t=\cH_t\cup\cB_t$ and
$|\cB_t|\le f_t<|\cS_t|/2$. Byzantine clients may coordinate and submit
arbitrary vectors; the algorithm receives a declared rejection budget $f_t$.
\label{assump:screening_representation}
The server fixes a coordinate subset for screening. Screening uses only each
submission restricted to these coordinates, the trusted reference, and
quantities derived from them; it has no side information revealing client
Byzantine status. Retained clients contribute their original full-model
submissions to aggregation.
\end{assumption}

\section{From Simplex to Cone}
\label{sec:geometry}

Standard structure-aware aggregation under label skew models honest updates
as class mixtures. We show that $q$-fair weighting preserves this mixture
structure but removes its unit-sum constraint, changing the appropriate
observation model from a simplex to a cone.

\subsection{Fairness Scaling Breaks the Simplex}

\textbf{Magnitude ambiguity.}
For $F_i>0$, $q>0$, and any $\lambda>0$,
\[
(\lambda F_i)^q(\lambda^{-q}\Delta_i)=F_i^q\Delta_i.
\]
The update-only message does not separately identify fairness scale and
descent magnitude. Magnitude alone therefore cannot distinguish legitimate
$q$-fair amplification from adversarial scaling.

\begin{assumption}[Shared class-conditional label skew]
\label{assump:label_skew}
For every honest client $i$,
\[
P_i=\sum_{c=1}^{C}p_{ic}Q_c,
\qquad p_i\succeq0,
\qquad \mathbf 1^\top p_i=1,
\]
where $Q_c$ is the class-$c$ conditional distribution shared by honest
clients. At the broadcast model, the population class losses satisfy
$0<\ell_{\min}\le\ell_c\le\ell_{\max}<\infty$.
\end{assumption}

Here $p_i$ contains client $i$'s label proportions. The guarantees do not
cover arbitrary violations of shared class-conditionals; covariate shifts are
evaluated empirically in \Cref{app:domain-shift}.

Let $\ell=(\ell_1,\ldots,\ell_C)^\top$ and
$G^\star=[g_1^\star,\ldots,g_C^\star]$ contain the population class losses
and class-conditioned updates, respectively. Under
\Cref{assump:label_skew}, linearity of expectation gives
\begin{equation}
\begin{aligned}
\overline F_i&=\ell^\top p_i,\qquad
\overline\Delta_i=G^\star p_i,\\
\overline u_i&=(\ell^\top p_i)^qG^\star p_i=G^\star a_i^\star,\qquad
a_i^\star:=(\ell^\top p_i)^q p_i\succeq0,\qquad
\|a_i^\star\|_1=(\ell^\top p_i)^q.
\end{aligned}
\label{eq:fair_cone_identity}
\end{equation}

\noindent\textbf{Key consequence.}
Without fairness scaling, class-mixture coefficients satisfy
$\mathbf 1^\top p_i=1$ and lie on a simplex. Fairness scaling preserves
coefficient nonnegativity but changes their mass to $F_i^q$. Thus the honest
observation model is no longer a unit-sum simplex but a radially varying
nonnegative cone.

\section{\methodname{} Robust Aggregator}
\label{sec:method}

\begin{figure}[h!]
\centering
\includegraphics[width=0.86\linewidth]{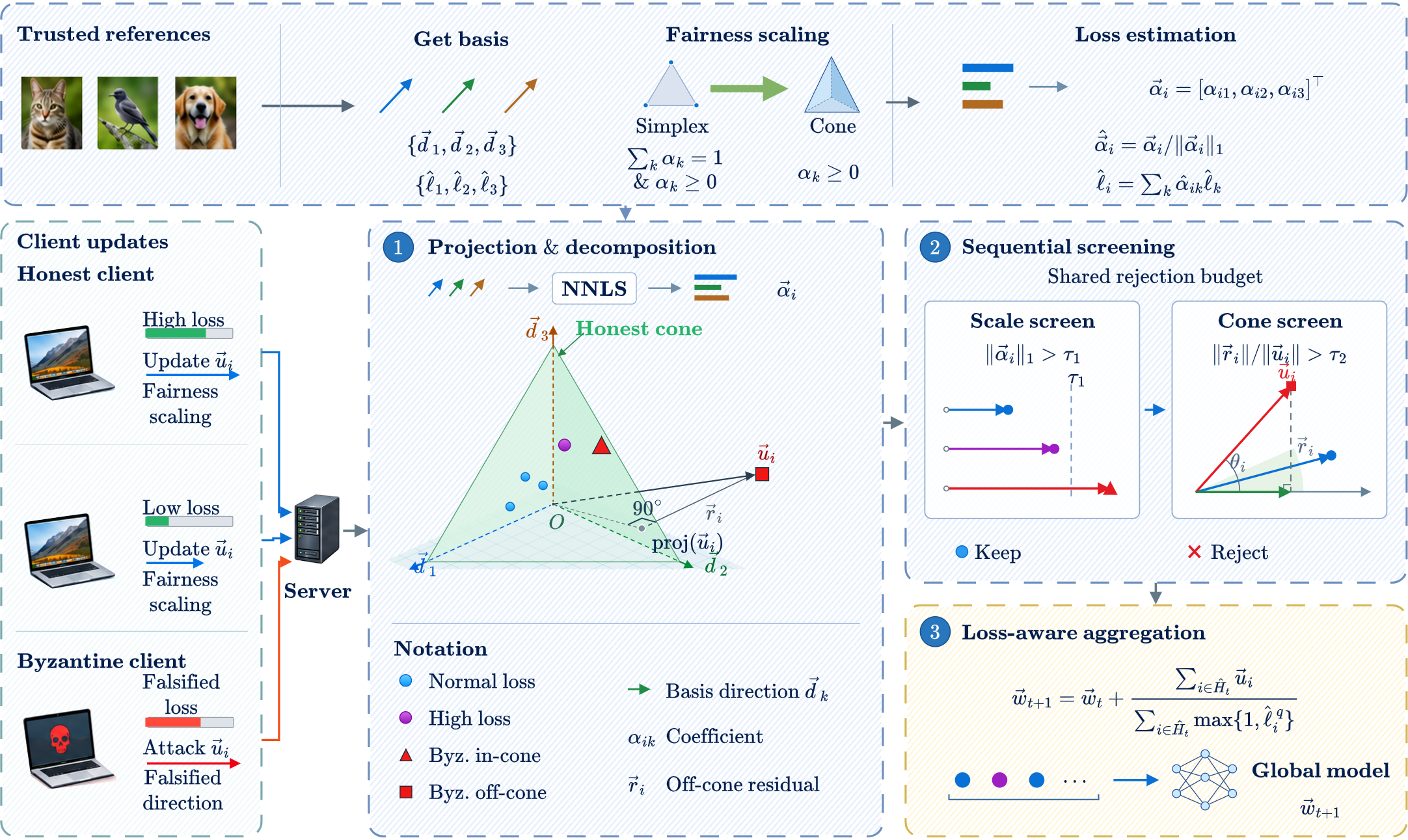}
\caption{\textbf{\methodname{} overview.}
Each fair update is decomposed over trusted class-conditioned directions by
nonnegative least squares. The recovered mass $L_i=\|\widehat a_i\|_1$ tests
for anomalous radial scaling; the normalized cone residual $D_i^{\rm cone}$
detects updates inconsistent with any nonnegative class mixture.
\methodname{} removes scale outliers first, spends the remaining budget on the
largest cone residuals, and reconstructs the client loss from the normalized
coefficients for post-screening step-size control.}
\label{fig:algorithm}
\end{figure}

In this section, we construct a robust aggregator from the cone structure
identified above. To alleviate the ambiguity between legitimate scale and
malicious deviation, \methodname{} combines trusted class references,
coefficient recovery, two-stage screening, and post-screening step-size
control, as illustrated in \Cref{fig:algorithm}.

\textbf{Trusted class reference.}
The server uses a small labeled set to construct class-conditioned reference
directions $G=[g_1,\ldots,g_C]$ and trusted class losses $\widehat\ell$ in the
screening coordinates:
\begin{equation}
\begin{aligned}
\widehat\ell_c
&=
\frac{1}{|\mathcal D_s^c|}
\sum_{\xi\in\mathcal D_s^c}\ell(w_t;\xi),
\\
g_c&=
\left[
-\eta\nabla_w
\frac{1}{|\mathcal D_s^c|}
\sum_{\xi\in\mathcal D_s^c}\ell(w_t;\xi)
\right]_{\rm scr},
\\
G&=[g_1,\ldots,g_C],\qquad
\widehat\ell=(\widehat\ell_1,\ldots,\widehat\ell_C)^\top.
\end{aligned}
\label{eq:trusted_reference}
\end{equation}
Define the trusted cone $\mathcal C_G:=\{Ga:a\succeq0\}$.

\textbf{Coefficient recovery and screening scores.}
Since honest submissions are approximately nonnegative combinations of the
trusted directions, \methodname{} recovers their latent coefficients by
nonnegative least squares. Specifically, for each submission it solves
\begin{equation}
\widehat a_i\in\arg\min_{a\succeq0}\|v_i-Ga\|_2^2,
\qquad
L_i=\|\widehat a_i\|_1,
\qquad
D_i^{\rm cone}=
\frac{\|v_i-G\widehat a_i\|_2^2}
{\max\{\|v_i\|_2^2,\epsilon\}}.
\label{eq:recovery_scores}
\end{equation}
$L_i$ measures radial scale, while $D_i^{\rm cone}$ measures structural
incompatibility with the trusted nonnegative cone. Stage I removes up to $f$
submissions with implausibly large coefficient mass
$L_i>\kappa L_{\rm med}$; Stage II spends the remaining rejection budget on
the largest cone residuals, yielding the retained set $\widehat\cH_t$.

\textbf{Trusted loss reconstruction and aggregation.}
Finally, normalized coefficients estimate class composition and hence the
client loss:
\begin{equation}
\widehat p_i=\frac{\widehat a_i}{\max\{L_i,\epsilon\}},
\qquad
\widetilde F_i=\widehat\ell^\top\widehat p_i,
\qquad
w_{t+1}=w_t+
\frac{\sum_{i\in\widehat\cH_t}u_i}
{\sum_{i\in\widehat\cH_t}\max\{1,\widetilde F_i^q\}}.
\label{eq:aggregation}
\end{equation}
In the ideal model, $\widehat p_i=p_i$ and $\widetilde F_i=\ell^\top p_i$.
The denominator only rescales the retained mean and therefore does not change
its direction. Additional implementation details are provided in
\Cref{app:scale-implementation}.

\begin{algorithm}[!t]
\caption{\methodname{} at round $t$}
\label{alg:scboba}
\begin{algorithmic}[1]
\REQUIRE submissions $\{u_i\}_{i\in\cS_t}$, trusted class data
$\{\mathcal D_s^c\}_{c=1}^{C}$, budget $f$, $q$, $\kappa$, $\epsilon$
\STATE Construct $(G,\widehat\ell)$ using Eq.~\eqref{eq:trusted_reference}
\FOR{each $i\in\cS_t$}
    \STATE Recover $\widehat a_i$ by NNLS in Eq.~\eqref{eq:recovery_scores}
    \STATE Compute $L_i$ and $D_i^{\rm cone}$ using Eq.~\eqref{eq:recovery_scores}
\ENDFOR
\STATE Reject scale anomalies using $L_i>\kappa L_{\rm med}$
\STATE Use the remaining budget to reject the largest cone scores
\STATE For each retained client, compute $\widehat p_i$ and $\widetilde F_i$
using Eq.~\eqref{eq:aggregation}
\STATE Update $w_{t+1}$ using Eq.~\eqref{eq:aggregation}
\ENSURE $w_{t+1}$
\end{algorithmic}
\end{algorithm}

\section{Experiments}
\label{sec:experiments}

This section summarizes the key results of our experiments. In all that
follows, $q=1$ and the declared rejection budget is $f=4$. We evaluate
\methodname{} against fair, robust, and structure-aware baselines in
\Cref{sec:main-results}, and audit whether its robustness preserves the
high-loss clients that $q$-FFL emphasizes in \Cref{sec:high-loss}. We isolate
the geometric mechanism and conduct a component ablation in
\Cref{sec:mechanism}; further results are reported in the appendix.

\subsection{Experimental Setup}

\textbf{Datasets and models.}
We evaluate CIFAR-10/ResNet-18~\citep{krizhevsky2009cifar,he2016resnet},
MNIST/MLP~\citep{lecun1998mnist}, EMNIST-Balanced/CNN~\citep{cohen2017emnist},
and AGNews/GRU~\citep{zhang2015agnews,cho2014gru}. Data are partitioned by
Dirichlet label skew with $\alpha=1$, and
each round samples 10 of 100 honest clients. \methodname{} uses 20 trusted
labeled examples per class. Screening uses the classifier block on the three
vision benchmarks and all model parameters on AGNews; retained clients always
contribute their original full-model submissions.

\textbf{Threat model and baselines.}
We consider Sign Flip, IPM, Gaussian, ALIE, Min-Max, Min-Sum, and Magnitude
Scaling~\citep{xie2020ipm,baruch2019alie,shejwalkar2021manipulating}. Four
Byzantine Sybils are appended in attacked rounds. We compare with
$q$-FFL~\citep{li2020fair}, H-nobs~\citep{zhou2023hnobs}, Trimmed
Mean~\citep{yin2018robust}, Krum~\citep{blanchard2017krum},
FLTrust~\citep{cao2021fltrust}, BOBA~\citep{bao2024boba}, and
FedAA~\citep{he2025fedaa} under the same fairness-scaled message whenever
applicable.

\textbf{Evaluation.}
We conduct each experiment with five registered seeds (40--44), and report
global and worst-10\% client accuracy (W10) averaged over the final ten rounds.
Our primary statistic is the seed-wise minimum across the seven attacks, \ie
the accuracy an adversary can force by picking the best attack for that seed.
Runs triggering the registered collapse criterion are not replaced by
pre-collapse maxima, and the number of completed seeds is reported as $^{(k)}$.
We include additional details in \Cref{app:setup}.

\subsection{Robustness under Byzantine Attacks}
\label{sec:main-results}

\FloatBarrier
\input{formal_results_5seed.tex}

In \Cref{tab:main-results}, we observe that \methodname{} consistently attains
the highest seed-wise worst-attack global and W10 accuracy on all four
benchmarks, improving worst-attack global accuracy over the best baseline by
30.1, 3.9, 17.1, and 13.4 percentage points on CIFAR-10, MNIST,
EMNIST-Balanced, and AGNews respectively. Beyond worst-case performance, it
attains the highest accuracy in 13 of the 28 dataset--attack pairs; across the
remaining 15 its largest deficit to the best method is 3.5 percentage points,
under Sign Flip on AGNews. Relative to FLTrust, the strongest baseline overall,
the paired global difference is positive on all five seeds of each vision
benchmark, with bootstrap 95\% intervals $[20.2,37.8]$, $[1.6,6.3]$ and
$[12.3,21.9]$. The advantage is particularly pronounced against the
invalid-mixture attacks: on CIFAR-10 the baselines drop to 13--62\% under
Min-Max and Min-Sum while \methodname{} stays above 82\%. The worst-attack
column makes the general pattern visible: every baseline has at least one
attack for which its accuracy floor falls well below its own clean
performance. This robustness does not uniformly improve clean utility:
\methodname{} trails the best clean baseline by 2.6, 1.2, and 1.6 percentage
points on MNIST, EMNIST-Balanced, and AGNews, and leads it by 0.3 on
CIFAR-10. We regard this as the price of a substantially higher adversarial
floor.

\subsection{Preserving High-Loss Honest Clients}
\label{sec:high-loss}

A robust aggregator should not obtain its robustness by discarding the
high-loss honest clients that $q$-FFL is designed to emphasize, nor buy a clean
fairness record by screening nothing at all. We therefore audit both sides of
the screening decision under matched seeds, attacks, rejection budget $f=4$,
evaluation grid and round window. We focus on the two closest routes: H-nobs,
which filters fair updates by norm, and BOBA, which fits a unit-sum
class-mixture model on classifier coordinates.

\input{high_loss_matched_main.tex}

In \Cref{tab:high-loss-main}, we observe that the two baselines fail on
opposite sides. Norm screening pays a steep fairness price: H-nobs rejects
$22.2-35.7\%$ of the highest-loss honest submissions, and up to $91.7\%$ of
them under its worst attack, in exchange for a $61.1-68.6\%$ Byzantine
rejection rate. BOBA barely touches the fair tail ($0.0-5.6\%$), but only
because it rejects almost nothing: it removes $14.3-36.8\%$ of the Byzantine
submissions, and nothing at all under Sign Flip and ALIE. \methodname{} is the
only method that is conservative on both counts, rejecting $0.0-9.7\%$ of the
high-loss honest submissions while removing $85.7-99.8\%$ of the Byzantine
ones. Its robustness therefore does not arise from suppressing the $q$-fair
tail. The one exception is ALIE on AGNews, which \methodname{} does not
separate within this window and which produces the $0.00$ worst-case entry; it
is also the mildest attack in the suite, costing 3.7 percentage points of
accuracy in \Cref{tab:main-results}. The complete per-attack audit is in
\Cref{app:diagnostics}.

\FloatBarrier

\section{Why Does Cone Geometry Matter?}
\label{sec:mechanism}

The results above establish robustness, but do not show that the gain comes
from the fairness-warped cone geometry rather than from some other ingredient
of the aggregator. We therefore ask whether $q$-fair scaling breaks the simplex
model in practice, whether the mismatch changes behavior in an otherwise
identical pipeline, which components are responsible, and how the method
behaves as its operating parameters vary.

\subsection{Fairness Scaling Breaks the Simplex in Practice}

\begin{figure}[!t]
\centering
\begin{subfigure}[t]{0.375\linewidth}
\centering
\includegraphics[width=\linewidth]{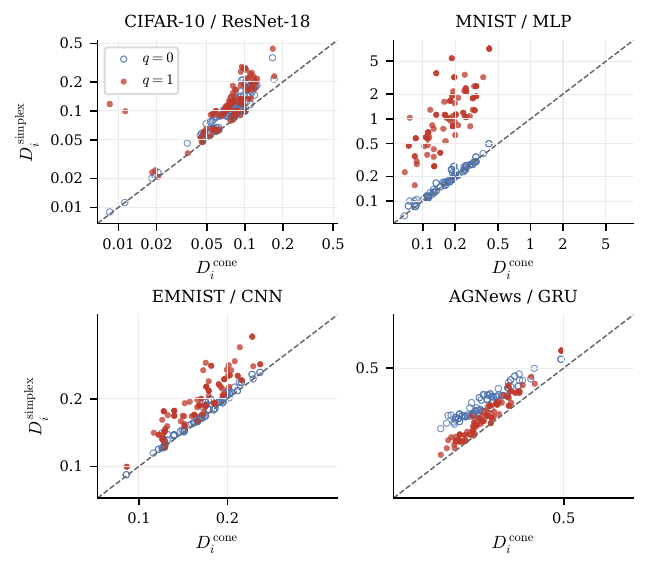}
\caption{Controlled simplex--cone mismatch.}
\end{subfigure}\hspace{0.03\linewidth}%
\begin{subfigure}[t]{0.375\linewidth}
\centering
\includegraphics[width=\linewidth]{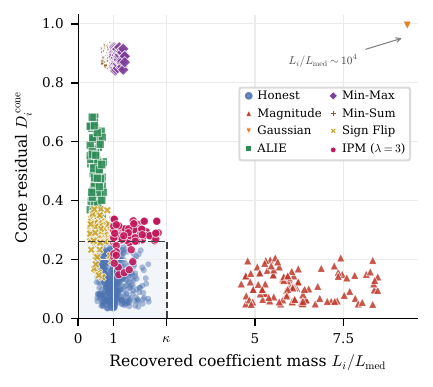}
\caption{Scale and cone statistics in training.}
\end{subfigure}
\caption{\textbf{Fairness-warped geometry in practice.}
(a) Fairness scaling moves the simplex residual off the diagonal while leaving
the cone residual nearly unchanged; hollow and filled points are $q=0$ and
$q=1$. (b) On MNIST (seed 42, $q=1$), radial and structural attacks occupy
different regions of the $(L_i,D_i^{\rm cone})$ plane, which is what motivates
screening the two in sequence rather than with one test
(\Cref{app:simplex-cone,app:stagewise-screening}).}
\label{fig:pipeline}
\end{figure}

Holding the checkpoint, honest submissions, and trusted basis fixed and
varying only the coefficient constraint, the median cone/simplex residual on
MNIST is $0.178/0.185$ at $q=0$ but $0.178/1.077$ after $q=1$ scaling: radial
scaling preserves the cone, whereas the unit-sum simplex becomes mismatched
(\Cref{fig:pipeline}(a)). The recovered coefficient mass tracks the fairness
scale (Pearson $r=0.940$ between $L_i$ and $F_i^q$), and normalized
coefficients reconstruct client losses with $r=0.923$ and MAE $0.053$.

\subsection{Matched Training Consequence of the Simplex Mismatch}

To isolate the geometry, we replace only \methodname{}'s Stage-II cone score by
a unit-sum simplex score on MNIST. Both scores saturate on Gaussian and
Magnitude Scaling, so the geometry is visible only where the structural test
does the work. There the cone dominates: under IPM, Min-Max, and Min-Sum it
lifts Byzantine recall from $0.45/0.56/0.62$ to $0.69/1.00/1.00$ and cuts the
honest false-positive rate from $0.219/0.175/0.153$ to $0.124/0.001/0.000$,
worth 2.44, 1.03, and 1.13 percentage points. Sign Flip is an informative
counterexample, where the simplex score is the better detector ($0.92$ against
$0.58$). The cone is not universally superior; it is the representation matched
to fairness-scaled honest mixtures (\Cref{app:simplex-cone}).

\subsection{Which Components Matter?}

\input{component_ablation_cifar_main.tex}

The full $2^3$ ablation assigns each component the role our geometry predicts.
Removing the scale screen exposes \methodname{} to Magnitude Scaling (worst
attack 77.50 to 52.38) while leaving the other attacks intact, confirming it is
the radial test. Removing the cone screen instead eliminates the IPM terminal
window and breaks the invalid-mixture attacks (ALIE 78.50 to 55.16, Min-Max
79.28 to 36.71, Min-Sum 78.71 to 38.94); removing both also breaks Gaussian
(26.11). The screens are thus complementary, while loss normalization is an
optimization component, raising the floor from 70.06 to 77.50
(\Cref{app:factorial-ablation}).

\subsection{Sensitivity to $q$, $\kappa$, and the Byzantine Budget}

\input{sensitivity_budget_main.tex}

Worst-attack global/W10 accuracy varies smoothly from 91.57/86.21 to
90.47/84.85 across the tested $q$, so the method does not rely on a particular
fairness strength. Increasing $\kappa$ lowers the honest false-positive rate
and raises the retained fair mass without materially changing accuracy, and
over-declaring the budget costs fairness rather than accuracy
(\Cref{app:diagnostics}). Neither statistic can certify maliciousness for a
submission inside the trusted cone with a plausible recovered mass: the
adaptive attack of \Cref{app:adaptive-attacks} reaches 65.6\% on CIFAR-10
against 82.4\% under clean training, an observable boundary of
\methodname{}'s two screening statistics rather than a missing threshold.

\FloatBarrier

\section{Theoretical Analysis}
\label{sec:theory}

We now characterize when the fairness-warped geometry exploited by
\methodname{} is reliably observable, and when it suffices to exactly recover
the honest set, thereby matching the post-screening aggregate obtained with
oracle knowledge of the honest identities. The analysis proceeds in two steps:
stable recovery of the latent fair coefficients, then conditions under which
the two screening stages recover the honest set exactly.

\subsection{Recoverability of the Fairness-Warped Geometry}

The cone identity in \Cref{sec:geometry} is a population-level
description. In practice, both client updates and the trusted class basis
are estimated from finite data. For an honest client, write its
screening-coordinate submission as
\begin{equation}
    v_i = G a_i + e_i,
    \qquad
    a_i := F_i^q p_i \succeq 0,
    \qquad
    \|a_i\|_1 = F_i^q,
    \label{eq:finite_cone_model}
\end{equation}
where $e_i$ collects the finite-sample client and trusted-basis
approximation errors. We are now ready to state the corresponding
recovery guarantee.

\begin{theorem}[Stable recovery of the fairness-warped representation]
\label{thm:fair_warped_geometry}
If $G$ has full column rank, the NNLS estimator in
\eqref{eq:recovery_scores} satisfies
$
    \|\widehat a_i-a_i\|_2
    \le
    \frac{2\|e_i\|_2}{\sigma_{\min}(G)},
    \qquad
    \left|L_i-F_i^q\right|
    \le
    \frac{2\sqrt{C}\|e_i\|_2}{\sigma_{\min}(G)}.
$
\end{theorem}

Recovery is therefore controlled by the finite-sample representation error and
the conditioning of the trusted basis: when honest submissions stay close to
the trusted cone, $L_i=\|\widehat a_i\|_1$ stably recovers the fairness scale
$F_i^q$. The same argument yields stable composition and trusted-loss
recovery; the full bounds and proof are in \Cref{appendix:scboba_theory},
and the prediction matches the diagnostics in \Cref{fig:pipeline}.

\subsection{Robustness of Two-Stage Screening}

Stable recovery alone does not guarantee robust screening: a Byzantine
submission must also be separated from the natural variation of honest
submissions. Let
$
    m_H:=\operatorname{med}_{h\in\cH_t}\log L_h,
    \qquad
    r_H :=
    \max_{h\in\cH_t}
    \left|
    \log L_h - m_H
    \right|,
    \qquad
    \epsilon_H^{\rm cone}
    :=
    \max_{h\in\cH_t}D_h^{\rm cone},
$
and, for a vector $c$, let $(c)_-:=\max\{-c,0\}$ denote its coordinate-wise
negative part. Stage~I measures radial separation, whereas Stage~II measures
structural separation from the trusted nonnegative cone.

\begin{theorem}[Oracle recovery under geometric separation]
\label{thm:scale_cone_separation}
Assume that $G$ has full column rank, honest submissions form a strict
majority, $f_t=|\cB_t|$, and $\log\kappa>2r_H$.
Any Byzantine submission satisfying
$
    \log L_b-m_H>\log\kappa+r_H
$
is rejected by Stage~I; in particular, for a positive radial copy
$v_b=\lambda v_h$ this holds whenever
$
    \log\lambda>\log\kappa+2r_H.
$
For any Byzantine submission surviving Stage~I, write
$v_b=Gc_b+r_b$ with $r_b\perp\operatorname{span}(G)$. If
$
    \frac{
        \|r_b\|_2^2
        +
        \sigma_{\min}^2(G)\|(c_b)_-\|_2^2
    }{\|v_b\|_2^2}
    >
    \epsilon_H^{\rm cone},
    \label{eq:structural_separation}
$
then $v_b$ ranks above every honest submission in Stage~II.
Consequently, if every Byzantine submission satisfies one of these two
separation conditions, then
$
    \widehat\cH_t=\cH_t,
$
and Eq.~\eqref{eq:aggregation} returns exactly the aggregate that would be
obtained by running the same post-screening rule with oracle knowledge of the
honest identities.
\end{theorem}

The theorem makes the roles of the two statistics explicit. Robustness
is controlled by the honest radial spread $r_H$, the honest cone
variation $\epsilon_H^{\rm cone}$, and the radial or structural
deviation of the Byzantine submission. Stage~I detects excessive radial
scaling, while Stage~II detects off-span energy or negative mixture mass
beyond honest finite-sample variation. Their complementary behavior is
observed empirically in \Cref{fig:pipeline} and
\Cref{app:stagewise-screening}. If neither separation condition holds,
the theorem provides no certificate of maliciousness; the adaptive
cone-preserving attack in \Cref{app:adaptive-attacks} probes this
observable boundary.

\section{Conclusion}

We formalized Byzantine-robust $q$-fair federated learning under an
update-only interface, and showed that fairness scaling relaxes the unit-sum
simplex of honest class mixtures into a nonnegative cone with client-dependent
radial mass. Building on this geometry, we proposed \methodname{}, which
screens the recovered coefficient mass and the cone residual in sequence and
rescales the retained mean by a reconstructed loss. Our experiments show that
\methodname{} consistently improves worst-attack accuracy while preserving the
high-loss honest clients, and our adaptive attack delimits the interface: only
deviations the observable statistics separate can be resisted.

\subsection*{Reproducibility Statement}

All details needed to reproduce our experiments are given in
\Cref{sec:experiments} and \Cref{app:setup}: registered seeds,
data partitions, hyperparameters, attacks, and the reporting protocol.
Component, boundary, and geometry diagnostics are in
\Cref{app:diagnostics}.

\section*{AI Use Statement}

Generative AI assisted execution, debugging, and language polishing. The
authors developed the research and verified all assisted content and results.

\bibliography{iclr2027_conference}
\bibliographystyle{iclr2027_conference}

\input{appendix/appendix.tex}

\end{document}

%% file: math_commands.tex
\usepackage{amsmath,amsfonts,bm}

\def\eqref#1{(\ref{#1})}

\def\1{\bm{1}}

\DeclareMathAlphabet{\mathsfit}{\encodingdefault}{\sfdefault}{m}{sl}
\SetMathAlphabet{\mathsfit}{bold}{\encodingdefault}{\sfdefault}{bx}{n}

\newcommand{\R}{\mathbb{R}}



%% file: authors.tex
\author{%
\parbox{0.98\textwidth}{%
\centering
\textbf{Tiantong Sun$^{1,\dagger}$, Jinlin Liu$^{2}$, Xingru Wei$^{3}$, Zequan Fang$^{4}$,}\\
\textbf{Wei Chen$^{3}$, Qiang Yang$^{5}$, Xiaojin Zhang$^{3,*}$}\\[0.2em]
{\normalfont\footnotesize
$^1$George Mason University \quad
$^2$The University of Hong Kong\\
$^3$Huazhong University of Science and Technology\\
$^4$The Chinese University of Hong Kong, Shenzhen \quad
$^5$The Hong Kong Polytechnic University\\
\texttt{tsun20@gmu.edu},
\texttt{jinlinliu@connect.hku.hk},
\texttt{u202315638@hust.edu.cn}\\
\texttt{zequanfang@link.cuhk.edu.cn},
\texttt{lemuria\_chen@hust.edu.cn}\\
\texttt{profqiang.yang@polyu.edu.hk},
\texttt{xiaojinzhang@hust.edu.cn}
}%
}%
}

%% file: formal_results_5seed.tex
\begin{table}[t]
\centering
\scriptsize
\setlength{\tabcolsep}{2.6pt}
\renewcommand{\arraystretch}{0.82}
\setlength{\aboverulesep}{0.30ex}
\setlength{\belowrulesep}{0.30ex}
\caption{\textbf{Robustness across datasets and attacks.} Mean global accuracy (\%) over the seeds reaching the registered endpoint; $^{(k)}$ denotes $k$ completed seeds. The last column is the seed-wise worst attack over the seven attacks, reported as global (G) / worst-10\% client (W10) accuracy. Per-attack W10 results and across-seed standard deviations are in Appendix~\ref{app:full-main-results}.}
\label{tab:main-results}
\makebox[\textwidth][c]{%
\begin{tabular}{@{}llrrrrrrrrc@{}}
\toprule
Dataset/model & Method & Clean & Sign & IPM & Gauss. & ALIE & Min-Max & Min-Sum & Mag. & Worst attack (G / W10)\\
\midrule
\multirow{8}{*}{\shortstack[l]{\textbf{CIFAR-10}\\ResNet-18}} & q-FFL & 76.96 & 76.47 & -- & 13.81$^{(3)}$ & 63.43 & 14.76$^{(4)}$ & 14.05 & 43.45$^{(4)}$ & --\\
 & H-nobs & 76.96 & 12.19$^{(3)}$ & 57.13 & 79.28 & 35.65 & 42.86 & 20.15 & 76.96 & 12.19 / 0.35$^{(3)}$\\
 & Trim. & 70.56 & 59.66 & 44.84 & 79.45 & 39.27 & 13.37 & 14.06 & 74.35 & 13.18 / 1.28\\
 & Krum & 63.66 & 44.69 & 62.78 & 60.21 & 22.24 & \underline{61.58} & \underline{45.62} & 63.56 & 21.67 / 9.78\\
 & FLTrust & 81.31 & \textbf{81.29} & \textbf{81.31} & 81.25 & 55.13 & 48.84 & 43.81 & \underline{79.44} & \underline{43.81 / 30.53}\\
 & BOBA (cls.) & \underline{81.93} & 72.68 & 10.93 & \underline{81.58} & \underline{72.00} & 17.31 & 17.83 & 73.64 & 10.93 / 0.66\\
 & FedAA & 75.07 & 45.32 & 74.07$^{(4)}$ & 72.60 & 61.44 & 22.54 & 43.07 & 66.07 & 22.15 / 10.27$^{(4)}$\\
\rowcolor{scalecolumn}  & \textbf{\methodname{}} & \textbf{82.25} & \underline{79.29} & \underline{80.59} & \textbf{82.30} & \textbf{77.74} & \textbf{82.17} & \textbf{82.19} & \textbf{81.12} & \textbf{73.88 / 63.58}\\
\midrule
\multirow{8}{*}{\shortstack[l]{\textbf{MNIST}\\MLP}} & q-FFL & 94.90 & 93.24 & -- & 62.99$^{(3)}$ & 94.29 & 48.54$^{(3)}$ & 56.03$^{(3)}$ & 68.76$^{(3)}$ & --\\
 & H-nobs & 94.90 & 14.82$^{(3)}$ & 91.09 & 94.90 & 93.45 & \textbf{94.80} & 65.54$^{(3)}$ & 94.90 & 14.82 / 1.91$^{(3)}$\\
 & Trim. & 94.90 & 85.77 & 87.30$^{(4)}$ & 94.84 & 65.65$^{(3)}$ & 48.61$^{(3)}$ & 48.74$^{(3)}$ & 95.33 & 48.24 / 37.25$^{(3)}$\\
 & Krum & 93.99 & 90.45 & \textbf{93.42} & 93.99 & 63.99$^{(3)}$ & \underline{94.00} & \underline{93.76} & 93.99 & 63.99 / 56.14$^{(3)}$\\
 & FLTrust & 94.96 & \underline{93.30} & \underline{93.34} & 94.98 & 93.10 & 89.97 & 91.77 & 95.07 & \underline{88.52 / 79.64}\\
 & BOBA (cls.) & \underline{96.41} & \textbf{94.45} & 11.23 & \textbf{96.41} & \underline{95.70} & 80.55 & 82.26 & \textbf{96.60} & 11.23 / 0.16\\
 & FedAA & \textbf{96.42} & 51.53 & 85.86 & \underline{96.29} & \textbf{96.30} & 74.04 & 60.57 & \underline{96.22} & 43.00 / 31.26\\
\rowcolor{scalecolumn}  & \textbf{\methodname{}} & 93.83 & 93.15 & 92.42 & 93.83 & 93.81 & 93.84 & \textbf{93.84} & 93.83 & \textbf{92.42 / 87.07}\\
\midrule
\multirow{8}{*}{\shortstack[l]{\textbf{EMNIST-Bal.}\\CNN}} & q-FFL & 63.79 & \underline{74.93} & -- & 2.13$^{(1)}$ & 53.66$^{(3)}$ & 44.08$^{(3)}$ & 27.01$^{(3)}$ & 65.81 & --\\
 & H-nobs & 63.79 & 28.70$^{(2)}$ & \underline{78.66} & 63.79 & 78.55$^{(3)}$ & 52.75$^{(3)}$ & 50.65$^{(3)}$ & 63.79 & 28.70 / 21.64$^{(2)}$\\
 & Trim. & \underline{80.30} & 63.17 & 54.93$^{(4)}$ & 63.28 & 66.82$^{(4)}$ & 25.45$^{(3)}$ & 34.35$^{(3)}$ & 64.13 & 37.11 / 26.83$^{(2)}$\\
 & Krum & 78.49 & 7.35$^{(2)}$ & 78.64 & 78.49 & 71.34$^{(3)}$ & 47.92$^{(3)}$ & 4.12$^{(3)}$ & 78.49 & 5.12 / 0.97$^{(2)}$\\
 & FLTrust & 71.34 & 71.02 & 71.46 & 72.17 & 75.48 & 63.45 & 63.88 & 68.45 & \underline{61.21 / 50.22}\\
 & BOBA (cls.) & \textbf{80.34} & 73.19 & -- & \textbf{80.45} & \underline{78.91} & 72.64 & \underline{75.60} & \textbf{79.59} & --\\
 & FedAA & 78.15 & 47.32$^{(4)}$ & 71.70$^{(2)}$ & 78.26 & 71.43 & \underline{75.34} & 60.20 & 78.39 & 30.43 / 23.53$^{(2)}$\\
\rowcolor{scalecolumn}  & \textbf{\methodname{}} & 79.12 & \textbf{79.51} & \textbf{79.41} & \underline{79.67} & \textbf{79.61} & \textbf{78.89} & \textbf{79.11} & \underline{79.12} & \textbf{78.31 / 70.00}\\
\midrule
\multirow{8}{*}{\shortstack[l]{\textbf{AGNews}\\GRU}} & q-FFL & 81.52 & 64.00 & -- & 27.06$^{(4)}$ & 73.23 & 40.49 & 45.40$^{(4)}$ & 76.35 & --\\
 & H-nobs & 81.52 & 26.69$^{(3)}$ & \underline{81.62} & 81.52 & 77.53 & 72.91 & 73.71 & 81.52 & 26.69 / 0.56$^{(3)}$\\
 & Trim. & 81.57 & 36.61 & 35.88 & 79.67 & 74.76 & 38.63 & 39.37 & 79.89 & 30.85 / 5.71\\
 & Krum & 72.69 & 77.17 & 76.62 & 72.69 & 72.18 & \underline{77.42} & \underline{74.77} & 72.69 & \underline{64.43 / 48.20}\\
 & FLTrust & \textbf{83.92} & \textbf{83.92} & \textbf{83.92} & \textbf{83.98} & 77.19 & 66.82 & 64.26 & \underline{83.03} & 64.26 / 45.46\\
 & BOBA (cls.) & \underline{83.82} & 60.57 & 27.65 & \underline{83.82} & \textbf{80.82} & 56.44 & 59.43 & \textbf{83.53} & 27.65 / 1.65\\
 & FedAA & 78.36 & 38.32 & 57.68$^{(2)}$ & 77.83 & 75.29 & 41.18 & 40.34 & 79.67 & 31.29 / 4.44$^{(2)}$\\
\rowcolor{scalecolumn}  & \textbf{\methodname{}} & 82.29 & \underline{80.46}$^{(4)}$ & 81.12 & 82.17 & \underline{78.62} & \textbf{81.44} & \textbf{81.36} & 82.29 & \textbf{77.84 / 63.75}$^{(4)}$\\
\bottomrule
\end{tabular}%
}
\end{table}

%% file: high_loss_matched_main.tex
\begin{table}[tb]
\centering\scriptsize\setlength{\tabcolsep}{5pt}\renewcommand{\arraystretch}{0.92}
\caption{\textbf{Whom each screen actually rejects, five seeds.} Left: the fraction (\%) of the highest-loss 20\% of honest submissions rejected, which fairness wants low. Right: the fraction (\%) of Byzantine submissions rejected, which robustness wants high. Both pool the first 300 rounds and the seven attacks within a seed, reported as mean$_{\pm\text{std}}$ over seeds 40--44; ``worst'' is the attack at which that column is worst in each seed. Best in bold, runner-up underlined.}
\label{tab:high-loss-main}
\begin{tabular}{llcccc}
\toprule
 & & \multicolumn{2}{c}{High-loss honest rejected $\downarrow$} & \multicolumn{2}{c}{Byzantine rejected $\uparrow$}\\
\cmidrule(lr){3-4}\cmidrule(lr){5-6}
Dataset/model & Method & Pooled & Worst & Pooled & Worst\\
\midrule
CIFAR-10/ResNet-18 & H-nobs & $34.95_{\scriptscriptstyle\pm1.62}$ & $90.67_{\scriptscriptstyle\pm2.53}$ & $\underline{63.31}_{\scriptscriptstyle\pm1.09}$ & $\underline{0.00}_{\scriptscriptstyle\pm0.00}$\\
 & BOBA (cls.) & $\underline{0.33}_{\scriptscriptstyle\pm0.27}$ & $\underline{1.67}_{\scriptscriptstyle\pm1.18}$ & $27.12_{\scriptscriptstyle\pm0.79}$ & $\underline{0.00}_{\scriptscriptstyle\pm0.00}$\\
 & \textbf{\methodname{}} & $\mathbf{0.19}_{\scriptscriptstyle\pm0.20}$ & $\mathbf{1.00}_{\scriptscriptstyle\pm0.91}$ & $\mathbf{98.71}_{\scriptscriptstyle\pm0.70}$ & $\mathbf{94.83}_{\scriptscriptstyle\pm2.16}$\\
\midrule
MNIST/MLP & H-nobs & $35.71_{\scriptscriptstyle\pm1.40}$ & $91.67_{\scriptscriptstyle\pm3.33}$ & $\underline{61.07}_{\scriptscriptstyle\pm1.43}$ & $\underline{0.00}_{\scriptscriptstyle\pm0.00}$\\
 & BOBA (cls.) & $\underline{5.62}_{\scriptscriptstyle\pm0.87}$ & $\underline{21.67}_{\scriptscriptstyle\pm2.04}$ & $36.81_{\scriptscriptstyle\pm2.40}$ & $\underline{0.00}_{\scriptscriptstyle\pm0.00}$\\
 & \textbf{\methodname{}} & $\mathbf{1.67}_{\scriptscriptstyle\pm1.04}$ & $\mathbf{5.67}_{\scriptscriptstyle\pm2.53}$ & $\mathbf{96.05}_{\scriptscriptstyle\pm1.56}$ & $\mathbf{85.83}_{\scriptscriptstyle\pm5.14}$\\
\midrule
EMNIST-Balanced/CNN & H-nobs & $\underline{23.95}_{\scriptscriptstyle\pm1.06}$ & $\underline{70.15}_{\scriptscriptstyle\pm7.32}$ & $\underline{66.13}_{\scriptscriptstyle\pm0.12}$ & $\underline{0.00}_{\scriptscriptstyle\pm0.00}$\\
 & BOBA (cls.) & $\mathbf{0.00}_{\scriptscriptstyle\pm0.00}$ & $\mathbf{0.00}_{\scriptscriptstyle\pm0.00}$ & $14.31_{\scriptscriptstyle\pm0.04}$ & $\underline{0.00}_{\scriptscriptstyle\pm0.00}$\\
 & \textbf{\methodname{}} & $\mathbf{0.00}_{\scriptscriptstyle\pm0.00}$ & $\mathbf{0.00}_{\scriptscriptstyle\pm0.00}$ & $\mathbf{99.81}_{\scriptscriptstyle\pm0.25}$ & $\mathbf{98.67}_{\scriptscriptstyle\pm1.73}$\\
\midrule
AGNews/GRU & H-nobs & $22.19_{\scriptscriptstyle\pm3.58}$ & $80.00_{\scriptscriptstyle\pm11.79}$ & $\underline{68.62}_{\scriptscriptstyle\pm0.56}$ & $0.00_{\scriptscriptstyle\pm0.00}$\\
 & BOBA (cls.) & $\mathbf{0.00}_{\scriptscriptstyle\pm0.00}$ & $\mathbf{0.00}_{\scriptscriptstyle\pm0.00}$ & $29.71_{\scriptscriptstyle\pm0.64}$ & $0.00_{\scriptscriptstyle\pm0.00}$\\
 & \textbf{\methodname{}} & $\underline{9.71}_{\scriptscriptstyle\pm1.42}$ & $\underline{68.00}_{\scriptscriptstyle\pm9.96}$ & $\mathbf{85.71}_{\scriptscriptstyle\pm0.00}$ & $0.00_{\scriptscriptstyle\pm0.00}$\\
\bottomrule
\end{tabular}
\end{table}

%% file: component_ablation_cifar_main.tex
\begin{table}[t]
\centering
\caption{\textbf{Component ablation of \methodname{}.} CIFAR-10/ResNet-18, seed 42, 1000 rounds. Attack columns give global accuracy (\%) over the final ten rounds; the worst-attack column gives global / worst-10\% client accuracy and is the componentwise minimum over the seven attacks. Clean uses $f=0$; ``--'' marks a run with no terminal window.}
\label{tab:component-ablation-main}
\scriptsize
\setlength{\tabcolsep}{2.2pt}
\renewcommand{\arraystretch}{0.88}
\resizebox{\textwidth}{!}{\begin{tabular}{lccc|ccccccccc}
\toprule
Variant & Scale & Cone & Loss norm. & Clean & Magnitude & Sign & IPM$_{\lambda=3}$ & Gauss. & ALIE & Min-Max & Min-Sum & Worst attack\\
\midrule
\textbf{\methodname{}} & \checkmark & \checkmark & \checkmark & \textbf{79.03} & \textbf{78.97} & \underline{77.50} & \textbf{77.83} & \textbf{78.82} & \underline{78.50} & \textbf{79.28} & \underline{78.71} & \textbf{77.50}/\textbf{68.37}\\
w/o scale & -- & \checkmark & \checkmark & \textbf{79.03} & 52.38 & \textbf{78.20} & \underline{76.08} & \textbf{78.82} & \textbf{78.97} & \underline{79.00} & \textbf{79.03} & 52.38/40.25\\
w/o cone & \checkmark & -- & \checkmark & \textbf{79.03} & \underline{78.67} & 69.95 & -- & \textbf{78.82} & 55.16 & 36.71 & 38.94 & --\\
w/o scale and cone & -- & -- & \checkmark & \textbf{79.03} & 63.23 & 70.16 & -- & 26.11 & 63.20 & 34.07 & 35.66 & --\\
w/o loss norm. & \checkmark & \checkmark & -- & \underline{78.04} & 70.06 & 73.33 & 74.01 & \underline{77.89} & 76.98 & 76.04 & 76.02 & \underline{70.06}/\underline{59.88}\\
cone only & -- & \checkmark & -- & \underline{78.04} & 10.89 & 71.59 & 75.30 & 77.65 & 76.87 & 78.04 & 78.04 & 10.89/0.05\\
scale only & \checkmark & -- & -- & \underline{78.04} & 73.14 & 71.66 & -- & 77.74 & 46.45 & 13.87 & -- & --\\
no component & -- & -- & -- & \underline{78.04} & 16.91 & 72.56 & -- & -- & 48.22 & 15.41 & 15.04 & --\\
\bottomrule
\end{tabular}}
\end{table}

%% file: sensitivity_budget_main.tex
\begin{figure}[t]
\centering
\includegraphics[width=0.68\linewidth]{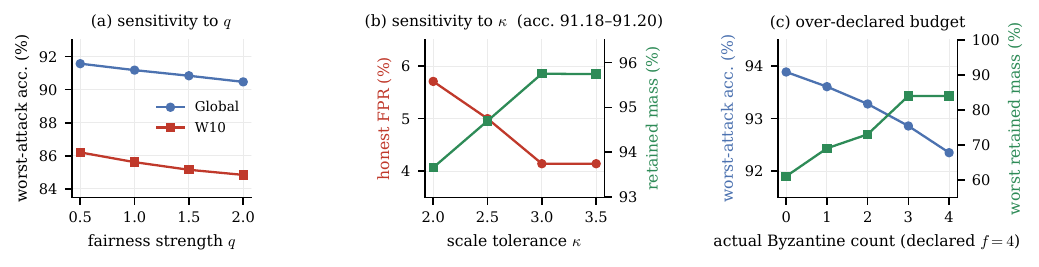}
\caption{\textbf{Sensitivity and budget misspecification on MNIST/MLP, seed 42.}
(a) Worst-attack accuracy over $q$. (b) Larger $\kappa$ lowers honest FPR and
raises retained fair mass, with accuracy within 91.18--91.20\%. (c) When the
declared budget stays at $f=4$ while the true number of Byzantine submissions
falls from 4 to 0, robustness is stable, but the surplus rejection budget is
spent on honest clients: the worst retained fair mass drops from 84\% to
61\%. Over-declaring $f$ therefore costs fairness rather than accuracy
(\Cref{app:diagnostics}).}
\label{fig:sensitivity-budget}
\end{figure}

%% file: appendix/appendix.tex
\clearpage
\appendix
\crefalias{section}{appendix}
\crefalias{subsection}{appendix}
\crefalias{subsubsection}{appendix}
\raggedbottom
\setcounter{topnumber}{4}
\setcounter{bottomnumber}{3}
\setcounter{totalnumber}{6}
\renewcommand{\topfraction}{0.92}
\renewcommand{\bottomfraction}{0.85}
\renewcommand{\textfraction}{0.06}
\renewcommand{\floatpagefraction}{0.78}
\setlength{\floatsep}{7pt plus 2pt minus 2pt}
\setlength{\textfloatsep}{8pt plus 2pt minus 2pt}
\setlength{\intextsep}{7pt plus 2pt minus 2pt}
\setlength{\abovecaptionskip}{4pt}
\setlength{\belowcaptionskip}{2pt}
\input{appendix/related_work.tex}
\input{appendix/experimental_setup.tex}
\input{appendix/additional_results.tex}
\input{appendix/mechanism_validation.tex}
\input{appendix/ablations_diagnostics.tex}
\input{appendix/theory.tex}

%% file: appendix/related_work.tex
\section{Related Work}
\label{app:related-work}

\textbf{Client-level fairness.}
Client-level fairness has been pursued through $q$-FFL, agnostic and tilted
objectives, multi-objective optimization, and proportional, coalitional, or
gradient-conflict principles~\citep{li2020fair,mohri2019agnostic,li2021term,
hu2022fedmgda,zhang2023propfair,chaudhury2022core,pan2024fedlf}. Yet, the
Byzantine robustness of these schemes has received limited attention. An
initial preliminary work was conducted by \citet{zhou2023hnobs}, who combine
fairness-aware learning with a norm filter, as does
\citet{basharat2025trustworthy}. However, a norm filter reads exactly the
quantity that fairness is designed to inflate, so it suppresses the high-loss
clients it is meant to protect. Closest in intent, FedAA learns aggregation
weights with a reinforcement agent driven by pairwise distances between client
models~\citep{he2025fedaa}; its weights are still a function of how far a
submission sits from the others, so it too never models the $F_i^q\Delta_i$
product that makes a high-loss honest client distant by construction. We
instead treat that ambiguity as the object of study and exploit the
class-mixture structure that survives fairness scaling.

\textbf{Byzantine distributed learning.}
Byzantine robustness has become a central topic in distributed learning, where
malicious clients may send arbitrary updates during
training~\citep{blanchard2017krum,yin2018robust,chen2017byzantine,
pillutla2022rfa,mhamdi2018bulyan,cao2021fltrust,shejwalkar2021manipulating,
li2019rsa,karimireddy2022bucketing,liu2023gradient,allouah2023heterogeneity}.
Robust aggregators in this line filter by distances, order statistics, trusted
directions, spectral structure, regularization, bucketing, or gradient
decomposition, all of which treat honest updates as concentrated around a
common mean. Our setting differs in that every honest $q$-fair update undergoes
a loss-dependent radial transformation before reaching the server, so honest
submissions are deliberately dispersed in magnitude.

\textbf{Structure-aware Byzantine aggregation.}
A parallel line of work exploits the structure of updates under label skew
rather than their concentration. BOBA, our closest structural comparison,
models unscaled updates with a unit-sum class-mixture
simplex~\citep{bao2024boba}. These approaches, however, assume normalized
honest coefficients, an assumption which does not hold here: fairness weighting
preserves nonnegativity but scales the coefficients by $F_i^q$. \methodname{}
therefore retains the radial mass as a screening statistic in its own right,
and normalizes the coefficients only when class composition is needed.

%% file: appendix/experimental_setup.tex
\section{Experimental Setup and Attack Details}
\label{app:setup}

\subsection{Datasets, models, and optimization}

The benchmark contains 100 honest clients and samples 10 honest clients per communication round. Attacked rounds append four Byzantine Sybil submissions, while clean rounds append none. Client data follow a Dirichlet label-skew partition with concentration $\alpha=1.0$ and class-availability probability $0.7$. Each honest client reserves 2\% of its local data for validation. All reported runs use one local epoch, a whole-client batch, SGD without momentum, and fairness strength $q=1$ unless a sensitivity experiment states otherwise.

\begin{table}[!htbp]
\centering
\small
\caption{Dataset, model, and optimization settings used by the main comparison.}
\label{tab:dataset-settings}
\resizebox{\textwidth}{!}{%
\begin{tabular}{lcccccc}
\toprule
Dataset & Model & LR & Rounds & Classes & Batch & Weight decay\\
\midrule
CIFAR-10 & ResNet-18 & 0.03 & 1500 & 10 & whole client & 0\\
MNIST & MLP (784--200--10) & 0.10 & 1000 & 10 & whole client & $5\times10^{-4}$\\
EMNIST-Balanced & five-layer CNN & 0.03 & 1000 & 47 & whole client & $5\times10^{-4}$\\
AGNews & one-layer GRU & 0.05 & 2000 & 4 & whole client & $5\times10^{-4}$\\
\bottomrule
\end{tabular}}
\end{table}

The five-layer CNN contains three unpadded $3\times3$ convolutions with 32, 64, and 64 output channels, followed by fully connected layers of width 64 and the dataset-specific classifier. MNIST uses one hidden layer with 200 ReLU units. AGNews uses 300-dimensional embeddings and a one-layer unidirectional GRU with hidden dimension 256; tokens with frequency below two are removed and sequences are truncated to length 128.

\subsection{Attack definitions and adversarial information}

The attack suite contains seven Byzantine attacks. Sign Flip submits $-\bar u_H$, where $\bar u_H$ is the current honest submitted mean. IPM submits $-3\bar u_H$. Gaussian uses standard deviation 200. ALIE uses the original-$z$ rule with configured value approximately $0.2533$ for 10 honest and four Byzantine submissions. Min-Max and Min-Sum use $\gamma_0=10$, search tolerance $10^{-5}$, and shared Sybils. Magnitude Scaling copies the sampled honest submission with the highest start loss and multiplies its magnitude by five.

The attack generator therefore receives the honest submitted mean for Sign Flip and IPM and simulator-side start losses for the Magnitude source choice. These quantities are not inputs to \methodname{}. The adversary knows the learning rule, model, defense, and hyperparameters but not the realized trusted examples or trusted class losses. Trusted server records are assumed clean.

\subsection{\methodname{} implementation}
\label{app:scale-implementation}

Each client uploads only $u_i=F_i(w_t)^q\Delta_i$. \methodname{} uses 20 unique trusted examples per class. CIFAR-10/ResNet-18 uses 24 stochastic augmentation views per trusted example; the other main dataset/model configurations use one view. The scale threshold is $\kappa=2.5$. The algorithm receives a round-specific declared budget $f_t$: the main benchmark uses $f_t=4$ in attacked conditions and $f_t=0$ in clean conditions, matching the number of Byzantine submissions. Appendix~\ref{app:always-on} evaluates the over-declared clean setting $f_t=4$.

Screening uses the classifier parameters for ResNet-18, MNIST/MLP, and
EMNIST/CNN, and all model parameters for AGNews/GRU. Accepted clients
contribute their original full-parameter submissions to
Eq.~\eqref{eq:aggregation}. To recover coefficients efficiently, the
implementation forms
$
K=G^\top G,\qquad b_i=G^\top v_i,
$
removes numerically degenerate reference directions, and solves NNLS in the retained at-most-$C$-dimensional Gram space. Let $\widehat a_i$ denote this implementation-level NNLS solution. Forming the inner products still depends on the selected parameter dimension; only the constrained numerical solve is low dimensional.

The trusted start-of-round class-loss vector $\widehat\ell$ is evaluated on the same broadcast model as the client fairness losses. The final method uses
$
\widetilde F_i
=
\widehat\ell^\top
\frac{\widehat a_i}{\max\{\|\widehat a_i\|_1,\epsilon\}},
\qquad
d_i=\max\{1,\widetilde F_i^q\},
$
No client loss or additional scalar is communicated.

\subsection{Baselines and information compatibility}

q-FFL-GD averages the same fairness-scaled gradient carrier used by the benchmark. H-nobs follows its norm-based screening mechanism: it removes the declared number of largest-$\ell_2$ fair gradients and averages the retained vectors. Median, Trimmed Mean, Krum, Multi-Krum, and DnC operate on the same submitted fair vector. FLTrust additionally constructs a clean server root gradient. FedAA and Ditto retain their method-specific aggregation or personalization mechanisms.

All matched DnC runs triggered the registered collapse criterion before their
final reporting windows and are therefore not included in the main endpoint
tables. BOBA is evaluated over the matched five seeds with classifier-coordinate
geometry; the original full-parameter implementation is
reproduced for seed 42 only and is compared with the classifier-coordinate
variant in Appendix~\ref{app:boba-fidelity}.
Table~\ref{tab:baseline-compatibility} summarizes the information-interface
compatibility of the evaluated methods under our benchmark.

\begin{table}[!htbp]
\centering
\setlength{\tabcolsep}{4.2pt}
\renewcommand{\arraystretch}{1.05}
\small

\caption{
Baseline information compatibility.
``Same fair update'' indicates whether the method is applied to the
fairness-scaled client submission $u_i = F_i^q \Delta_i$ used in our benchmark.
A dash indicates that the corresponding detail is not documented or cannot be
verified from the current manuscript repository.
}
\label{tab:baseline-compatibility}

\resizebox{\textwidth}{!}{
\begin{tabular}{lcccc}
\toprule
Method
& Same fair update $u_i$
& Extra trusted server data
& Output
& Uses declared $f$ \\
\midrule

q-FFL-GD (mean)
& Yes
& No
& Shared global
& No \\

H-nobs
& Yes
& No
& Shared global
& Yes \\

Trimmed Mean
& Yes
& No
& Shared global
& -- \\

Median
& Yes
& No
& Shared global
& No documented dependence \\

Krum
& Yes
& No
& Shared global
& -- \\

Multi-Krum
& Yes
& No
& Shared global
& -- \\

FLTrust
& Yes
& Yes, clean root gradient
& Shared global
& -- \\

FedAA
& --
& --
& Shared aggregation
& -- \\

Ditto
& --
& No
& Personalized
& -- \\

DnC
& Yes
& No
& Shared global
& -- \\

BOBA
& Yes
& Yes, per-class references from the same pool
& Shared global
& Yes \\

\textbf{\methodname{}}
& Yes
& Yes, clean labeled server data
& Shared global
& Yes \\

\bottomrule
\end{tabular}
}
\end{table}

\subsection{BOBA implementation and full-space fidelity check}
\label{app:boba-fidelity}

BOBA~\citep{bao2024boba} fits an affine subspace and a class-mixture simplex
to the received updates and accepts the submissions whose simplex fit is
plausible. In the original implementation both fits are computed in the full
parameter space; on ResNet-18 this costs 30--47 GPU-hours per 1500-round run
(20 Stage-1 subspace iterations on a $14\times11$M matrix), which is why the
matched five-seed benchmark uses classifier-coordinate BOBA: the Stage-1 subspace, the
simplex coefficients, and the acceptance decision are computed on the
classifier block ($5{,}130$ coordinates for ResNet-18, $2{,}010$ for the MLP,
$68{,}655$ for the CNN, $1{,}028$ for the GRU), the classifier coordinates are
aggregated by BOBA's projection, and all other parameters are averaged over
the accepted clients. For the three vision benchmarks, this matches the
coordinate choice used by \methodname{} for screening, so the two methods see
the same geometry there. All other BOBA
settings follow the reference implementation ($p_{\min}=-0.5$, 20 Stage-1
iterations, the same 20 trusted examples per class as \methodname{} and
FLTrust).

Table~\ref{tab:boba-full-seed42} reports the full-parameter reproduction for
the only seed on which it exists. It is a sensitivity and fidelity check, not
the primary statistical comparison, and a single seed does not support a
statistical claim in either direction. Where both CIFAR-10 variants are
available for seed 42, the classifier-coordinate variant is not uniformly
weaker: it is stronger under clean training (82.19 vs.\ 78.34), Sign Flip
(72.50 vs.\ 11.01), Gaussian (81.66 vs.\ 76.29), ALIE (73.70 vs.\ 24.27),
and Magnitude (72.02 vs.\ 62.50), while the full-space variant is stronger
under Min-Max (45.57 vs.\ 17.53) and Min-Sum (65.26 vs.\ 20.15). Both fail
severely under IPM (11.84 for the classifier variant; the full-space run
collapsed at round 18), and the full-space MNIST and EMNIST IPM runs also
collapsed or ended near chance (2.69). The five-seed BOBA rows in
Tables~\ref{tab:main-results} and
\ref{tab:cifar10-meanstd-5seed}--\ref{tab:agnews-meanstd-5seed} are the
numbers the paper's comparison rests on.

\begin{table}[!htbp]
\centering\small
\setlength{\tabcolsep}{2.5pt}
\caption{\textbf{Full-parameter BOBA, seed 42 (fidelity check).} Affine
subspace and simplex fitted in the full parameter space; global / worst-10\%
accuracy (\%) over the registered final window. ``collapsed ($r$)'' =
terminated by the registered collapse rule at round $r$; ``n/a'' = not run;
``incomplete'' = did not reach the endpoint within the compute budget.}
\label{tab:boba-full-seed42}
\resizebox{\textwidth}{!}{\begin{tabular}{lccccccccc}
\toprule
Dataset & Clean & Sign & IPM & Gauss. & ALIE & Min-Max & Min-Sum & Mag. & Worst attack\\
\midrule
CIFAR-10 / ResNet-18 & 78.34/68.84 & 11.01/0.30 & collapsed (18) & 76.29/66.97 & 24.27/11.68 & 45.57/33.37 & 65.26/53.62 & 62.50/51.41 & 11.01/0.30\\
MNIST / MLP & n/a & 92.57/87.26 & collapsed (17) & n/a & collapsed (779) & n/a & n/a & 94.73/90.52 & --\\
EMNIST-Balanced / CNN & n/a & n/a & 2.69/0.17 & n/a & n/a & n/a & n/a & n/a & --\\
AGNews / GRU & n/a & n/a & incomplete (1047) & n/a & n/a & n/a & n/a & n/a & --\\
\bottomrule
\end{tabular}}
\end{table}

\subsection{Metrics and reporting}

We report global test accuracy and worst-10\% client accuracy. For each seed, both metrics are averaged over the final ten rounds: 1490--1499 for CIFAR-10, 990--999 for MNIST and EMNIST-Balanced, and 1990--1999 for AGNews. Main tables then average the seed-level window means over seeds 40--44 that reach the registered endpoint. Superscript $(k)$ indicates that only $k$ seeds reach the endpoint; a dash indicates that no valid complete result is available. Pre-collapse maxima are never substituted.

The main worst-attack result is seed-wise: for every seed completing all seven attacks, we first take the minimum attacked accuracy and then average these minima across seeds. This conservative statistic supports paired comparisons and differs from the minimum of the seven attack-column means. Clean runs use $f_t=0$. Selection recall, honest FPR, retained fairness mass, and loss-quantile audits use simulator-side labels or losses only for offline evaluation; they are not selector inputs. The high-loss retention audit in Table~\ref{tab:high-loss-main} uses all five registered seeds and pools rounds $r$ with $(r+1)\bmod 10=0$, the grid common to every logging cadence; the H-nobs fairness comparison and the loss-quantile breakdown in Table~\ref{tab:loss-quantile-full} remain seed-42 audits. A collapsed trajectory contributes diagnostic observations only through its registered stopping round.

\subsection{Three-component ablation protocol}

The factorial ablation traverses all $2^3$ combinations of the scale screen,
cone screen, and loss-normalized aggregation.  All four dataset/model pairs
use 1000 communication rounds under matched client partitions and attack
realizations.  The complete $2^3$ factorial ablation for all four dataset/model pairs is
reported in Appendix~\ref{app:factorial-ablation}; the main text shows the
CIFAR-10 instance as a representative mechanism study.  Auxiliary runs that reached
round 299 were continued from their matched checkpoints.  Clean trajectories
are reused across selector-component rows within an aggregation mode because
clean runs use $f=0$ and therefore perform no screening rejection.

After this reuse, 58 CIFAR-10, 53 EMNIST-Balanced, and 60 AGNews conceptual
conditions reach their rounds 990--999 reporting windows.  A missing terminal
window is retained as ``--'' rather than replaced by a pre-collapse maximum;
all such dashes correspond to a registered algorithmic collapse, not an
unfinished scheduler job.

%% file: appendix/additional_results.tex
\section{Five-Seed Results and Training Dynamics}
\label{app:results}

\subsection{Full five-seed endpoint results}
\label{app:full-main-results}

Tables~\ref{tab:cifar10-meanstd-5seed}--\ref{tab:agnews-meanstd-5seed}
report all ten baselines and \methodname{} over seeds 40--44. Each attacked
column first averages the final reporting window within a seed and then
averages completed seeds. The last column instead computes the minimum across
all seven attacks within each seed before averaging; it is therefore the
seed-wise worst-attack statistic used throughout the main paper. Superscript $(k)$
indicates that only $k$ seeds reached the endpoint. The BOBA rows use
classifier-coordinate geometry (Appendix~\ref{app:boba-fidelity});
full-space BOBA is available for seed 42 only. The AGNews \methodname{}
rows screen all GRU parameters.

\input{appendix/results_5seed_meanstd.tex}

Table~\ref{tab:w10-main-results} mirrors the main-paper robustness table using
worst-10\% client accuracy instead of global accuracy. Table~\ref{tab:worst-attack-paired}
then reports paired seed-wise comparisons for the global worst-attack metric.

\paragraph{Paired interpretation.}
Against FLTrust, \methodname{} improves the seed-wise global worst-attack accuracy by
30.07 points on CIFAR-10, 3.90 on MNIST, and 17.09 on EMNIST; the paired
differences are positive on all five seeds and their bootstrap 95\% intervals
exclude zero. The exact two-sided sign-flip test is nevertheless limited to
$p=0.0625$ with five pairs. On AGNews the floor is 77.84 over the four seeds
completing all attacks, 13.03 points above FLTrust ([+11.84, +14.22]) and
15.40 above Krum ([+4.62, +26.18]) on the same seeds; with four pairs the
sign-flip test is limited to $p=0.125$.

\subsection{Training dynamics}
\label{app:training-dynamics}

\begin{figure}[p]
\centering
\includegraphics[width=0.96\textwidth]{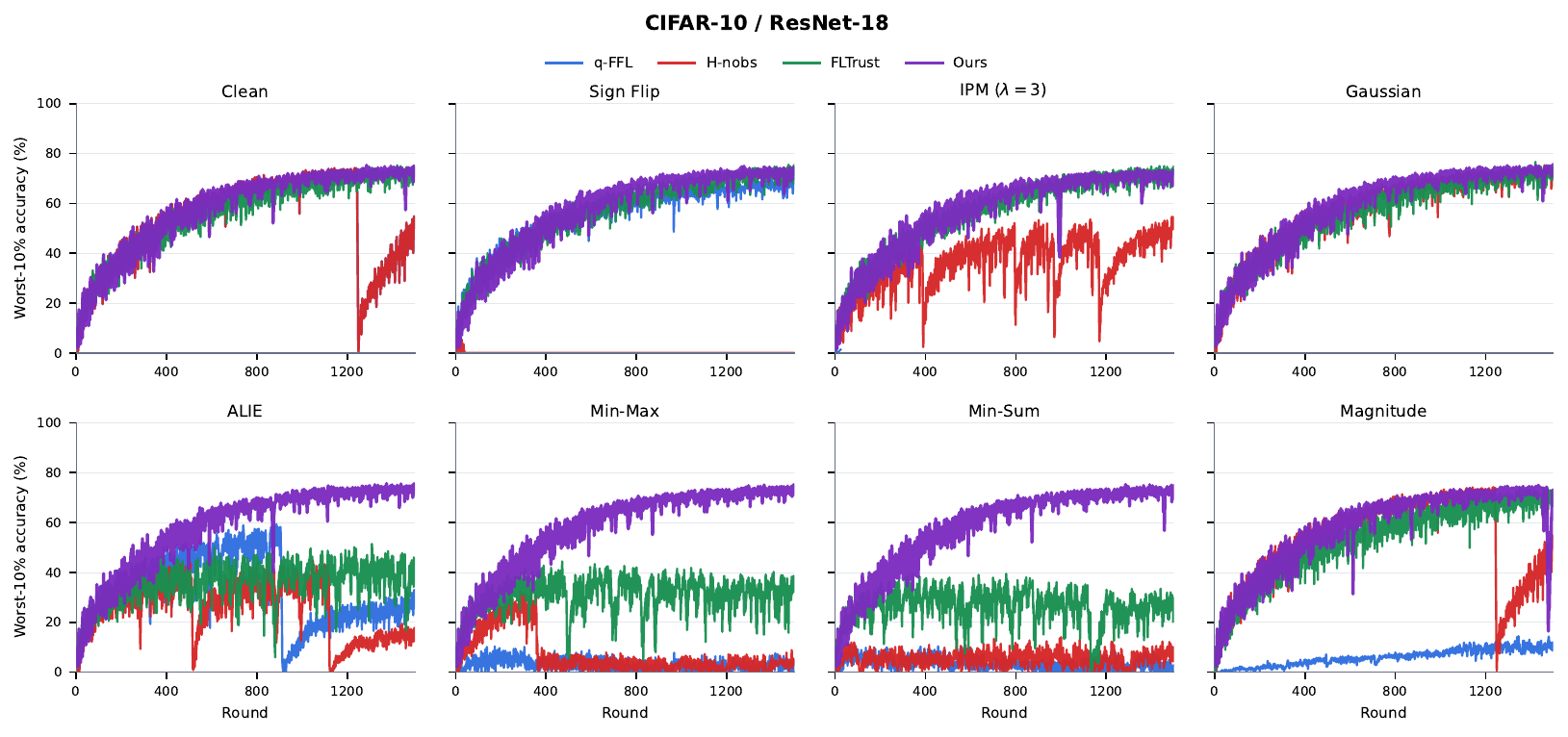}
\caption{\textbf{CIFAR-10/ResNet-18 training dynamics.} Worst-10\% client
accuracy (\%) over 1500 communication rounds.}
\label{fig:training-cifar}
\end{figure}

\begin{figure}[p]
\centering
\includegraphics[width=0.96\textwidth]{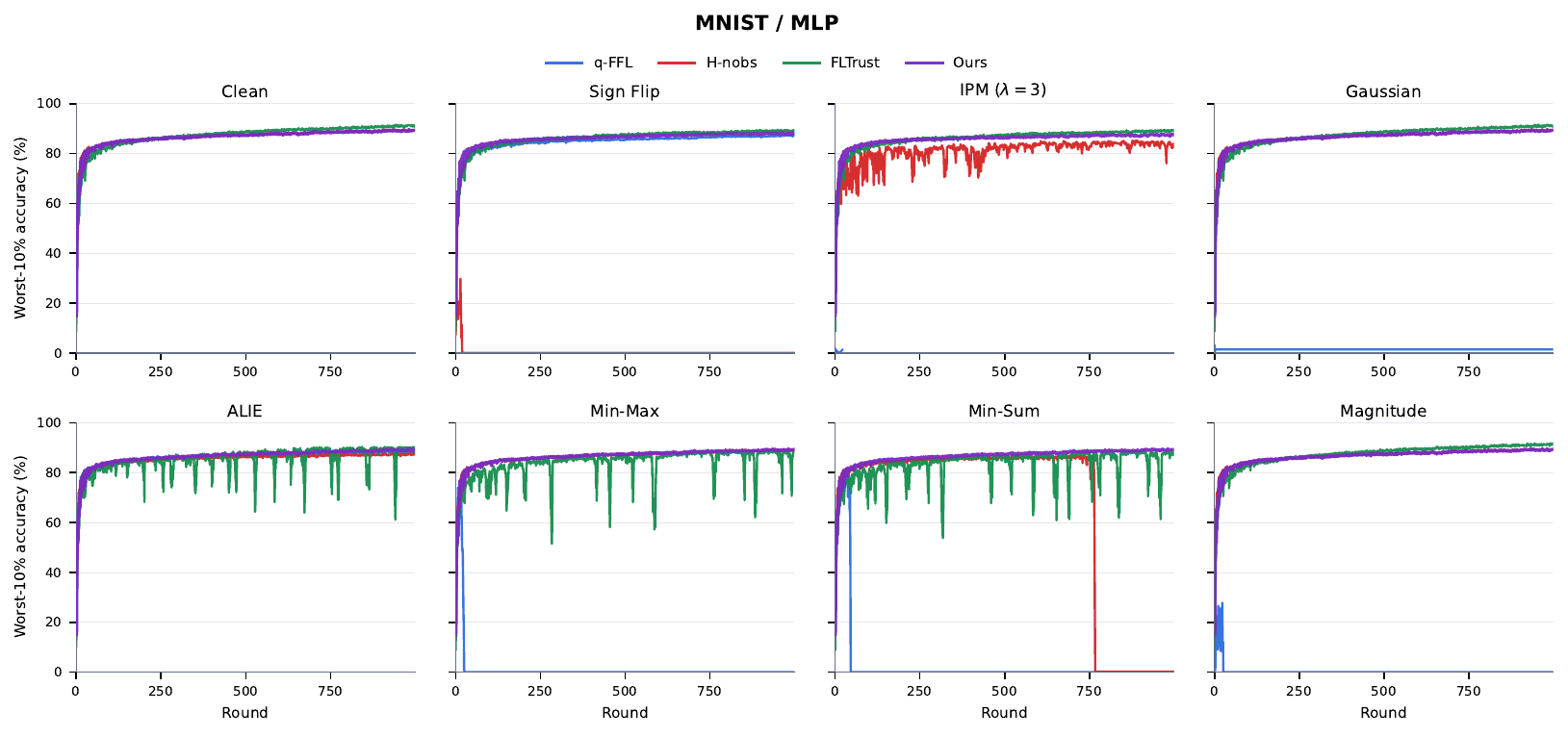}
\caption{\textbf{MNIST/MLP training dynamics.} Worst-10\% client accuracy
(\%) over 1000 communication rounds.}
\label{fig:training-mnist}
\end{figure}

\begin{figure}[p]
\centering
\includegraphics[width=0.96\textwidth]{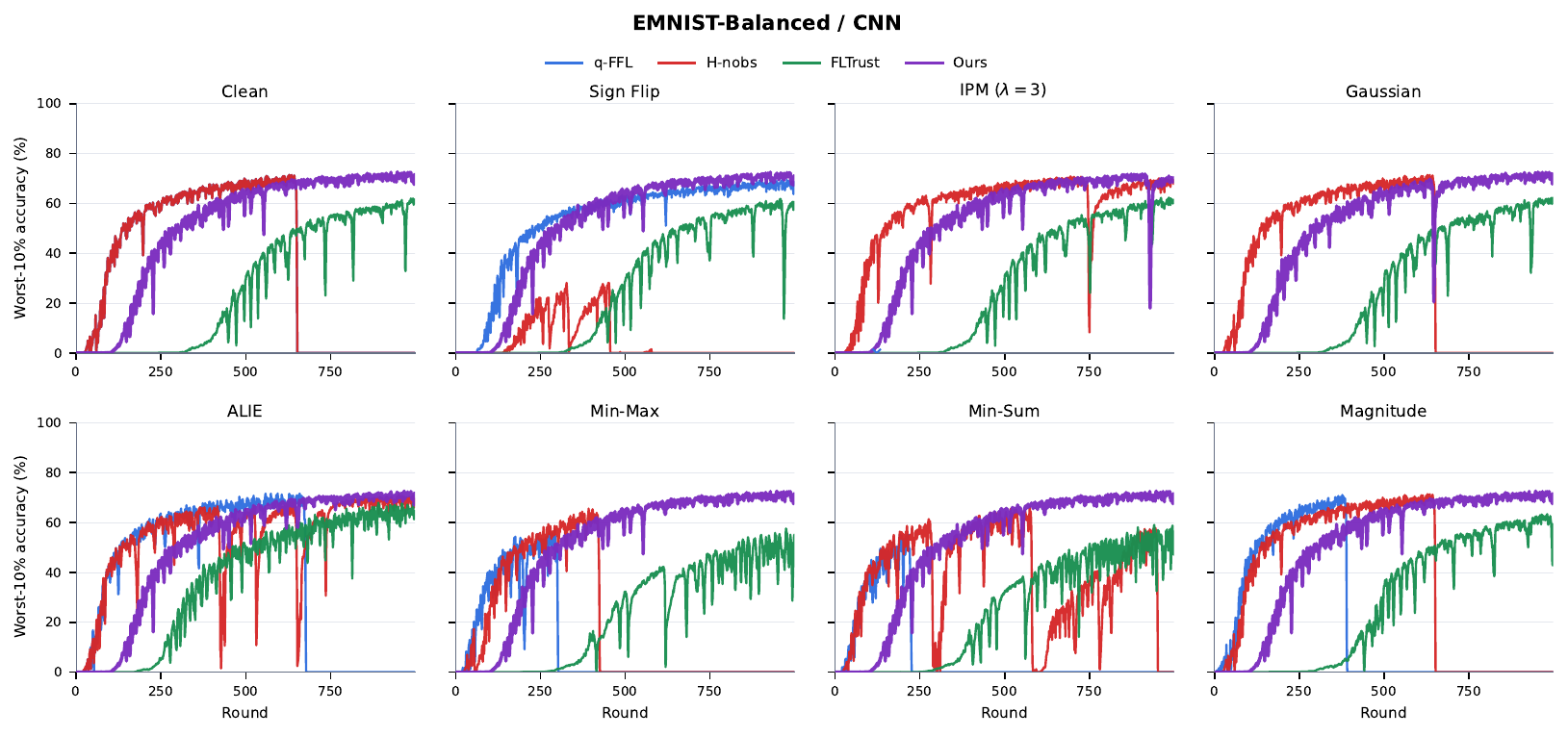}
\caption{\textbf{EMNIST-Balanced/CNN training dynamics.} Worst-10\% client
accuracy (\%) over 1000 communication rounds.}
\label{fig:training-emnist}
\end{figure}

\begin{figure}[p]
\centering
\includegraphics[width=0.96\textwidth]{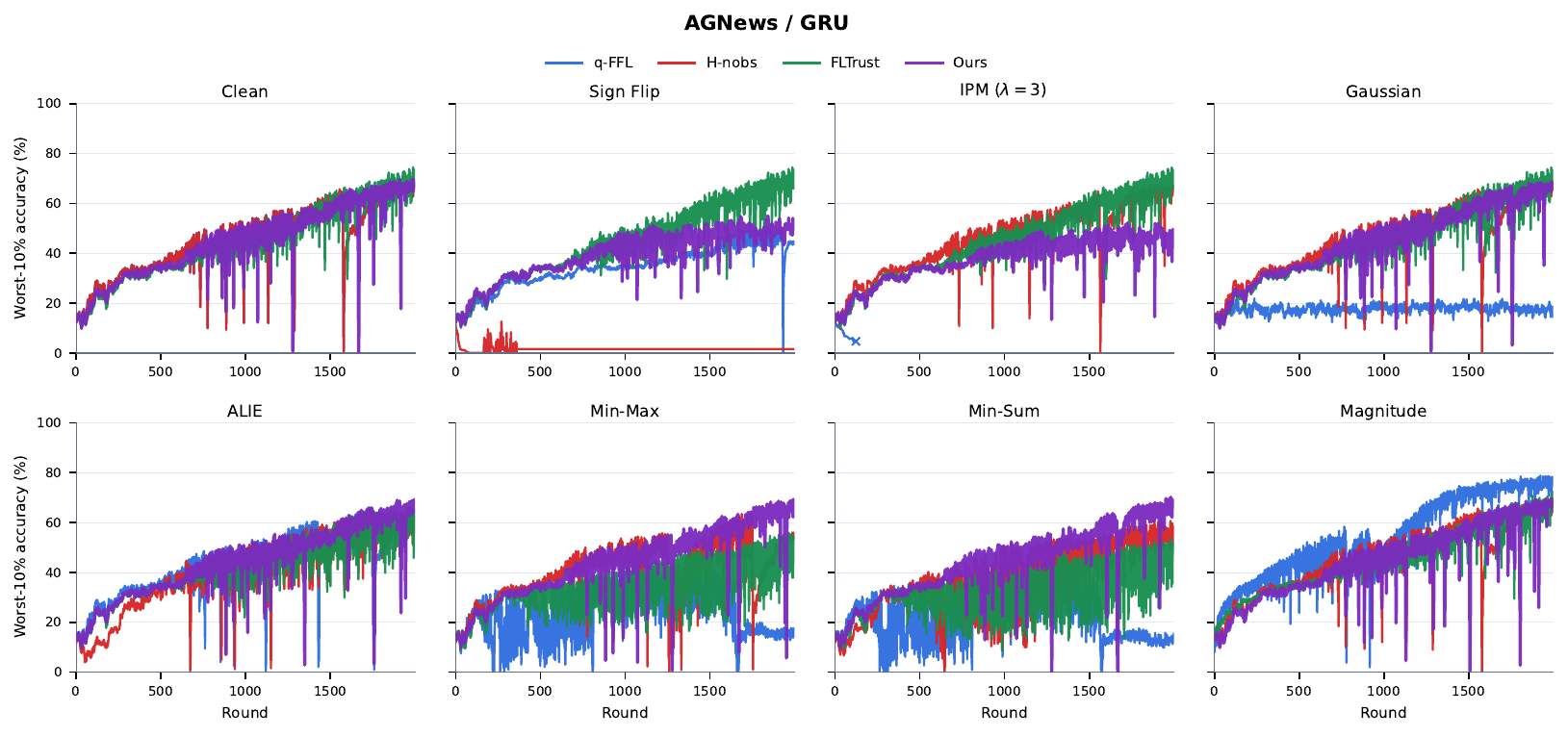}
\caption{\textbf{AGNews/GRU training dynamics.} Worst-10\% client accuracy
(\%) over 2000 communication rounds.}
\label{fig:training-agnews}
\end{figure}

Figures~\ref{fig:training-cifar}--\ref{fig:training-agnews} show the
seed-42 worst-10\% client-accuracy trajectories for every dataset/model pair
and all seven attacks, together with the clean condition. We retain three
representative baselines spanning fair optimization, norm screening, and
trusted-gradient aggregation. ``Ours'' denotes \methodname{}. A cross marks a
trajectory that terminates before the registered training horizon.

\FloatBarrier

\FloatBarrier

%% file: appendix/results_5seed_meanstd.tex

\begin{table}[!t]
\centering\setlength{\tabcolsep}{1.0pt}\renewcommand{\arraystretch}{0.9}\scriptsize
\caption{CIFAR-10 / ResNet-18, rounds 1490--1499. Entries are global / worst-10\% client accuracy (\%), mean over seeds 40, 41, 42, 43, 44 with the across-seed standard deviation as subscript; the worst-attack column is the seed-wise minimum over the seven attacks, also mean with std subscript, over seeds completing all seven attacks.}
\label{tab:cifar10-meanstd-5seed}
\resizebox{\textwidth}{!}{\begin{tabular}{lccccccccc}
\toprule
Method & Clean & Sign & IPM$_{\lambda=3}$ & Gauss. & ALIE & Min-Max & Min-Sum & Magnitude & Worst attack\\
\midrule
q-FFL-GD (mean) & $76.96_{\scriptscriptstyle\pm8.6}$/$67.34_{\scriptscriptstyle\pm9.9}$ & $76.47_{\scriptscriptstyle\pm2.1}$/$66.94_{\scriptscriptstyle\pm2.7}$ & - & $13.81_{\scriptscriptstyle\pm3.3}$/$4.71_{\scriptscriptstyle\pm4.2}$$^{(3)}$ & $63.43_{\scriptscriptstyle\pm13.4}$/$51.90_{\scriptscriptstyle\pm14.7}$ & $14.76_{\scriptscriptstyle\pm1.4}$/$2.24_{\scriptscriptstyle\pm0.6}$$^{(4)}$ & $14.05_{\scriptscriptstyle\pm1.9}$/$2.04_{\scriptscriptstyle\pm0.9}$ & $43.45_{\scriptscriptstyle\pm31.2}$/$32.27_{\scriptscriptstyle\pm31.3}$$^{(4)}$ & -\\
H-nobs & $76.96_{\scriptscriptstyle\pm8.6}$/$67.34_{\scriptscriptstyle\pm9.9}$ & $12.19_{\scriptscriptstyle\pm1.9}$/$0.35_{\scriptscriptstyle\pm0.5}$$^{(3)}$ & $57.13_{\scriptscriptstyle\pm24.6}$/$46.25_{\scriptscriptstyle\pm24.8}$ & $79.28_{\scriptscriptstyle\pm3.7}$/$70.03_{\scriptscriptstyle\pm4.4}$ & $35.65_{\scriptscriptstyle\pm20.3}$/$23.39_{\scriptscriptstyle\pm19.8}$ & $42.86_{\scriptscriptstyle\pm32.1}$/$31.99_{\scriptscriptstyle\pm33.3}$ & $20.15_{\scriptscriptstyle\pm11.2}$/$8.34_{\scriptscriptstyle\pm9.5}$ & $76.96_{\scriptscriptstyle\pm8.6}$/$67.34_{\scriptscriptstyle\pm9.9}$ & $12.19_{\scriptscriptstyle\pm1.9}$/$0.35_{\scriptscriptstyle\pm0.5}$$^{(3)}$\\
Trimmed Mean & $70.56_{\scriptscriptstyle\pm18.4}$/$60.32_{\scriptscriptstyle\pm20.4}$ & $59.66_{\scriptscriptstyle\pm7.7}$/$46.72_{\scriptscriptstyle\pm9.3}$ & $44.84_{\scriptscriptstyle\pm13.6}$/$31.29_{\scriptscriptstyle\pm15.2}$ & $79.45_{\scriptscriptstyle\pm3.5}$/$70.42_{\scriptscriptstyle\pm3.4}$ & $39.27_{\scriptscriptstyle\pm9.6}$/$25.86_{\scriptscriptstyle\pm8.8}$ & $13.37_{\scriptscriptstyle\pm2.1}$/$1.31_{\scriptscriptstyle\pm1.2}$ & $14.06_{\scriptscriptstyle\pm1.8}$/$2.08_{\scriptscriptstyle\pm1.1}$ & $74.35_{\scriptscriptstyle\pm7.3}$/$64.41_{\scriptscriptstyle\pm8.6}$ & $13.18_{\scriptscriptstyle\pm1.9}$/$1.28_{\scriptscriptstyle\pm1.1}$\\
Median & $45.02_{\scriptscriptstyle\pm16.4}$/$31.89_{\scriptscriptstyle\pm15.2}$ & $27.99_{\scriptscriptstyle\pm2.4}$/$14.01_{\scriptscriptstyle\pm2.4}$ & $23.83_{\scriptscriptstyle\pm2.5}$/$10.14_{\scriptscriptstyle\pm2.8}$ & $41.65_{\scriptscriptstyle\pm14.1}$/$28.89_{\scriptscriptstyle\pm13.2}$ & $24.71_{\scriptscriptstyle\pm10.7}$/$12.87_{\scriptscriptstyle\pm9.3}$ & $12.85_{\scriptscriptstyle\pm1.5}$/$1.32_{\scriptscriptstyle\pm0.8}$ & $12.73_{\scriptscriptstyle\pm2.5}$/$1.22_{\scriptscriptstyle\pm1.0}$ & $47.89_{\scriptscriptstyle\pm9.8}$/$34.11_{\scriptscriptstyle\pm10.1}$ & $12.34_{\scriptscriptstyle\pm1.9}$/$1.10_{\scriptscriptstyle\pm0.9}$\\
Krum & $63.66_{\scriptscriptstyle\pm17.4}$/$51.97_{\scriptscriptstyle\pm19.3}$ & $44.69_{\scriptscriptstyle\pm7.9}$/$29.35_{\scriptscriptstyle\pm6.8}$ & $62.78_{\scriptscriptstyle\pm17.7}$/$51.13_{\scriptscriptstyle\pm19.6}$ & $60.21_{\scriptscriptstyle\pm20.7}$/$48.53_{\scriptscriptstyle\pm21.7}$ & $22.24_{\scriptscriptstyle\pm11.2}$/$10.14_{\scriptscriptstyle\pm10.2}$ & $61.58_{\scriptscriptstyle\pm18.0}$/$50.09_{\scriptscriptstyle\pm19.5}$ & $45.62_{\scriptscriptstyle\pm14.3}$/$32.93_{\scriptscriptstyle\pm15.0}$ & $63.56_{\scriptscriptstyle\pm17.3}$/$51.66_{\scriptscriptstyle\pm18.9}$ & $21.67_{\scriptscriptstyle\pm10.4}$/$9.78_{\scriptscriptstyle\pm9.7}$\\
Multi-Krum & $80.15_{\scriptscriptstyle\pm3.6}$/$71.05_{\scriptscriptstyle\pm4.6}$ & $15.49_{\scriptscriptstyle\pm4.9}$/$3.79_{\scriptscriptstyle\pm3.3}$$^{(3)}$ & $68.36_{\scriptscriptstyle\pm3.1}$/$57.04_{\scriptscriptstyle\pm2.4}$ & $81.90_{\scriptscriptstyle\pm1.5}$/$73.17_{\scriptscriptstyle\pm1.8}$ & $39.70_{\scriptscriptstyle\pm10.9}$/$27.01_{\scriptscriptstyle\pm10.2}$ & $17.25_{\scriptscriptstyle\pm5.7}$/$4.71_{\scriptscriptstyle\pm4.7}$ & $19.64_{\scriptscriptstyle\pm4.9}$/$7.13_{\scriptscriptstyle\pm3.8}$ & $80.15_{\scriptscriptstyle\pm3.6}$/$71.05_{\scriptscriptstyle\pm4.6}$ & $12.97_{\scriptscriptstyle\pm2.6}$/$1.48_{\scriptscriptstyle\pm1.3}$$^{(3)}$\\
FLTrust & $81.31_{\scriptscriptstyle\pm0.3}$/$72.22_{\scriptscriptstyle\pm0.5}$ & $81.29_{\scriptscriptstyle\pm0.3}$/$72.26_{\scriptscriptstyle\pm0.4}$ & $81.31_{\scriptscriptstyle\pm0.3}$/$72.27_{\scriptscriptstyle\pm0.4}$ & $81.25_{\scriptscriptstyle\pm0.4}$/$72.16_{\scriptscriptstyle\pm0.5}$ & $55.13_{\scriptscriptstyle\pm1.9}$/$42.07_{\scriptscriptstyle\pm1.5}$ & $48.84_{\scriptscriptstyle\pm1.8}$/$35.47_{\scriptscriptstyle\pm1.8}$ & $43.81_{\scriptscriptstyle\pm2.5}$/$30.53_{\scriptscriptstyle\pm2.5}$ & $79.44_{\scriptscriptstyle\pm1.0}$/$70.05_{\scriptscriptstyle\pm1.1}$ & $43.81_{\scriptscriptstyle\pm2.5}$/$30.53_{\scriptscriptstyle\pm2.5}$\\
BOBA (cls.) & $81.93_{\scriptscriptstyle\pm1.1}$/$73.13_{\scriptscriptstyle\pm1.2}$ & $72.68_{\scriptscriptstyle\pm1.8}$/$62.12_{\scriptscriptstyle\pm1.8}$ & $10.93_{\scriptscriptstyle\pm0.7}$/$0.66_{\scriptscriptstyle\pm0.8}$ & $81.58_{\scriptscriptstyle\pm1.3}$/$72.73_{\scriptscriptstyle\pm1.6}$ & $72.00_{\scriptscriptstyle\pm3.6}$/$60.85_{\scriptscriptstyle\pm3.8}$ & $17.31_{\scriptscriptstyle\pm0.8}$/$5.27_{\scriptscriptstyle\pm1.0}$ & $17.83_{\scriptscriptstyle\pm1.9}$/$5.35_{\scriptscriptstyle\pm2.2}$ & $73.64_{\scriptscriptstyle\pm1.5}$/$62.39_{\scriptscriptstyle\pm2.1}$ & $10.93_{\scriptscriptstyle\pm0.7}$/$0.66_{\scriptscriptstyle\pm0.8}$\\
FedAA & $75.07_{\scriptscriptstyle\pm1.1}$/$64.87_{\scriptscriptstyle\pm1.3}$ & $45.32_{\scriptscriptstyle\pm19.1}$/$32.48_{\scriptscriptstyle\pm19.3}$ & $74.07_{\scriptscriptstyle\pm2.6}$/$62.88_{\scriptscriptstyle\pm3.7}$$^{(4)}$ & $72.60_{\scriptscriptstyle\pm4.2}$/$61.23_{\scriptscriptstyle\pm5.8}$ & $61.44_{\scriptscriptstyle\pm18.1}$/$49.78_{\scriptscriptstyle\pm19.4}$ & $22.54_{\scriptscriptstyle\pm6.3}$/$10.34_{\scriptscriptstyle\pm5.7}$ & $43.07_{\scriptscriptstyle\pm18.9}$/$30.00_{\scriptscriptstyle\pm19.9}$ & $66.07_{\scriptscriptstyle\pm6.1}$/$53.75_{\scriptscriptstyle\pm7.5}$ & $22.15_{\scriptscriptstyle\pm7.2}$/$10.27_{\scriptscriptstyle\pm6.4}$$^{(4)}$\\
Ditto & $81.73_{\scriptscriptstyle\pm1.1}$/$68.94_{\scriptscriptstyle\pm2.6}$ & $75.10_{\scriptscriptstyle\pm1.2}$/$24.59_{\scriptscriptstyle\pm5.9}$ & - & $15.55_{\scriptscriptstyle\pm0.9}$/$5.25_{\scriptscriptstyle\pm1.8}$ & $75.21_{\scriptscriptstyle\pm1.8}$/$35.95_{\scriptscriptstyle\pm3.7}$ & $33.69_{\scriptscriptstyle\pm1.2}$/$16.73_{\scriptscriptstyle\pm0.6}$ & $34.39_{\scriptscriptstyle\pm1.9}$/$19.40_{\scriptscriptstyle\pm3.2}$ & $75.52_{\scriptscriptstyle\pm3.0}$/$63.50_{\scriptscriptstyle\pm3.1}$ & -\\
\textbf{\methodname{}} & $82.25_{\scriptscriptstyle\pm0.5}$/$73.88_{\scriptscriptstyle\pm0.8}$ & $79.29_{\scriptscriptstyle\pm4.9}$/$70.15_{\scriptscriptstyle\pm5.9}$ & $80.59_{\scriptscriptstyle\pm1.1}$/$71.62_{\scriptscriptstyle\pm1.4}$ & $82.30_{\scriptscriptstyle\pm0.5}$/$73.91_{\scriptscriptstyle\pm0.8}$ & $77.74_{\scriptscriptstyle\pm10.2}$/$68.35_{\scriptscriptstyle\pm12.1}$ & $82.17_{\scriptscriptstyle\pm0.5}$/$73.91_{\scriptscriptstyle\pm0.9}$ & $82.19_{\scriptscriptstyle\pm0.5}$/$73.87_{\scriptscriptstyle\pm0.8}$ & $81.12_{\scriptscriptstyle\pm2.4}$/$72.67_{\scriptscriptstyle\pm3.1}$ & $73.88_{\scriptscriptstyle\pm9.1}$/$63.58_{\scriptscriptstyle\pm10.7}$\\
\bottomrule
\end{tabular}}
\end{table}

\begin{table}[!t]
\centering\setlength{\tabcolsep}{1.0pt}\renewcommand{\arraystretch}{0.9}\scriptsize
\caption{MNIST / MLP, rounds 990--999. Entries are global / worst-10\% client accuracy (\%), mean over seeds 40, 41, 42, 43, 44 with the across-seed standard deviation as subscript; the worst-attack column is the seed-wise minimum over the seven attacks, also mean with std subscript, over seeds completing all seven attacks.}
\label{tab:mnist-meanstd-5seed}
\resizebox{\textwidth}{!}{\begin{tabular}{lccccccccc}
\toprule
Method & Clean & Sign & IPM$_{\lambda=3}$ & Gauss. & ALIE & Min-Max & Min-Sum & Magnitude & Worst attack\\
\midrule
q-FFL-GD (mean) & $94.90_{\scriptscriptstyle\pm1.3}$/$90.72_{\scriptscriptstyle\pm2.1}$ & $93.24_{\scriptscriptstyle\pm1.1}$/$88.31_{\scriptscriptstyle\pm1.5}$ & - & $62.99_{\scriptscriptstyle\pm42.5}$/$54.32_{\scriptscriptstyle\pm45.8}$$^{(3)}$ & $94.29_{\scriptscriptstyle\pm1.1}$/$89.66_{\scriptscriptstyle\pm1.6}$ & $48.54_{\scriptscriptstyle\pm33.0}$/$37.24_{\scriptscriptstyle\pm32.5}$$^{(3)}$ & $56.03_{\scriptscriptstyle\pm40.0}$/$46.60_{\scriptscriptstyle\pm40.4}$$^{(3)}$ & $68.76_{\scriptscriptstyle\pm49.7}$/$63.08_{\scriptscriptstyle\pm54.6}$$^{(3)}$ & -\\
H-nobs & $94.90_{\scriptscriptstyle\pm1.3}$/$90.72_{\scriptscriptstyle\pm2.1}$ & $14.82_{\scriptscriptstyle\pm4.0}$/$1.91_{\scriptscriptstyle\pm2.1}$$^{(3)}$ & $91.09_{\scriptscriptstyle\pm1.2}$/$84.66_{\scriptscriptstyle\pm2.3}$ & $94.90_{\scriptscriptstyle\pm1.3}$/$90.72_{\scriptscriptstyle\pm2.1}$ & $93.45_{\scriptscriptstyle\pm1.0}$/$88.26_{\scriptscriptstyle\pm1.3}$ & $94.80_{\scriptscriptstyle\pm1.4}$/$90.59_{\scriptscriptstyle\pm2.0}$ & $65.54_{\scriptscriptstyle\pm44.3}$/$56.83_{\scriptscriptstyle\pm49.0}$$^{(3)}$ & $94.90_{\scriptscriptstyle\pm1.3}$/$90.72_{\scriptscriptstyle\pm2.1}$ & $14.82_{\scriptscriptstyle\pm4.0}$/$1.91_{\scriptscriptstyle\pm2.1}$$^{(3)}$\\
Trimmed Mean & $94.90_{\scriptscriptstyle\pm1.3}$/$90.71_{\scriptscriptstyle\pm2.1}$ & $85.77_{\scriptscriptstyle\pm4.4}$/$75.80_{\scriptscriptstyle\pm7.0}$ & $87.30_{\scriptscriptstyle\pm3.6}$/$78.31_{\scriptscriptstyle\pm6.1}$$^{(4)}$ & $94.84_{\scriptscriptstyle\pm1.4}$/$90.46_{\scriptscriptstyle\pm2.3}$ & $65.65_{\scriptscriptstyle\pm48.6}$/$59.07_{\scriptscriptstyle\pm51.2}$$^{(3)}$ & $48.61_{\scriptscriptstyle\pm34.4}$/$37.79_{\scriptscriptstyle\pm32.7}$$^{(3)}$ & $48.74_{\scriptscriptstyle\pm33.7}$/$37.52_{\scriptscriptstyle\pm32.5}$$^{(3)}$ & $95.33_{\scriptscriptstyle\pm1.3}$/$91.29_{\scriptscriptstyle\pm1.9}$ & $48.24_{\scriptscriptstyle\pm34.1}$/$37.25_{\scriptscriptstyle\pm32.3}$$^{(3)}$\\
Median & $66.04_{\scriptscriptstyle\pm39.1}$/$55.57_{\scriptscriptstyle\pm42.6}$$^{(4)}$ & $61.37_{\scriptscriptstyle\pm44.3}$/$51.79_{\scriptscriptstyle\pm45.0}$$^{(3)}$ & $86.33_{\scriptscriptstyle\pm5.0}$/$78.61_{\scriptscriptstyle\pm6.5}$$^{(3)}$ & $66.10_{\scriptscriptstyle\pm48.8}$/$59.72_{\scriptscriptstyle\pm51.7}$$^{(3)}$ & $67.35_{\scriptscriptstyle\pm41.2}$/$58.41_{\scriptscriptstyle\pm47.1}$$^{(3)}$ & $47.54_{\scriptscriptstyle\pm32.9}$/$36.27_{\scriptscriptstyle\pm31.4}$$^{(3)}$ & $47.25_{\scriptscriptstyle\pm32.3}$/$35.64_{\scriptscriptstyle\pm30.9}$$^{(3)}$ & $73.93_{\scriptscriptstyle\pm42.4}$/$68.19_{\scriptscriptstyle\pm45.5}$$^{(4)}$ & $65.88_{\scriptscriptstyle\pm0.4}$/$53.38_{\scriptscriptstyle\pm0.6}$$^{(2)}$\\
Krum & $93.99_{\scriptscriptstyle\pm1.4}$/$89.24_{\scriptscriptstyle\pm2.5}$ & $90.45_{\scriptscriptstyle\pm1.5}$/$83.30_{\scriptscriptstyle\pm2.7}$ & $93.42_{\scriptscriptstyle\pm1.3}$/$88.42_{\scriptscriptstyle\pm2.3}$ & $93.99_{\scriptscriptstyle\pm1.4}$/$89.24_{\scriptscriptstyle\pm2.5}$ & $63.99_{\scriptscriptstyle\pm46.7}$/$56.77_{\scriptscriptstyle\pm49.2}$$^{(3)}$ & $94.00_{\scriptscriptstyle\pm1.5}$/$89.24_{\scriptscriptstyle\pm2.6}$ & $93.76_{\scriptscriptstyle\pm1.2}$/$88.61_{\scriptscriptstyle\pm1.7}$ & $93.99_{\scriptscriptstyle\pm1.4}$/$89.24_{\scriptscriptstyle\pm2.5}$ & $63.99_{\scriptscriptstyle\pm46.7}$/$56.14_{\scriptscriptstyle\pm48.6}$$^{(3)}$\\
Multi-Krum & $94.72_{\scriptscriptstyle\pm1.4}$/$90.39_{\scriptscriptstyle\pm2.3}$ & $59.90_{\scriptscriptstyle\pm43.4}$/$46.41_{\scriptscriptstyle\pm40.2}$$^{(3)}$ & $90.27_{\scriptscriptstyle\pm0.5}$/$83.57_{\scriptscriptstyle\pm1.1}$ & $94.72_{\scriptscriptstyle\pm1.4}$/$90.39_{\scriptscriptstyle\pm2.3}$ & $66.22_{\scriptscriptstyle\pm48.9}$/$59.88_{\scriptscriptstyle\pm51.9}$$^{(3)}$ & $94.16_{\scriptscriptstyle\pm1.0}$/$89.52_{\scriptscriptstyle\pm1.4}$ & $61.54_{\scriptscriptstyle\pm44.4}$/$54.29_{\scriptscriptstyle\pm47.1}$$^{(3)}$ & $94.72_{\scriptscriptstyle\pm1.4}$/$90.39_{\scriptscriptstyle\pm2.3}$ & $59.90_{\scriptscriptstyle\pm43.4}$/$46.41_{\scriptscriptstyle\pm40.2}$$^{(3)}$\\
FLTrust & $94.96_{\scriptscriptstyle\pm0.6}$/$90.53_{\scriptscriptstyle\pm0.8}$ & $93.30_{\scriptscriptstyle\pm0.8}$/$88.05_{\scriptscriptstyle\pm1.7}$ & $93.34_{\scriptscriptstyle\pm0.8}$/$88.24_{\scriptscriptstyle\pm1.6}$ & $94.98_{\scriptscriptstyle\pm0.5}$/$90.54_{\scriptscriptstyle\pm0.8}$ & $93.10_{\scriptscriptstyle\pm3.4}$/$87.16_{\scriptscriptstyle\pm5.7}$ & $89.97_{\scriptscriptstyle\pm3.9}$/$82.26_{\scriptscriptstyle\pm6.4}$ & $91.77_{\scriptscriptstyle\pm2.3}$/$85.34_{\scriptscriptstyle\pm3.4}$ & $95.07_{\scriptscriptstyle\pm0.6}$/$90.76_{\scriptscriptstyle\pm0.8}$ & $88.52_{\scriptscriptstyle\pm3.3}$/$79.64_{\scriptscriptstyle\pm5.2}$\\
BOBA (cls.) & $96.41_{\scriptscriptstyle\pm0.1}$/$92.68_{\scriptscriptstyle\pm0.4}$ & $94.45_{\scriptscriptstyle\pm0.2}$/$89.83_{\scriptscriptstyle\pm0.3}$ & $11.23_{\scriptscriptstyle\pm0.5}$/$0.16_{\scriptscriptstyle\pm0.4}$ & $96.41_{\scriptscriptstyle\pm0.1}$/$92.68_{\scriptscriptstyle\pm0.4}$ & $95.70_{\scriptscriptstyle\pm0.1}$/$91.40_{\scriptscriptstyle\pm0.5}$ & $80.55_{\scriptscriptstyle\pm3.8}$/$70.92_{\scriptscriptstyle\pm5.8}$ & $82.26_{\scriptscriptstyle\pm5.4}$/$72.58_{\scriptscriptstyle\pm7.2}$ & $96.60_{\scriptscriptstyle\pm0.1}$/$93.00_{\scriptscriptstyle\pm0.2}$ & $11.23_{\scriptscriptstyle\pm0.5}$/$0.16_{\scriptscriptstyle\pm0.4}$\\
FedAA & $96.42_{\scriptscriptstyle\pm0.2}$/$92.71_{\scriptscriptstyle\pm0.4}$ & $51.53_{\scriptscriptstyle\pm40.3}$/$39.47_{\scriptscriptstyle\pm37.7}$ & $85.86_{\scriptscriptstyle\pm10.7}$/$77.56_{\scriptscriptstyle\pm14.4}$ & $96.29_{\scriptscriptstyle\pm0.3}$/$92.31_{\scriptscriptstyle\pm0.9}$ & $96.30_{\scriptscriptstyle\pm0.1}$/$92.39_{\scriptscriptstyle\pm0.4}$ & $74.04_{\scriptscriptstyle\pm10.2}$/$62.88_{\scriptscriptstyle\pm13.1}$ & $60.57_{\scriptscriptstyle\pm18.8}$/$47.28_{\scriptscriptstyle\pm20.0}$ & $96.22_{\scriptscriptstyle\pm0.1}$/$92.41_{\scriptscriptstyle\pm0.2}$ & $43.00_{\scriptscriptstyle\pm32.2}$/$31.26_{\scriptscriptstyle\pm28.8}$\\
Ditto & $96.41_{\scriptscriptstyle\pm0.1}$/$92.72_{\scriptscriptstyle\pm0.6}$ & $94.53_{\scriptscriptstyle\pm0.1}$/$89.05_{\scriptscriptstyle\pm1.2}$ & - & $87.36_{\scriptscriptstyle\pm0.2}$/$78.18_{\scriptscriptstyle\pm3.4}$ & $95.95_{\scriptscriptstyle\pm0.1}$/$91.73_{\scriptscriptstyle\pm1.0}$ & $81.79_{\scriptscriptstyle\pm1.4}$/$63.73_{\scriptscriptstyle\pm3.9}$ & $76.89_{\scriptscriptstyle\pm4.9}$/$61.49_{\scriptscriptstyle\pm1.4}$ & $97.33_{\scriptscriptstyle\pm0.2}$/$94.41_{\scriptscriptstyle\pm0.8}$ & -\\
\textbf{\methodname{}} & $93.83_{\scriptscriptstyle\pm0.1}$/$89.16_{\scriptscriptstyle\pm0.2}$ & $93.15_{\scriptscriptstyle\pm0.1}$/$88.30_{\scriptscriptstyle\pm0.4}$ & $92.42_{\scriptscriptstyle\pm0.3}$/$87.07_{\scriptscriptstyle\pm0.2}$ & $93.83_{\scriptscriptstyle\pm0.1}$/$89.16_{\scriptscriptstyle\pm0.2}$ & $93.81_{\scriptscriptstyle\pm0.1}$/$89.17_{\scriptscriptstyle\pm0.1}$ & $93.84_{\scriptscriptstyle\pm0.1}$/$89.14_{\scriptscriptstyle\pm0.2}$ & $93.84_{\scriptscriptstyle\pm0.1}$/$89.15_{\scriptscriptstyle\pm0.2}$ & $93.83_{\scriptscriptstyle\pm0.1}$/$89.19_{\scriptscriptstyle\pm0.2}$ & $92.42_{\scriptscriptstyle\pm0.3}$/$87.07_{\scriptscriptstyle\pm0.2}$\\
\bottomrule
\end{tabular}}
\end{table}

\begin{table}[!t]
\centering\setlength{\tabcolsep}{1.0pt}\renewcommand{\arraystretch}{0.9}\scriptsize
\caption{EMNIST-Balanced / CNN, rounds 990--999. Entries are global / worst-10\% client accuracy (\%), mean over seeds 40, 41, 42, 43, 44 with the across-seed standard deviation as subscript; the worst-attack column is the seed-wise minimum over the seven attacks, also mean with std subscript, over seeds completing all seven attacks.}
\label{tab:emnist-meanstd-5seed}
\resizebox{\textwidth}{!}{\begin{tabular}{lccccccccc}
\toprule
Method & Clean & Sign & IPM$_{\lambda=3}$ & Gauss. & ALIE & Min-Max & Min-Sum & Magnitude & Worst attack\\
\midrule
q-FFL-GD (mean) & $63.79_{\scriptscriptstyle\pm34.5}$/$57.08_{\scriptscriptstyle\pm32.0}$ & $74.93_{\scriptscriptstyle\pm2.5}$/$66.23_{\scriptscriptstyle\pm3.0}$ & - & $2.13_{\scriptscriptstyle\pm0.0}$/$0.00_{\scriptscriptstyle\pm0.0}$$^{(1)}$ & $53.66_{\scriptscriptstyle\pm44.6}$/$46.71_{\scriptscriptstyle\pm40.4}$$^{(3)}$ & $44.08_{\scriptscriptstyle\pm36.5}$/$36.43_{\scriptscriptstyle\pm31.8}$$^{(3)}$ & $27.01_{\scriptscriptstyle\pm38.2}$/$21.24_{\scriptscriptstyle\pm34.3}$$^{(3)}$ & $65.81_{\scriptscriptstyle\pm35.6}$/$58.92_{\scriptscriptstyle\pm33.0}$ & -\\
H-nobs & $63.79_{\scriptscriptstyle\pm34.5}$/$57.08_{\scriptscriptstyle\pm32.0}$ & $28.70_{\scriptscriptstyle\pm37.6}$/$21.64_{\scriptscriptstyle\pm30.6}$$^{(2)}$ & $78.66_{\scriptscriptstyle\pm1.5}$/$70.66_{\scriptscriptstyle\pm1.5}$ & $63.79_{\scriptscriptstyle\pm34.5}$/$57.08_{\scriptscriptstyle\pm32.0}$ & $78.55_{\scriptscriptstyle\pm0.6}$/$69.11_{\scriptscriptstyle\pm0.4}$$^{(3)}$ & $52.75_{\scriptscriptstyle\pm43.8}$/$45.63_{\scriptscriptstyle\pm39.5}$$^{(3)}$ & $50.65_{\scriptscriptstyle\pm42.0}$/$43.19_{\scriptscriptstyle\pm37.4}$$^{(3)}$ & $63.79_{\scriptscriptstyle\pm34.5}$/$57.08_{\scriptscriptstyle\pm32.0}$ & $28.70_{\scriptscriptstyle\pm37.6}$/$21.64_{\scriptscriptstyle\pm30.6}$$^{(2)}$\\
Trimmed Mean & $80.30_{\scriptscriptstyle\pm0.5}$/$72.43_{\scriptscriptstyle\pm0.6}$ & $63.17_{\scriptscriptstyle\pm1.7}$/$52.69_{\scriptscriptstyle\pm2.0}$ & $54.93_{\scriptscriptstyle\pm4.5}$/$44.44_{\scriptscriptstyle\pm3.9}$$^{(4)}$ & $63.28_{\scriptscriptstyle\pm34.3}$/$56.28_{\scriptscriptstyle\pm31.6}$ & $66.82_{\scriptscriptstyle\pm18.7}$/$57.29_{\scriptscriptstyle\pm18.9}$$^{(4)}$ & $25.45_{\scriptscriptstyle\pm20.2}$/$17.89_{\scriptscriptstyle\pm15.6}$$^{(3)}$ & $34.35_{\scriptscriptstyle\pm29.2}$/$26.51_{\scriptscriptstyle\pm24.6}$$^{(3)}$ & $64.13_{\scriptscriptstyle\pm34.7}$/$56.61_{\scriptscriptstyle\pm31.7}$ & $37.11_{\scriptscriptstyle\pm1.8}$/$26.83_{\scriptscriptstyle\pm2.4}$$^{(2)}$\\
Median & $71.43_{\scriptscriptstyle\pm10.3}$/$60.98_{\scriptscriptstyle\pm11.1}$$^{(4)}$ & $45.35_{\scriptscriptstyle\pm21.3}$/$34.82_{\scriptscriptstyle\pm19.9}$ & $58.80_{\scriptscriptstyle\pm6.6}$/$47.41_{\scriptscriptstyle\pm7.3}$ & $59.14_{\scriptscriptstyle\pm30.2}$/$49.74_{\scriptscriptstyle\pm29.3}$$^{(3)}$ & $65.63_{\scriptscriptstyle\pm9.7}$/$56.19_{\scriptscriptstyle\pm9.3}$$^{(4)}$ & $17.57_{\scriptscriptstyle\pm26.8}$/$12.35_{\scriptscriptstyle\pm21.4}$$^{(3)}$ & $30.96_{\scriptscriptstyle\pm25.1}$/$22.84_{\scriptscriptstyle\pm20.0}$$^{(3)}$ & $78.51_{\scriptscriptstyle\pm0.5}$/$68.41_{\scriptscriptstyle\pm0.1}$$^{(3)}$ & $15.72_{\scriptscriptstyle\pm23.6}$/$10.51_{\scriptscriptstyle\pm18.2}$$^{(3)}$\\
Krum & $78.49_{\scriptscriptstyle\pm0.8}$/$70.12_{\scriptscriptstyle\pm0.7}$ & $7.35_{\scriptscriptstyle\pm2.5}$/$1.24_{\scriptscriptstyle\pm1.0}$$^{(2)}$ & $78.64_{\scriptscriptstyle\pm0.7}$/$70.15_{\scriptscriptstyle\pm0.7}$ & $78.49_{\scriptscriptstyle\pm0.8}$/$70.12_{\scriptscriptstyle\pm0.7}$ & $71.34_{\scriptscriptstyle\pm4.5}$/$60.74_{\scriptscriptstyle\pm4.7}$$^{(3)}$ & $47.92_{\scriptscriptstyle\pm39.7}$/$39.93_{\scriptscriptstyle\pm34.6}$$^{(3)}$ & $4.12_{\scriptscriptstyle\pm3.5}$/$1.02_{\scriptscriptstyle\pm1.8}$$^{(3)}$ & $78.49_{\scriptscriptstyle\pm0.8}$/$70.12_{\scriptscriptstyle\pm0.7}$ & $5.12_{\scriptscriptstyle\pm4.2}$/$0.97_{\scriptscriptstyle\pm1.4}$$^{(2)}$\\
Multi-Krum & $80.49_{\scriptscriptstyle\pm0.5}$/$72.46_{\scriptscriptstyle\pm0.6}$ & $62.69_{\scriptscriptstyle\pm2.9}$/$52.43_{\scriptscriptstyle\pm3.1}$$^{(3)}$ & $78.71_{\scriptscriptstyle\pm1.3}$/$70.46_{\scriptscriptstyle\pm1.2}$$^{(4)}$ & $80.49_{\scriptscriptstyle\pm0.5}$/$72.46_{\scriptscriptstyle\pm0.6}$ & $60.16_{\scriptscriptstyle\pm22.1}$/$49.78_{\scriptscriptstyle\pm23.1}$$^{(4)}$ & $38.19_{\scriptscriptstyle\pm31.3}$/$30.24_{\scriptscriptstyle\pm26.3}$$^{(3)}$ & $38.14_{\scriptscriptstyle\pm31.2}$/$30.05_{\scriptscriptstyle\pm26.1}$$^{(3)}$ & $80.49_{\scriptscriptstyle\pm0.5}$/$72.46_{\scriptscriptstyle\pm0.6}$ & 38.14/30.05\\
FLTrust & $71.34_{\scriptscriptstyle\pm2.7}$/$62.53_{\scriptscriptstyle\pm3.5}$ & $71.02_{\scriptscriptstyle\pm2.8}$/$62.29_{\scriptscriptstyle\pm3.7}$ & $71.46_{\scriptscriptstyle\pm2.6}$/$62.52_{\scriptscriptstyle\pm3.6}$ & $72.17_{\scriptscriptstyle\pm1.3}$/$63.44_{\scriptscriptstyle\pm2.1}$ & $75.48_{\scriptscriptstyle\pm0.7}$/$65.71_{\scriptscriptstyle\pm1.0}$ & $63.45_{\scriptscriptstyle\pm4.4}$/$52.76_{\scriptscriptstyle\pm5.1}$ & $63.88_{\scriptscriptstyle\pm5.2}$/$53.36_{\scriptscriptstyle\pm5.8}$ & $68.45_{\scriptscriptstyle\pm7.3}$/$59.25_{\scriptscriptstyle\pm8.1}$ & $61.21_{\scriptscriptstyle\pm5.4}$/$50.22_{\scriptscriptstyle\pm5.9}$\\
BOBA (cls.) & $80.34_{\scriptscriptstyle\pm0.5}$/$72.19_{\scriptscriptstyle\pm1.0}$ & $73.19_{\scriptscriptstyle\pm1.1}$/$64.32_{\scriptscriptstyle\pm1.3}$ & - & $80.45_{\scriptscriptstyle\pm0.5}$/$72.31_{\scriptscriptstyle\pm0.9}$ & $78.91_{\scriptscriptstyle\pm0.5}$/$69.21_{\scriptscriptstyle\pm0.5}$ & $72.64_{\scriptscriptstyle\pm7.9}$/$63.86_{\scriptscriptstyle\pm9.1}$ & $75.60_{\scriptscriptstyle\pm1.7}$/$66.98_{\scriptscriptstyle\pm2.4}$ & $79.59_{\scriptscriptstyle\pm0.6}$/$70.85_{\scriptscriptstyle\pm1.3}$ & -\\
FedAA & $78.15_{\scriptscriptstyle\pm0.6}$/$68.99_{\scriptscriptstyle\pm0.6}$ & $47.32_{\scriptscriptstyle\pm28.0}$/$38.31_{\scriptscriptstyle\pm25.1}$$^{(4)}$ & $71.70_{\scriptscriptstyle\pm5.9}$/$61.89_{\scriptscriptstyle\pm8.0}$$^{(2)}$ & $78.26_{\scriptscriptstyle\pm0.6}$/$69.34_{\scriptscriptstyle\pm0.4}$ & $71.43_{\scriptscriptstyle\pm4.0}$/$60.91_{\scriptscriptstyle\pm4.8}$ & $75.35_{\scriptscriptstyle\pm4.4}$/$66.42_{\scriptscriptstyle\pm5.1}$ & $60.20_{\scriptscriptstyle\pm31.9}$/$52.65_{\scriptscriptstyle\pm29.3}$ & $78.39_{\scriptscriptstyle\pm0.6}$/$69.18_{\scriptscriptstyle\pm1.0}$ & $30.43_{\scriptscriptstyle\pm38.4}$/$23.53_{\scriptscriptstyle\pm32.5}$$^{(2)}$\\
Ditto & $80.47_{\scriptscriptstyle\pm0.4}$/$65.80_{\scriptscriptstyle\pm1.5}$ & $72.16_{\scriptscriptstyle\pm1.2}$/$42.02_{\scriptscriptstyle\pm2.5}$ & - & $2.13_{\scriptscriptstyle\pm0.0}$/$0.00_{\scriptscriptstyle\pm0.0}$$^{(1)}$ & $78.07_{\scriptscriptstyle\pm0.8}$/$61.66_{\scriptscriptstyle\pm0.7}$ & $61.34_{\scriptscriptstyle\pm1.4}$/$53.18_{\scriptscriptstyle\pm2.6}$ & $62.93_{\scriptscriptstyle\pm0.8}$/$54.56_{\scriptscriptstyle\pm1.0}$ & $82.58_{\scriptscriptstyle\pm0.5}$/$74.83_{\scriptscriptstyle\pm0.5}$ & -\\
\textbf{\methodname{}} & $79.12_{\scriptscriptstyle\pm2.1}$/$71.04_{\scriptscriptstyle\pm2.1}$ & $79.51_{\scriptscriptstyle\pm0.8}$/$71.52_{\scriptscriptstyle\pm1.2}$ & $79.41_{\scriptscriptstyle\pm0.5}$/$71.30_{\scriptscriptstyle\pm0.9}$ & $79.67_{\scriptscriptstyle\pm1.0}$/$71.63_{\scriptscriptstyle\pm1.1}$ & $79.61_{\scriptscriptstyle\pm1.0}$/$71.67_{\scriptscriptstyle\pm1.2}$ & $78.89_{\scriptscriptstyle\pm2.6}$/$70.77_{\scriptscriptstyle\pm2.8}$ & $79.11_{\scriptscriptstyle\pm2.1}$/$71.05_{\scriptscriptstyle\pm2.1}$ & $79.12_{\scriptscriptstyle\pm2.1}$/$71.04_{\scriptscriptstyle\pm2.1}$ & $78.31_{\scriptscriptstyle\pm2.3}$/$70.00_{\scriptscriptstyle\pm2.4}$\\
\bottomrule
\end{tabular}}
\end{table}

\begin{table}[!t]
\centering\setlength{\tabcolsep}{1.0pt}\renewcommand{\arraystretch}{0.9}\scriptsize
\caption{AGNews / GRU, rounds 1990--1999. Entries are global / worst-10\% client accuracy (\%), mean over seeds 40, 41, 42, 43, 44 with the across-seed standard deviation as subscript; the worst-attack column is the seed-wise minimum over the seven attacks, also mean with std subscript, over seeds completing all seven attacks.}
\label{tab:agnews-meanstd-5seed}
\resizebox{\textwidth}{!}{\begin{tabular}{lccccccccc}
\toprule
Method & Clean & Sign & IPM$_{\lambda=3}$ & Gauss. & ALIE & Min-Max & Min-Sum & Magnitude & Worst attack\\
\midrule
q-FFL-GD (mean) & $81.52_{\scriptscriptstyle\pm3.0}$/$68.95_{\scriptscriptstyle\pm2.7}$ & $64.00_{\scriptscriptstyle\pm2.0}$/$49.40_{\scriptscriptstyle\pm4.7}$ & - & $27.06_{\scriptscriptstyle\pm0.7}$/$16.92_{\scriptscriptstyle\pm1.0}$$^{(4)}$ & $73.23_{\scriptscriptstyle\pm9.0}$/$59.00_{\scriptscriptstyle\pm8.9}$ & $40.49_{\scriptscriptstyle\pm15.2}$/$23.90_{\scriptscriptstyle\pm11.6}$ & $45.40_{\scriptscriptstyle\pm18.0}$/$29.31_{\scriptscriptstyle\pm16.8}$$^{(4)}$ & $76.35_{\scriptscriptstyle\pm22.8}$/$64.38_{\scriptscriptstyle\pm26.1}$ & -\\
H-nobs & $81.52_{\scriptscriptstyle\pm3.0}$/$68.95_{\scriptscriptstyle\pm2.7}$ & $26.69_{\scriptscriptstyle\pm1.4}$/$0.56_{\scriptscriptstyle\pm0.9}$$^{(3)}$ & $81.62_{\scriptscriptstyle\pm2.6}$/$67.99_{\scriptscriptstyle\pm3.4}$ & $81.52_{\scriptscriptstyle\pm3.0}$/$68.95_{\scriptscriptstyle\pm2.7}$ & $77.53_{\scriptscriptstyle\pm0.5}$/$62.83_{\scriptscriptstyle\pm2.2}$ & $72.91_{\scriptscriptstyle\pm13.2}$/$58.16_{\scriptscriptstyle\pm13.7}$ & $73.71_{\scriptscriptstyle\pm1.5}$/$58.29_{\scriptscriptstyle\pm2.0}$ & $81.52_{\scriptscriptstyle\pm3.0}$/$68.95_{\scriptscriptstyle\pm2.7}$ & $26.69_{\scriptscriptstyle\pm1.4}$/$0.56_{\scriptscriptstyle\pm0.9}$$^{(3)}$\\
Trimmed Mean & $81.57_{\scriptscriptstyle\pm2.9}$/$69.06_{\scriptscriptstyle\pm2.5}$ & $36.61_{\scriptscriptstyle\pm2.9}$/$8.79_{\scriptscriptstyle\pm6.4}$ & $35.88_{\scriptscriptstyle\pm2.7}$/$6.90_{\scriptscriptstyle\pm6.6}$ & $79.67_{\scriptscriptstyle\pm5.9}$/$66.55_{\scriptscriptstyle\pm7.0}$ & $74.76_{\scriptscriptstyle\pm0.9}$/$59.20_{\scriptscriptstyle\pm2.3}$ & $38.63_{\scriptscriptstyle\pm13.4}$/$19.84_{\scriptscriptstyle\pm9.6}$ & $39.37_{\scriptscriptstyle\pm13.5}$/$22.11_{\scriptscriptstyle\pm8.3}$ & $79.89_{\scriptscriptstyle\pm3.0}$/$67.29_{\scriptscriptstyle\pm2.2}$ & $30.85_{\scriptscriptstyle\pm2.9}$/$5.71_{\scriptscriptstyle\pm4.7}$\\
Median & $66.83_{\scriptscriptstyle\pm4.0}$/$46.50_{\scriptscriptstyle\pm7.0}$ & $40.53_{\scriptscriptstyle\pm3.8}$/$15.32_{\scriptscriptstyle\pm5.6}$ & $37.58_{\scriptscriptstyle\pm4.7}$/$12.00_{\scriptscriptstyle\pm9.2}$ & $72.53_{\scriptscriptstyle\pm1.6}$/$54.11_{\scriptscriptstyle\pm3.7}$ & $70.64_{\scriptscriptstyle\pm1.3}$/$55.17_{\scriptscriptstyle\pm1.7}$ & $49.33_{\scriptscriptstyle\pm11.0}$/$30.59_{\scriptscriptstyle\pm9.1}$ & $49.01_{\scriptscriptstyle\pm10.7}$/$29.42_{\scriptscriptstyle\pm9.9}$ & $64.82_{\scriptscriptstyle\pm18.3}$/$48.88_{\scriptscriptstyle\pm17.5}$ & $33.50_{\scriptscriptstyle\pm3.8}$/$9.20_{\scriptscriptstyle\pm4.8}$\\
Krum & $72.69_{\scriptscriptstyle\pm10.1}$/$56.90_{\scriptscriptstyle\pm10.1}$ & $77.17_{\scriptscriptstyle\pm2.6}$/$60.86_{\scriptscriptstyle\pm5.0}$ & $76.62_{\scriptscriptstyle\pm6.7}$/$58.96_{\scriptscriptstyle\pm10.6}$ & $72.69_{\scriptscriptstyle\pm10.1}$/$56.90_{\scriptscriptstyle\pm10.1}$ & $72.18_{\scriptscriptstyle\pm1.6}$/$56.64_{\scriptscriptstyle\pm0.7}$ & $77.42_{\scriptscriptstyle\pm5.2}$/$62.50_{\scriptscriptstyle\pm6.7}$ & $74.77_{\scriptscriptstyle\pm13.3}$/$59.40_{\scriptscriptstyle\pm16.2}$ & $72.69_{\scriptscriptstyle\pm10.1}$/$56.90_{\scriptscriptstyle\pm10.1}$ & $64.43_{\scriptscriptstyle\pm9.9}$/$48.20_{\scriptscriptstyle\pm12.1}$\\
Multi-Krum & $76.02_{\scriptscriptstyle\pm9.9}$/$62.24_{\scriptscriptstyle\pm10.6}$ & $66.96_{\scriptscriptstyle\pm11.5}$/$46.28_{\scriptscriptstyle\pm12.4}$ & $73.08_{\scriptscriptstyle\pm13.2}$/$56.59_{\scriptscriptstyle\pm17.9}$ & $76.02_{\scriptscriptstyle\pm9.9}$/$62.24_{\scriptscriptstyle\pm10.6}$ & $74.87_{\scriptscriptstyle\pm1.1}$/$57.84_{\scriptscriptstyle\pm2.0}$ & $78.23_{\scriptscriptstyle\pm4.6}$/$64.39_{\scriptscriptstyle\pm3.1}$ & $66.27_{\scriptscriptstyle\pm5.5}$/$49.22_{\scriptscriptstyle\pm5.5}$ & $76.02_{\scriptscriptstyle\pm9.9}$/$62.24_{\scriptscriptstyle\pm10.6}$ & $58.34_{\scriptscriptstyle\pm5.5}$/$38.31_{\scriptscriptstyle\pm8.0}$\\
FLTrust & $83.92_{\scriptscriptstyle\pm1.1}$/$70.63_{\scriptscriptstyle\pm2.6}$ & $83.92_{\scriptscriptstyle\pm1.1}$/$70.63_{\scriptscriptstyle\pm2.6}$ & $83.92_{\scriptscriptstyle\pm1.1}$/$70.63_{\scriptscriptstyle\pm2.6}$ & $83.98_{\scriptscriptstyle\pm1.0}$/$70.79_{\scriptscriptstyle\pm2.4}$ & $77.19_{\scriptscriptstyle\pm1.0}$/$62.57_{\scriptscriptstyle\pm2.5}$ & $66.82_{\scriptscriptstyle\pm3.4}$/$49.24_{\scriptscriptstyle\pm4.8}$ & $64.26_{\scriptscriptstyle\pm3.6}$/$45.46_{\scriptscriptstyle\pm4.8}$ & $83.03_{\scriptscriptstyle\pm0.7}$/$69.91_{\scriptscriptstyle\pm1.8}$ & $64.26_{\scriptscriptstyle\pm3.6}$/$45.46_{\scriptscriptstyle\pm4.8}$\\
BOBA (cls.) & $83.82_{\scriptscriptstyle\pm1.0}$/$70.71_{\scriptscriptstyle\pm2.1}$ & $60.57_{\scriptscriptstyle\pm0.8}$/$44.07_{\scriptscriptstyle\pm2.9}$ & $27.65_{\scriptscriptstyle\pm1.4}$/$1.65_{\scriptscriptstyle\pm1.7}$ & $83.82_{\scriptscriptstyle\pm1.0}$/$70.71_{\scriptscriptstyle\pm2.1}$ & $80.82_{\scriptscriptstyle\pm2.0}$/$66.48_{\scriptscriptstyle\pm4.5}$ & $56.44_{\scriptscriptstyle\pm1.1}$/$38.29_{\scriptscriptstyle\pm2.5}$ & $59.43_{\scriptscriptstyle\pm1.5}$/$41.98_{\scriptscriptstyle\pm2.4}$ & $83.53_{\scriptscriptstyle\pm1.5}$/$69.52_{\scriptscriptstyle\pm3.3}$ & $27.65_{\scriptscriptstyle\pm1.4}$/$1.65_{\scriptscriptstyle\pm1.7}$\\
FedAA & $78.36_{\scriptscriptstyle\pm1.4}$/$61.25_{\scriptscriptstyle\pm3.1}$ & $38.32_{\scriptscriptstyle\pm12.0}$/$11.62_{\scriptscriptstyle\pm12.9}$ & $57.68_{\scriptscriptstyle\pm22.8}$/$38.52_{\scriptscriptstyle\pm30.1}$$^{(2)}$ & $77.83_{\scriptscriptstyle\pm3.0}$/$61.20_{\scriptscriptstyle\pm6.5}$ & $75.29_{\scriptscriptstyle\pm4.7}$/$60.29_{\scriptscriptstyle\pm6.7}$ & $41.18_{\scriptscriptstyle\pm22.2}$/$16.55_{\scriptscriptstyle\pm26.3}$ & $40.34_{\scriptscriptstyle\pm11.4}$/$15.72_{\scriptscriptstyle\pm13.2}$ & $79.67_{\scriptscriptstyle\pm1.4}$/$64.88_{\scriptscriptstyle\pm3.9}$ & $31.29_{\scriptscriptstyle\pm0.8}$/$4.44_{\scriptscriptstyle\pm0.7}$$^{(2)}$\\
Ditto & $83.82_{\scriptscriptstyle\pm1.0}$/$62.56_{\scriptscriptstyle\pm4.9}$ & $63.17_{\scriptscriptstyle\pm0.9}$/$38.63_{\scriptscriptstyle\pm2.6}$ & - & $25.05_{\scriptscriptstyle\pm0.8}$/$16.74_{\scriptscriptstyle\pm1.8}$$^{(3)}$ & $80.94_{\scriptscriptstyle\pm2.4}$/$62.93_{\scriptscriptstyle\pm3.1}$ & $39.97_{\scriptscriptstyle\pm5.1}$/$29.18_{\scriptscriptstyle\pm4.0}$ & $56.25_{\scriptscriptstyle\pm2.0}$/$39.36_{\scriptscriptstyle\pm3.2}$ & $86.53_{\scriptscriptstyle\pm0.6}$/$77.52_{\scriptscriptstyle\pm2.5}$ & -\\
\textbf{\methodname{}} & $82.29_{\scriptscriptstyle\pm1.3}$/$69.58_{\scriptscriptstyle\pm2.1}$ & $80.46_{\scriptscriptstyle\pm1.0}$/$66.94_{\scriptscriptstyle\pm2.8}$$^{(4)}$ & $81.12_{\scriptscriptstyle\pm1.0}$/$67.92_{\scriptscriptstyle\pm1.8}$ & $82.17_{\scriptscriptstyle\pm1.1}$/$69.43_{\scriptscriptstyle\pm2.5}$ & $78.62_{\scriptscriptstyle\pm4.1}$/$65.58_{\scriptscriptstyle\pm6.4}$ & $81.44_{\scriptscriptstyle\pm1.7}$/$68.25_{\scriptscriptstyle\pm3.1}$ & $81.36_{\scriptscriptstyle\pm1.5}$/$68.22_{\scriptscriptstyle\pm3.0}$ & $82.29_{\scriptscriptstyle\pm1.3}$/$69.58_{\scriptscriptstyle\pm2.1}$ & $77.84_{\scriptscriptstyle\pm4.4}$/$63.75_{\scriptscriptstyle\pm6.6}$$^{(4)}$\\
\bottomrule
\end{tabular}}
\end{table}

\begin{table}[p]
\centering
\small
\setlength{\tabcolsep}{3pt}
\renewcommand{\arraystretch}{0.82}
\setlength{\aboverulesep}{0.30ex}
\setlength{\belowrulesep}{0.30ex}
\caption{\textbf{Worst-10\% client robustness across datasets and attacks.}
Mean worst-10\% client accuracy (W10, \%) over seeds reaching the registered
endpoint. Worst attack is the seed-wise minimum over seven attacks;
$^{(k)}$ denotes $k$ completed seeds.}
\label{tab:w10-main-results}
\makebox[\textwidth][c]{%
\begin{tabular}{@{}llrrrrrrrrr@{}}
\toprule
Dataset/model & Method & Clean & Sign & IPM & Gauss. & ALIE & Min-Max & Min-Sum & Mag. & Worst attack\\
\midrule
\multirow{8}{*}{\shortstack[l]{\textbf{CIFAR-10}\\ResNet-18}} & q-FFL & 67.34 & 66.94 & -- & 4.71$^{(3)}$ & 51.90 & 2.24$^{(4)}$ & 2.04 & 32.27$^{(4)}$ & --\\
 & H-nobs & 67.34 & 0.35$^{(3)}$ & 46.25 & 70.03 & 23.39 & 31.99 & 8.34 & 67.34 & 0.35$^{(3)}$\\
 & Trim. & 60.32 & 46.72 & 31.29 & 70.42 & 25.86 & 1.31 & 2.08 & 64.41 & 1.28\\
 & Krum & 51.97 & 29.35 & 51.13 & 48.53 & 10.14 & \underline{50.09} & \underline{32.93} & 51.66 & 9.78\\
 & FLTrust & 72.22 & \textbf{72.26} & \textbf{72.27} & 72.16 & 42.07 & 35.47 & 30.53 & \underline{70.05} & \underline{30.53}\\
 & BOBA (cls.) & \underline{73.13} & 62.12 & 0.66 & \underline{72.73} & \underline{60.85} & 5.27 & 5.35 & 62.39 & 0.66\\
 & FedAA & 64.87 & 32.48 & 62.88$^{(4)}$ & 61.23 & 49.78 & 10.34 & 30.00 & 53.75 & 10.27$^{(4)}$\\
\rowcolor{scalecolumn} & \textbf{\methodname{}} & \textbf{73.88} & \underline{70.15} & \underline{71.62} & \textbf{73.91} & \textbf{68.35} & \textbf{73.91} & \textbf{73.87} & \textbf{72.67} & \textbf{63.58}\\
\midrule
\multirow{8}{*}{\shortstack[l]{\textbf{MNIST}\\MLP}} & q-FFL & 90.72 & \underline{88.31} & -- & 54.32$^{(3)}$ & 89.66 & 37.24$^{(3)}$ & 46.60$^{(3)}$ & 63.08$^{(3)}$ & --\\
 & H-nobs & 90.72 & 1.91$^{(3)}$ & 84.66 & 90.72 & 88.26 & \textbf{90.59} & 56.83$^{(3)}$ & 90.72 & 1.91$^{(3)}$\\
 & Trim. & 90.71 & 75.80 & 78.31$^{(4)}$ & 90.46 & 59.07$^{(3)}$ & 37.79$^{(3)}$ & 37.52$^{(3)}$ & 91.29 & 37.25$^{(3)}$\\
 & Krum & 89.24 & 83.30 & \textbf{88.42} & 89.24 & 56.77$^{(3)}$ & \underline{89.24} & \underline{88.61} & 89.24 & 56.14$^{(3)}$\\
 & FLTrust & 90.53 & 88.05 & \underline{88.24} & 90.54 & 87.16 & 82.26 & 85.34 & 90.76 & \underline{79.64}\\
 & BOBA (cls.) & \underline{92.68} & \textbf{89.83} & 0.16 & \textbf{92.68} & \underline{91.40} & 70.92 & 72.58 & \textbf{93.00} & 0.16\\
 & FedAA & \textbf{92.71} & 39.47 & 77.56 & \underline{92.31} & \textbf{92.39} & 62.88 & 47.28 & \underline{92.41} & 31.26\\
\rowcolor{scalecolumn} & \textbf{\methodname{}} & 89.16 & \underline{88.30} & 87.07 & 89.16 & 89.17 & 89.14 & \textbf{89.15} & 89.19 & \textbf{87.07}\\
\midrule
\multirow{8}{*}{\shortstack[l]{\textbf{EMNIST-Bal.}\\CNN}} & q-FFL & 57.08 & \underline{66.23} & -- & 0.00$^{(1)}$ & 46.71$^{(3)}$ & 36.43$^{(3)}$ & 21.24$^{(3)}$ & 58.92 & --\\
 & H-nobs & 57.08 & 21.64$^{(2)}$ & \underline{70.66} & 57.08 & 69.11$^{(3)}$ & 45.63$^{(3)}$ & 43.19$^{(3)}$ & 57.08 & 21.64$^{(2)}$\\
 & Trim. & \textbf{72.43} & 52.69 & 44.44$^{(4)}$ & 56.28 & 57.29$^{(4)}$ & 17.89$^{(3)}$ & 26.51$^{(3)}$ & 56.61 & 26.83$^{(2)}$\\
 & Krum & 70.12 & 1.24$^{(2)}$ & 70.15 & 70.12 & 60.74$^{(3)}$ & 39.93$^{(3)}$ & 1.02$^{(3)}$ & 70.12 & 0.97$^{(2)}$\\
 & FLTrust & 62.53 & 62.29 & 62.52 & 63.44 & 65.71 & 52.76 & 53.36 & 59.25 & \underline{50.22}\\
 & BOBA (cls.) & \underline{72.19} & 64.32 & -- & \textbf{72.31} & \underline{69.21} & 63.86 & \underline{66.98} & \underline{70.85} & --\\
 & FedAA & 68.99 & 38.31$^{(4)}$ & 61.89$^{(2)}$ & 69.34 & 60.91 & \underline{66.42} & 52.65 & 69.18 & 23.53$^{(2)}$\\
\rowcolor{scalecolumn} & \textbf{\methodname{}} & 71.04 & \textbf{71.52} & \textbf{71.30} & \underline{71.63} & \textbf{71.67} & \textbf{70.77} & \textbf{71.05} & \textbf{71.04} & \textbf{70.00}\\
\midrule
\multirow{8}{*}{\shortstack[l]{\textbf{AGNews}\\GRU}} & q-FFL & 68.95 & 49.40 & -- & 16.92$^{(4)}$ & 59.00 & 23.90 & 29.31$^{(4)}$ & 64.38 & --\\
 & H-nobs & 68.95 & 0.56$^{(3)}$ & \underline{67.99} & 68.95 & 62.83 & 58.16 & 58.29 & 68.95 & 0.56$^{(3)}$\\
 & Trim. & 69.06 & 8.79 & 6.90 & 66.55 & 59.20 & 19.84 & 22.11 & 67.29 & 5.71\\
 & Krum & 56.90 & 60.86 & 58.96 & 56.90 & 56.64 & \underline{62.50} & \underline{59.40} & 56.90 & \underline{48.20}\\
 & FLTrust & \underline{70.63} & \textbf{70.63} & \textbf{70.63} & \textbf{70.79} & 62.57 & 49.24 & 45.46 & \textbf{69.91} & 45.46\\
 & BOBA (cls.) & \textbf{70.71} & 44.07 & 1.65 & \underline{70.71} & \textbf{66.48} & 38.29 & 41.98 & 69.52 & 1.65\\
 & FedAA & 61.25 & 11.62 & 38.52$^{(2)}$ & 61.20 & 60.29 & 16.55 & 15.72 & 64.88 & 4.44$^{(2)}$\\
\rowcolor{scalecolumn} & \textbf{\methodname{}} & 69.58 & \underline{66.94}$^{(4)}$ & 67.92 & 69.43 & \underline{65.58} & \textbf{68.25} & \textbf{68.22} & \underline{69.58} & \textbf{63.75}$^{(4)}$\\
\bottomrule
\end{tabular}%
}
\end{table}

\begin{table}[!htbp]\centering\scriptsize
\caption{Seed-wise worst-attack accuracy of \methodname{} minus each baseline (global accuracy, percentage points). Worst-attack values are computed per seed as the minimum over the seven attacks; seeds in which either method collapsed under any attack are excluded from that pair ($n$ is the number of paired seeds). CI: 95\% percentile bootstrap over paired seeds; $p$: exact two-sided sign-flip test.}
\label{tab:worst-attack-paired}
\begin{tabular}{llcccc}\toprule
Dataset & Baseline & \shortstack{\methodname{}\\worst attack} &
\shortstack{Baseline\\worst attack} & $\Delta$ [95\% CI] & $p$ ($n$)\\\midrule
CIFAR-10 / ResNet-18 & q-FFL-GD (mean) & 73.88 & - & - & -\\
CIFAR-10 / ResNet-18 & H-nobs & 73.88 & 12.19 & +63.89 [+56.93, +67.83] & 0.250 (3)\\
CIFAR-10 / ResNet-18 & Trimmed Mean & 73.88 & 13.18 & +60.70 [+52.08, +67.62] & 0.062 (5)\\
CIFAR-10 / ResNet-18 & Median & 73.88 & 12.34 & +61.53 [+53.51, +68.25] & 0.062 (5)\\
CIFAR-10 / ResNet-18 & Krum & 73.88 & 21.67 & +52.21 [+43.96, +61.29] & 0.062 (5)\\
CIFAR-10 / ResNet-18 & Multi-Krum & 73.88 & 12.97 & +63.12 [+56.20, +66.92] & 0.250 (3)\\
CIFAR-10 / ResNet-18 & FLTrust & 73.88 & 43.81 & +30.07 [+20.24, +37.83] & 0.062 (5)\\
CIFAR-10 / ResNet-18 & BOBA (cls.) & 73.88 & 10.93 & +62.95 [+55.02, +69.61] & 0.062 (5)\\
CIFAR-10 / ResNet-18 & FedAA & 73.88 & 22.15 & +55.31 [+45.62, +64.99] & 0.125 (4)\\
CIFAR-10 / ResNet-18 & Ditto & 73.88 & - & - & -\\
\midrule
MNIST / MLP & q-FFL-GD (mean) & 92.42 & - & - & -\\
MNIST / MLP & H-nobs & 92.42 & 14.82 & +77.42 [+73.28, +81.00] & 0.250 (3)\\
MNIST / MLP & Trimmed Mean & 92.42 & 48.24 & +44.00 [+24.25, +83.43] & 0.250 (3)\\
MNIST / MLP & Median & 92.42 & 65.88 & +26.31 [+26.27, +26.34] & 0.500 (2)\\
MNIST / MLP & Krum & 92.42 & 63.99 & +28.25 [+0.76, +82.26] & 0.250 (3)\\
MNIST / MLP & Multi-Krum & 92.42 & 59.90 & +32.34 [+6.50, +82.55] & 0.250 (3)\\
MNIST / MLP & FLTrust & 92.42 & 88.52 & +3.90 [+1.56, +6.25] & 0.062 (5)\\
MNIST / MLP & BOBA (cls.) & 92.42 & 11.23 & +81.19 [+80.73, +81.75] & 0.062 (5)\\
MNIST / MLP & FedAA & 92.42 & 43.00 & +49.42 [+24.50, +74.34] & 0.062 (5)\\
MNIST / MLP & Ditto & 92.42 & - & - & -\\
\midrule
EMNIST-Balanced / CNN & q-FFL-GD (mean) & 78.31 & - & - & -\\
EMNIST-Balanced / CNN & H-nobs & 78.31 & 28.70 & +50.66 [+24.77, +76.55] & 0.500 (2)\\
EMNIST-Balanced / CNN & Trimmed Mean & 78.31 & 37.11 & +42.33 [+40.47, +44.19] & 0.500 (2)\\
EMNIST-Balanced / CNN & Median & 78.31 & 15.72 & +63.46 [+37.12, +76.72] & 0.250 (3)\\
EMNIST-Balanced / CNN & Krum & 78.31 & 5.12 & +73.64 [+70.74, +76.55] & 0.500 (2)\\
EMNIST-Balanced / CNN & Multi-Krum & 78.31 & - & - & -\\
EMNIST-Balanced / CNN & FLTrust & 78.31 & 61.21 & +17.09 [+12.33, +21.89] & 0.062 (5)\\
EMNIST-Balanced / CNN & BOBA (cls.) & 78.31 & - & - & -\\
EMNIST-Balanced / CNN & FedAA & 78.31 & 30.43 & +49.01 [+21.26, +76.76] & 0.500 (2)\\
EMNIST-Balanced / CNN & Ditto & 78.31 & - & - & -\\
\midrule
AGNews / GRU & q-FFL-GD (mean) & 77.84 & - & - & -\\
AGNews / GRU & H-nobs & 77.84 & 26.69 & +49.48 [+45.11, +53.86] & 0.500 (2)\\
AGNews / GRU & Trimmed Mean & 77.84 & 30.85 & +47.96 [+41.57, +51.37] & 0.125 (4)\\
AGNews / GRU & Median & 77.84 & 33.50 & +43.97 [+38.72, +49.22] & 0.125 (4)\\
AGNews / GRU & Krum & 77.84 & 64.43 & +15.40 [+4.62, +26.18] & 0.250 (4)\\
AGNews / GRU & Multi-Krum & 77.84 & 58.34 & +21.16 [+14.84, +27.08] & 0.125 (4)\\
AGNews / GRU & FLTrust & 77.84 & 64.26 & +13.03 [+11.84, +14.22] & 0.125 (4)\\
AGNews / GRU & BOBA (cls.) & 77.84 & 27.65 & +49.87 [+45.74, +52.88] & 0.125 (4)\\
AGNews / GRU & FedAA & 77.84 & 31.29 & +49.14 [+49.14, +49.14] & 1.000 (1)\\
AGNews / GRU & Ditto & 77.84 & - & - & -\\
\bottomrule\end{tabular}\end{table}

%% file: appendix/mechanism_validation.tex
\section{Experimental Validation of the Geometric Analysis}
\label{app:geometry-validation}

We validate the two predictions underlying Theorems~\ref{thm:fair_warped_geometry}
and~\ref{thm:scale_cone_separation}: coefficient mass should track the
fairness scale, and scale and cone residual should respond to complementary
deviations. The diagnostic uses MNIST, seed 42, and $q=1$; complete
stage-wise statistics are in Appendix~\ref{app:diagnostics}.

\begin{figure}[t]
\centering
\begin{subfigure}[t]{0.29\textwidth}
\centering
\includegraphics[width=\linewidth]{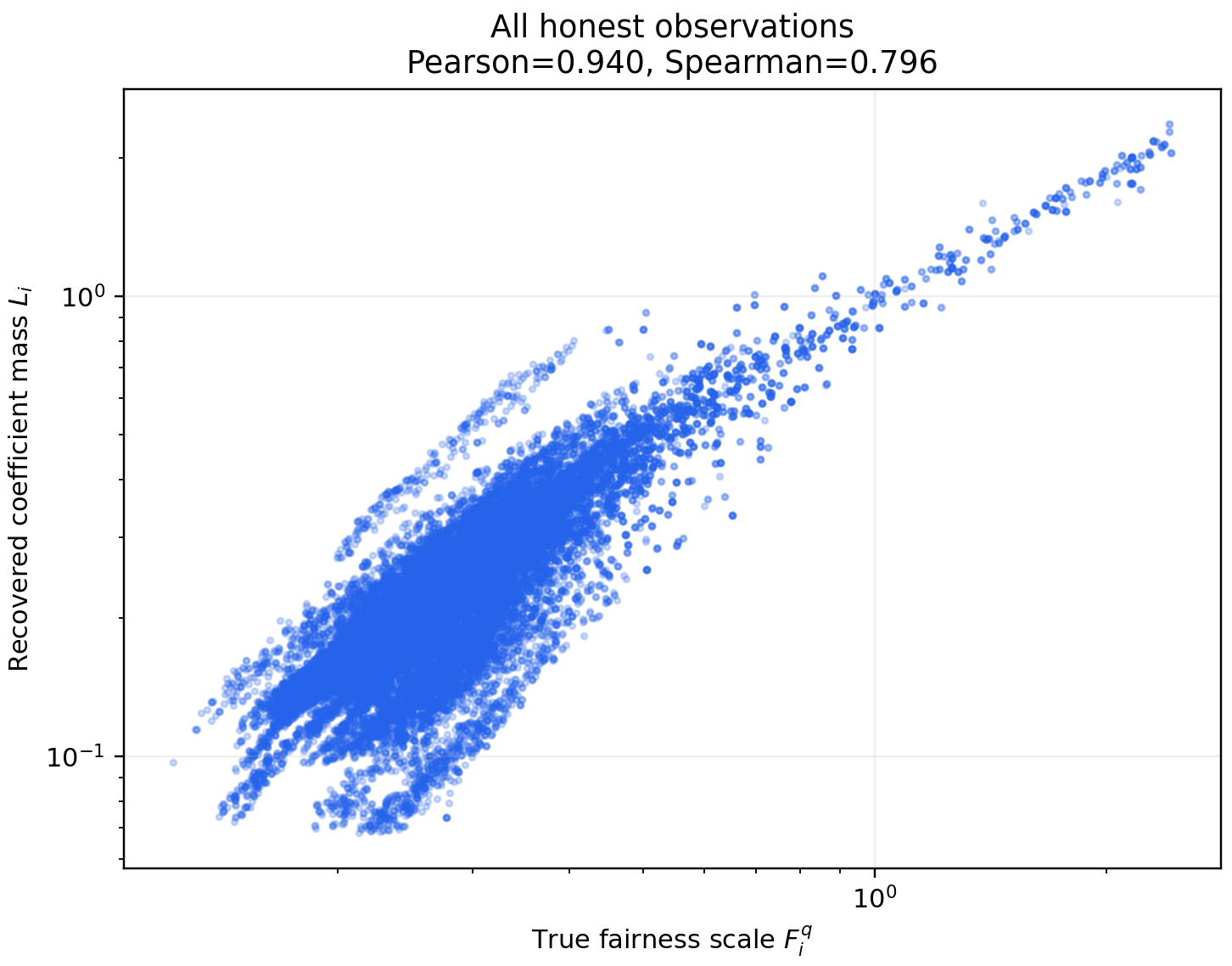}
\caption{$L_i$ versus $F_i^q$.}
\end{subfigure}\hfill
\begin{subfigure}[t]{0.29\textwidth}
\centering
\includegraphics[width=\linewidth]{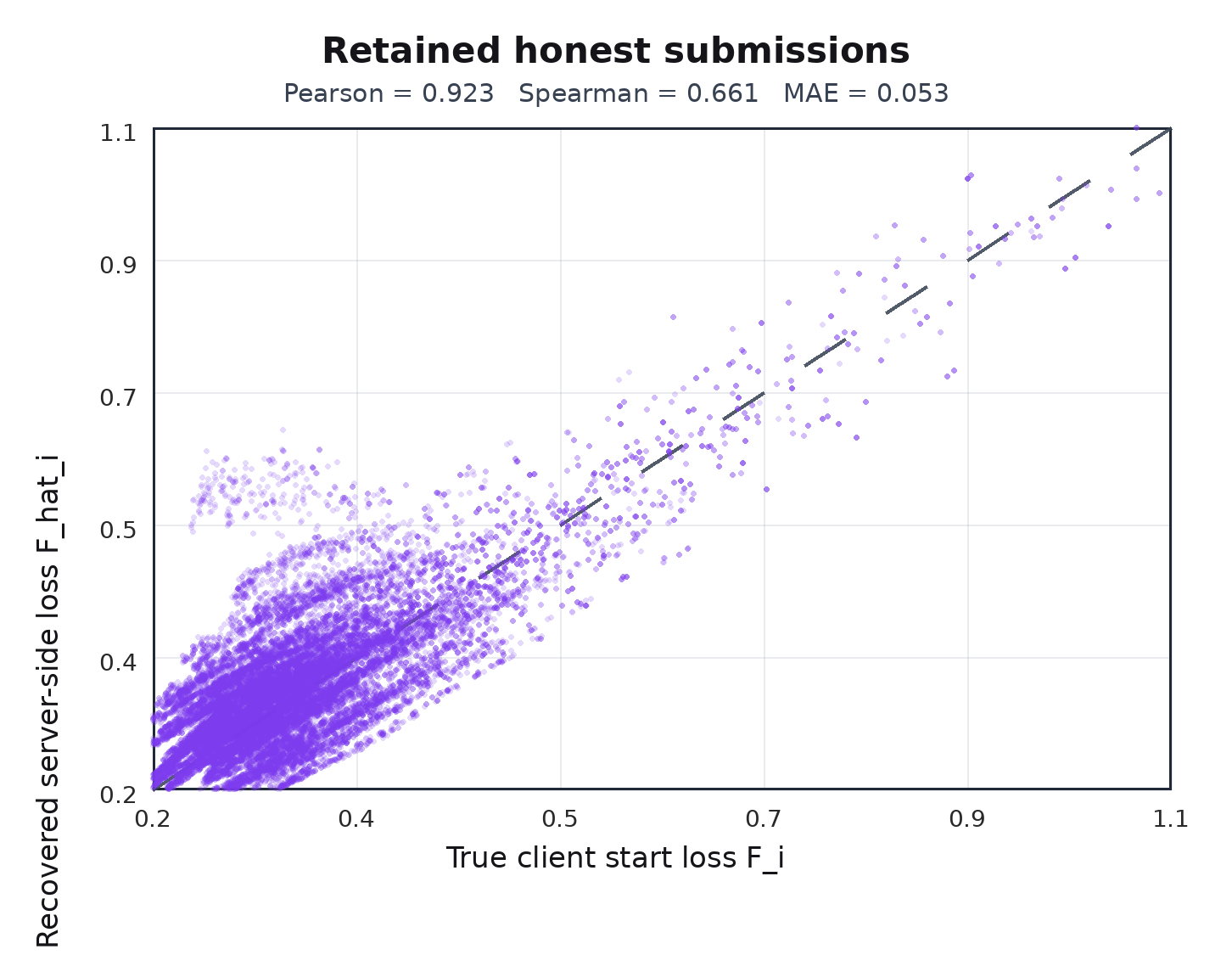}
\caption{$\widetilde F_i$ versus $F_i$.}
\end{subfigure}\hfill
\begin{subfigure}[t]{0.29\textwidth}
\centering
\includegraphics[width=\linewidth]{pic/scale_cone_diagnostic_compact.pdf}
\caption{Scale versus cone score.}
\end{subfigure}
\caption{\textbf{Experimental validation of the geometric analysis.}
Recovered mass tracks the fairness scale (a), normalized coefficients support
loss reconstruction (b), and the two screening statistics separate
complementary attack patterns (c).}
\label{fig:geometry-main}
\end{figure}

The recovered mass has Pearson correlation $r=0.940$ with $F_i^q$; the
trusted loss estimate has Pearson $r=0.923$ and MAE $0.053$ over 76,457
retained honest submissions. Gaussian and Magnitude Scaling primarily produce
large mass, whereas ALIE, Min-Max, and Min-Sum primarily produce large cone
residual. Sign Flip and IPM overlap more strongly with the honest region,
which illustrates the observable boundary of the update-only interface.

Finally, the controlled simplex--cone study in
Appendix~\ref{app:simplex-cone} holds the checkpoint, honest updates, and
trusted basis fixed while changing only the coefficient constraint. It shows
why BOBA's unit-sum simplex is mismatched after fairness scaling: the cone is
invariant to the legitimate radial transformation, whereas a fixed simplex is
not.

%% file: appendix/ablations_diagnostics.tex
\section{Ablations and Diagnostic Evidence}
\label{app:diagnostics}

This appendix groups mechanism attribution, fairness audits, sensitivity controls, trusted-geometry diagnostics, computational cost, and explicit robustness boundaries. Unless otherwise stated, diagnostic statistics use seed 42 and are not presented as cross-seed uncertainty estimates.

\subsection{Three-component factorial ablation}
\label{app:factorial-ablation}

The complete factorial ablation is reported here for all dataset/model pairs.
Tables~\ref{tab:factorial-cifar}--\ref{tab:factorial-agnews} use the identical
1000-round design on every dataset (including AGNews, whose main-table horizon
is 2000 rounds) and report all seven attacks. The three binary
components are the coefficient-scale screen, normalized cone screen, and
loss-normalized aggregation.  Runs that reached round 299 in the original
factorial were continued from their matched checkpoints; configurations that
had already met a registered collapse criterion retain a missing terminal
window.  Because clean training uses $f=0$, its trajectory depends on the
aggregation rule but not on the enabled screening stages, so the matched clean
trajectory is reused within each aggregation mode.

\input{appendix/component_ablation_appendix.tex}

\paragraph{Role of cone screening.}
Cone screening has the clearest structural-robustness effect.  Holding the
scale screen and loss-normalized aggregation fixed, removing the cone screen
eliminates the IPM terminal window on CIFAR-10, MNIST, EMNIST-Balanced, and
AGNews, whereas the complete \methodname{} configuration reaches the registered
endpoint on all four datasets.  The same pattern holds when both screens are
removed.  This supports the intended role of the cone stage against invalid
signed mixtures and other structural violations, rather than a claim that it
uniformly increases clean or per-attack endpoint accuracy.

\paragraph{Role of scale screening.}
With cone screening and loss normalization fixed, adding the scale stage
changes the worst-attack result from 52.38\%/40.25\% to 77.50\%/68.37\% on
CIFAR-10, from 91.83\%/86.29\% to 92.35\%/87.01\% on MNIST, from
79.21\%/69.89\% to 79.30\%/69.89\% on EMNIST-Balanced, and from
53.27\%/36.15\% to 53.34\%/38.33\% on
AGNews.  Thus the scale stage never lowers the paired worst-attack result and is
decisive on CIFAR-10, although the small differences on the other datasets
should not be over-interpreted from a single seed.  Its role is also visible
without loss normalization: it prevents the severe CIFAR-10 Magnitude failure
and the MNIST cone-only Magnitude collapse.  Conversely, removing the scale
stage improves the Magnitude endpoint on EMNIST and AGNews.  The evidence
therefore supports scale screening as protection against radial worst cases,
not as a component that uniformly raises every attack endpoint.

\paragraph{Role of loss normalization.}
Loss normalization is an optimization and stability component with a
dataset-dependent effect.  With both screens fixed, it raises the complete
CIFAR-10 worst-attack result from 70.06\%/59.88\% to 77.50\%/68.37\%.  On
EMNIST-Balanced,
the mean-$u$ Sign-Flip and IPM runs have no terminal window, while the
loss-normalized configuration completes all seven attacks.  The MNIST worst-attack results
are essentially unchanged (92.34\%/87.06\% for mean-$u$ and
92.35\%/87.01\% with loss normalization).  On AGNews, mean-$u$ instead has
the higher worst-attack result, 55.45\%/40.77\% versus 53.34\%/38.33\%.  Hence the
factorial supports a long-horizon
stabilization benefit on CIFAR-10 and EMNIST, a neutral effect on MNIST, and a
trade-off on AGNews; it does not support a claim of uniform endpoint
improvement from loss normalization.

Of the 64 conceptual conditions per auxiliary dataset, 58 CIFAR-10, 53
EMNIST-Balanced, and 60 AGNews conditions reach rounds 990--999 after clean
trajectory reuse.  Every remaining dash denotes a registered collapse rather
than an unfinished scheduler job.  In particular, two continued CIFAR-10
configurations and five continued EMNIST mean-$u$ configurations collapse
after round 299; all launched AGNews continuations reach the terminal window.

\subsection{Stage-wise screening behavior}
\label{app:stagewise-screening}

\begin{figure}[!htbp]
\centering
\includegraphics[width=0.9\textwidth]{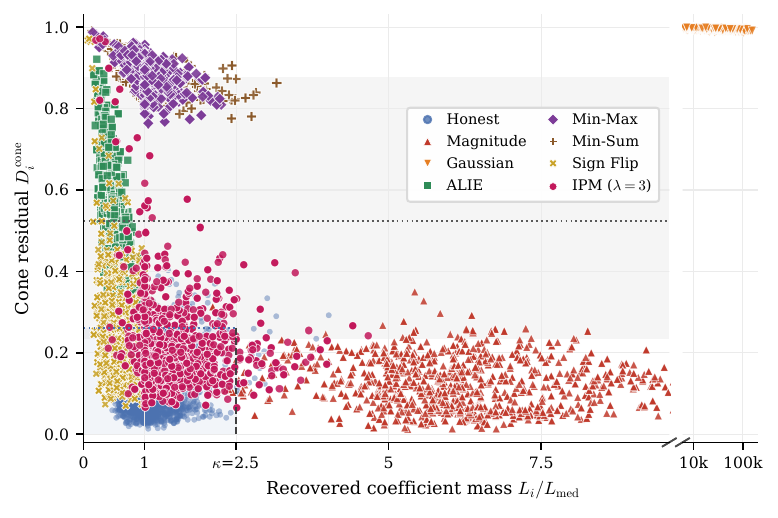}
\caption{\textbf{Uncurated version of Figure~\ref{fig:geometry-main}(c).}
MNIST, seed 42, $q=1$, all 1000 rounds of the clean and seven attacked runs; a
plain random subsample of 3,000 honest and 1,500 submissions per attack with no
quantile filtering. The shaded box is bounded by $\kappa=2.5$ and the honest
95th-percentile cone residual (0.26); the gray dotted line and band are the
median and interquartile range of the per-round Stage II cutoff (0.52,
[0.24, 0.88]). Gaussian submissions are on the broken right axis. Sign Flip
and IPM overlap the honest box in 69.8\% and 68.4\% of rounds; Min-Max and
Min-Sum coincide.}
\label{fig:cone-diagnostic-full}
\end{figure}

\begin{table}[!htbp]
\centering
\setlength{\tabcolsep}{4pt}
\scriptsize
\caption{\textbf{Stage-wise detection on MNIST, seed 42.} Statistics are
pooled over all 1000 rounds. Recall, honest FPR, and fair mass are percentages;
fair mass is the retained fraction of honest $F_i^q$ mass.}
\label{tab:detection-main}
\begin{tabular}{lccccc}
\toprule
Attack & Recall $\uparrow$ & Honest FPR $\downarrow$ & Fair mass $\uparrow$ & Scale used & Cone used\\
\midrule
Sign Flip & 58.4 & 16.7 & 82.9 & 0.15 & 3.85\\
IPM ($\lambda=3$) & 59.4 & 16.2 & 83.6 & 0.06 & 3.94\\
Gaussian & \textbf{100.0} & \textbf{0.0} & \textbf{100.0} & \textbf{4.00} & 0.00\\
ALIE & 94.9 & 2.0 & 97.3 & 0.17 & 3.83\\
Min-Max & 99.6 & 0.2 & \underline{99.8} & 0.01 & \textbf{3.98}\\
Min-Sum & \underline{99.7} & \underline{0.1} & \underline{99.8} & 0.03 & \underline{3.97}\\
Magnitude & 99.5 & 0.2 & \underline{99.8} & \underline{3.98} & 0.02\\
\bottomrule
\end{tabular}
\end{table}

Gaussian and Magnitude Scaling primarily consume scale-stage budget, whereas ALIE, Min-Max, and Min-Sum primarily consume cone-stage budget. Sign Flip and IPM are harder: their Byzantine recall is about 59\%, and the selector also rejects about 16\% of honest observations.

\subsection{From simplex to cone: controlled fairness-warp tests}
\label{app:simplex-cone}

This subsection separates two questions that the end-to-end BOBA comparison
of Section~\ref{sec:main-results} cannot separate: whether the unit-sum
simplex is the right observation model for fairness-scaled messages, and
whether a training run built on it is robust. The first question is answered
by a fixed-checkpoint reconstruction diagnostic; the second by the training
control of Table~\ref{tab:q-mechanism-main}.

\begin{figure}[!htbp]
\centering
\includegraphics[width=\textwidth]{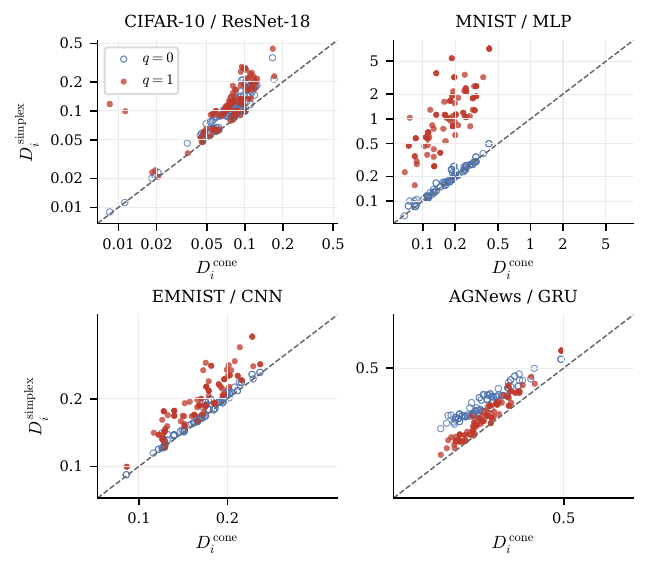}
\caption{\textbf{Unit-sum versus cone residual for the same honest
submissions.} Each panel uses that dataset's own screening coordinates
(classifier block on the three vision benchmarks, all parameters on AGNews),
seed-43 checkpoint at round 600, rounds 600--609 (100 submissions per
dataset); plain updates ($q=0$, hollow) and fairness-scaled submissions
($q=1$, filled). Every point lies above the diagonal; log axes.}
\label{fig:simplex-vs-cone}
\end{figure}

\paragraph{Fixed-checkpoint reconstruction residuals.}
Table~\ref{tab:simplex-vs-cone} and Figure~\ref{fig:simplex-vs-cone} use the
same checkpoint, the same sampled honest clients, and the same trusted basis
$G$ for both geometries, so that only the feasible set of the coefficient fit
differs. For $q=0$ the message is the plain update and its nonnegative cone
mass $\|\hat a_i\|_1$ is already close to one on MNIST and EMNIST (0.88 and 0.84), so the unit-sum and cone residuals nearly coincide there; on
CIFAR-10 and AGNews the plain updates have mass 0.65 and 0.74 and the unit-sum
fit is already worse. For $q=1$ the cone residual is unchanged on every submission,
because the cone is invariant to $u_i\mapsto F_iu_i$, whereas the unit-sum
residual moves with the distance of the true mass from one: on MNIST
($F_i\approx0.24$) it rises from 0.19 to 1.08, i.e.\ the simplex fit becomes
worse than predicting zero; on EMNIST ($F_i\approx0.77$) and CIFAR-10
($F_i\approx0.92$) the increase is moderate; on AGNews ($F_i\approx1.10$) the
scaling moves the mass toward one and the unit-sum penalty shrinks. The
direction of the effect therefore follows $|\log\|\hat a_i\|_1|$, not the
spread of the losses inside a round. The unit-sum residual is never smaller
than the cone residual (the simplex is a subset of the cone), and the
diagnostic makes no statement about attacks: a negatively scaled mean has
zero nonnegative mass under either model, so this experiment does not by
itself explain BOBA's IPM collapse.

\input{appendix/simplex_vs_cone_table.tex}

\paragraph{Training control.}
\begin{table}[!htbp]
\centering\small
\caption{Controlled fairness-warp mechanism test on MNIST/MLP, seed 42,
1000 rounds. Entries are global / worst-10\% client accuracy (\%) over rounds
990--999; ``n/a'' = condition not run. The two $q=1$ screening rows differ
only in the Stage II score (unit-sum simplex distance versus nonnegative-cone
distance); everything else, including Stage I and the loss-normalized
aggregation, is identical.}
\label{tab:q-mechanism-main}
\setlength{\tabcolsep}{3pt}
\begin{tabular}{llcccccc}
\toprule
Structural model & $q$ & Clean & Sign Flip & Magnitude & IPM & Min-Max & Min-Sum\\
\midrule
No screening & 1 & 93.95/89.26 & 92.39/87.11 & 95.20/90.06 & n/a & n/a & n/a\\
Simplex screening & 0 & 96.55/92.13 & 96.44/91.93 & 96.62/92.33 & n/a & n/a & n/a\\
Simplex screening & 1 & 93.95/89.26 & 93.85/89.08 & 93.95/89.26 & 89.91/83.98 & 92.91/87.46 & 92.83/87.40\\
Scale + cone & 1 & 93.95/89.26 & 93.20/88.09 & 93.96/89.27 & 92.35/87.01 & 93.94/89.23 & 93.96/89.26\\
\bottomrule
\end{tabular}
\end{table}

\begin{table}[!htbp]
\centering\small
\caption{Screening behavior behind the two $q=1$ rows of
Table~\ref{tab:q-mechanism-main}: Byzantine recall and honest false-rejection
rate, pooled over all 1000 rounds (4 Byzantine and 10 honest submissions per
round).}
\label{tab:q-mechanism-screening}
\setlength{\tabcolsep}{4pt}
\begin{tabular}{lcccccc}
\toprule
& \multicolumn{3}{c}{Simplex screening, $q=1$} & \multicolumn{3}{c}{Scale + cone, $q=1$}\\
\cmidrule(lr){2-4}\cmidrule(lr){5-7}
Attack & Recall & Honest FPR & Global acc. & Recall & Honest FPR & Global acc.\\
\midrule
Sign Flip & 0.92 & 0.03 & 93.85 & 0.58 & 0.17 & 93.20\\
Magnitude & 0.99 & 0.00 & 93.95 & 0.99 & 0.00 & 93.96\\
IPM & 0.45 & 0.22 & 89.91 & 0.62 & 0.15 & 92.35\\
Min-Max & 0.56 & 0.18 & 92.91 & 1.00 & 0.00 & 93.94\\
Min-Sum & 0.62 & 0.15 & 92.83 & 1.00 & 0.00 & 93.96\\
\bottomrule
\end{tabular}
\end{table}

Table~\ref{tab:q-mechanism-main} replaces \methodname{}'s nonnegative-cone
fit by a unit-sum simplex fit inside the same pipeline. The $q=0$ simplex row
is a sanity check of the unwarped geometry: with plain updates the simplex is
the appropriate model and performs well. It cannot, however, be compared
directly with the $q=1$ rows, because changing $q$ alters both the upload
geometry and the optimized fairness objective, so the accuracy difference does
not isolate geometry. The matched $q=1$ comparison is the relevant test of
observation-model mismatch, and Table~\ref{tab:q-mechanism-screening} shows
where the two models separate. Under Clean and Magnitude the rows coincide
(Stage I, shared by both, removes Magnitude). Under Sign Flip the simplex
score is the \emph{better} detector (recall 0.92 against 0.58, honest FPR
0.03 against 0.17): a negated honest mean has a negative coefficient sum, so
the unit-sum constraint rejects it directly, whereas the cone only measures
its residual; the MLP tolerates the leaked vectors and the accuracy gap is
0.65 points in the simplex's favor. Under the invalid-mixture attacks the
picture reverses. For Min-Max and Min-Sum the cone score removes essentially
every Byzantine submission (recall 1.00, FPR 0.00) while the simplex score
removes 56--62\% of them and rejects 15--18\% of honest clients instead; for
IPM the simplex recall drops to 0.45 with FPR 0.22 against 0.62 and 0.15
for the cone. The corresponding accuracy differences on MNIST are moderate
(1.0--2.4 percentage points global, 1.8--3.0 worst-10\%) because the MLP is
forgiving, but the screening statistics separate the two geometries
clearly: fairness-scaled honest updates have coefficient mass
$F_i^q\approx0.2$--$0.3$, so their unit-sum residuals overlap those of the
attacks (Table~\ref{tab:simplex-vs-cone}) and the simplex ranking spends its
budget on honest clients. These numbers do not license a claim of universal
cone superiority (Sign Flip is a counterexample), but they do isolate the
mechanism: on the attacks where end-to-end BOBA fails, replacing the
nonnegative cone by the unit-sum simplex inside \methodname{}'s own pipeline
reproduces the loss of recall, with no other component changed.

\subsection{Loss-scale recovery across fairness strengths}
\label{app:q-scale-recovery}

The ideal coefficient identity $L_i=F_i^q$ implies
$
\frac{1}{q}\log L_i=\log F_i,
\qquad q>0.
$
We test this identity in a matched MNIST/MLP checkpoint experiment at
$q\in\{0.1,0.5,1.0,1.5,2.0,3.0\}$.  For every value of $q$, \methodname{} is
trained from scratch with that same upload exponent to round 299, using the
same seed, partition, optimizer, scale-and-cone selector, and previous
server-loss q-FFL aggregation.  Trusted-reference augmentation is fixed to
one view throughout.  We then freeze the corresponding checkpoint, form
$u_i=F_i^q\Delta_i$ for all 100 honest clients, and rerun the full-parameter
nonnegative-cone recovery to obtain $L_i=\lVert\widehat a_i\rVert_1$.  Thus,
each panel contains 100 clients evaluated at one matched checkpoint; no
Byzantine submissions or pooled training-round observations enter this
diagnostic.

\begin{figure}[!htbp]
\centering
\includegraphics[width=\textwidth]{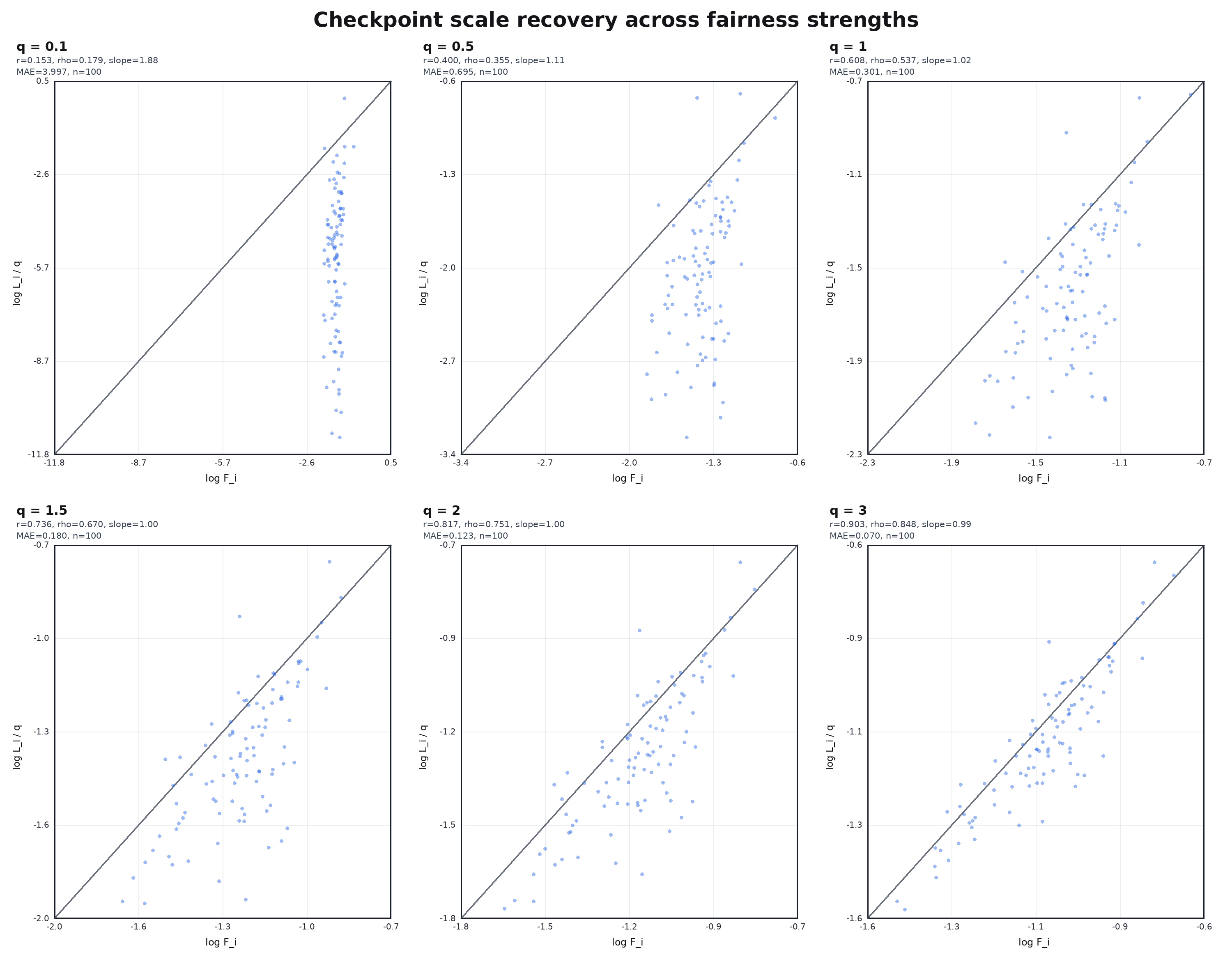}
\caption{\textbf{Checkpoint recovery of client loss scale across $q$.} Each
panel plots $x=\log F_i$ against
$y=q^{-1}\log L_i=\log L_i^{1/q}$ for all 100 honest clients at the matched
round-299 checkpoint; the diagonal is $y=x$. Pearson $r$, Spearman $\rho$,
ordinary-least-squares slope, and mean absolute log error (MALE) use all
points. Axes retain equal scaling and show the central 98\% only for
visualization.}
\label{fig:q-scale-recovery}
\end{figure}

\begin{table}[!htbp]
\centering\small
\caption{Numerical summary of Fig.~\ref{fig:q-scale-recovery}. Perfect loss-scale recovery has Pearson and Spearman correlation one, slope one, and zero mean absolute log error.}
\label{tab:q-scale-recovery}
\begin{tabular}{ccccc}
\toprule
$q$ & Pearson $r$ & Spearman $\rho$ & OLS slope & MALE\\
\midrule
0.1 & 0.153 & 0.179 & 1.88 & 3.997\\
0.5 & 0.400 & 0.355 & 1.11 & 0.695\\
1.0 & 0.608 & 0.537 & 1.02 & 0.301\\
1.5 & 0.736 & 0.670 & \textbf{1.00} & 0.180\\
2.0 & \underline{0.817} & \underline{0.751} & \textbf{1.00} & \underline{0.123}\\
3.0 & \textbf{0.903} & \textbf{0.848} & \underline{0.99} & \textbf{0.070}\\
\bottomrule
\end{tabular}
\end{table}

The inverse transformation becomes progressively better calibrated as $q$
increases in this checkpoint audit: the fitted slope is within $0.02$ of one
for $q\ge1$, while MALE decreases from $0.301$ at $q=1$ to $0.070$ at $q=3$.
Rank recovery also strengthens, reaching Spearman $\rho=0.848$ at $q=3$.
In contrast, $q=0.1$ is a clear failure regime. If
$\log L_i=q\log F_i+e_i$, then the reconstructed log loss is
$q^{-1}\log L_i=\log F_i+e_i/q$; hence coefficient-recovery error is amplified
by $1/q$. The result supports using $L_i$ directly as a radial screening
statistic, while cautioning against interpreting $L_i^{1/q}$ as an accurate
client-loss estimator when $q$ is small.

\subsection{Local-step sensitivity}
\label{app:local-steps}

The analysis of Section~\ref{sec:geometry} and the main benchmark use one
full-batch local step, for which $\Delta_i\approx Gp_i$ is a first-order
identity. Table~\ref{tab:local-steps} asks whether the geometry survives
several local steps. Honest clients take $E\in\{2,5\}$ full-batch steps
(learning rate 0.1) and upload $F_i(w_t)^q(w_i-w_t)$, with $F_i(w_t)$ the
train-mode loss of the first step; the server references $g_c$ are built with
the same $E$ steps, so the basis and the submissions undergo the same local
trajectory. Everything else follows the main protocol (MNIST/MLP, seed 42,
$q=1$, 1000 rounds; $E=1$ is the main-table run).

The mixture structure degrades gracefully rather than breaking. The honest
cone residual grows with $E$ (median 0.139, 0.246, 0.358 for $E=1,2,5$) but
stays well below the residuals of the structural attacks (0.49--1.00 in
Figure~\ref{fig:geometry-main}c), and the scale identity is preserved:
the correlation between $L_i$ and $F_i^q$ is 0.944, 0.981, and 0.976. Clean
accuracy rises with more local computation (93.95, 94.98, 96.06), every
attack except Sign Flip and IPM stays within 0.1 points of the corresponding
clean run, and the seven-attack floor moves from 92.35/87.01 to 91.33/85.96
and 91.45/85.78, i.e.\ IPM remains the binding attack and its gap to clean
widens from 1.6 to 3.7 and 4.6 points as the pooled recall over the seven
attacks falls from 0.88 to 0.81 and 0.79 (honest FPR 0.05 to 0.08). The
applicable regime is therefore not limited to a single step, but the
guarantees should be read as small-local-step statements: with more local
progress the honest updates drift away from the trusted class directions and
the cone-preserving attacks gain room.

\input{appendix/local_steps_table.tex}

\subsection{Operational sensitivity to fairness strength and scale tolerance}

\begin{table}[!htbp]
\centering
\small
\caption{Final-method MNIST/MLP sensitivity, seed 42, rounds 290--299.
Worst-attack results use all seven attacks; FPR and retained fair mass are pooled
across attacked conditions. Entries are percentages.}
\label{tab:operational-sensitivity}
\begin{tabular}{lcccc}
\toprule
$q$ & 0.5 & 1.0 & 1.5 & 2.0\\
\midrule
Worst-attack global accuracy & \textbf{91.57} & \underline{91.18} & 90.84 & 90.47\\
Worst-attack worst-10\% accuracy & \textbf{86.21} & \underline{85.62} & 85.16 & 84.85\\
\addlinespace
\midrule
$\kappa$ & 2.0 & 2.5 & 3.0 & 3.5\\
\midrule
Worst-attack global / worst-10\% & \textbf{91.20}/85.60 & 91.18/\textbf{85.62} & \underline{91.19}/\underline{85.61} & 91.18/\underline{85.61}\\
Honest FPR & 5.71 & \underline{5.00} & \textbf{4.14} & \textbf{4.14}\\
Retained fair mass & 93.66 & 94.70 & \textbf{95.75} & \underline{95.74}\\
\bottomrule
\end{tabular}
\end{table}

Theorem~\ref{thm:scale_cone_separation} predicts that larger $q$ widens honest
radial spread, while inverse scale recovery improves because log-scale error
is divided by $q$. Across the tested $\kappa$ values, the worst-attack accuracy is
nearly unchanged; increasing $\kappa$ from 2.0 to 3.0 reduces honest FPR and
raises retained fair mass.

\subsection{High-loss-client audit}


\begin{table}[!htbp]
\centering\small
\caption{Honest-client rejection by within-round start-loss quantile, pooled over the seven attacked conditions. Rejection share is the fraction of all rejected honest observations contributed by the row.}
\label{tab:loss-quantile-full}
\begin{tabular}{llccc}
\toprule
Dataset & Honest loss quantile & Rejection rate & Retained fair mass & Rejection share\\
\midrule
CIFAR-10/ResNet-18 & 0--50\% & \underline{3.15\%} & \underline{97.45\%} & \textbf{58.41\%}\\
 & 50--80\% & \textbf{1.44\%} & \textbf{98.99\%} & 16.05\%\\
 & 80--100\% & 3.45\% & 96.64\% & \underline{25.54\%}\\
MNIST & 0--50\% & \textbf{4.74\%} & \textbf{95.42\%} & \textbf{46.82\%}\\
 & 50--80\% & \underline{5.19\%} & \underline{95.20\%} & \underline{30.76\%}\\
 & 80--100\% & 5.67\% & 94.65\% & 22.41\%\\
EMNIST-Balanced & 0--50\% & 1.51\% & 99.26\% & \textbf{61.00\%}\\
 & 50--80\% & \textbf{0.89\%} & \textbf{99.55\%} & \underline{21.64\%}\\
 & 80--100\% & \underline{1.07\%} & \underline{99.48\%} & 17.36\%\\
AGNews & 0--50\% & \textbf{5.53\%} & \textbf{95.09\%} & \textbf{40.60\%}\\
 & 50--80\% & \underline{6.02\%} & \underline{94.30\%} & 26.52\%\\
 & 80--100\% & 11.19\% & 89.83\% & \underline{32.87\%}\\
\bottomrule
\end{tabular}
\end{table}

The AGNews tail is the least well preserved, consistent with the weaker IPM result in Table~\ref{tab:agnews-meanstd-5seed}.

Table~\ref{tab:high-loss-per-attack-5seed} extends the highest-loss-20\% audit to
the five registered seeds and splits it by attack. It pools the \emph{full}
training horizon, whereas the matched comparison of
Table~\ref{tab:high-loss-main} pools the first 300 rounds, the window over
which the H-nobs and BOBA audit runs were reproduced. The two windows are not
interchangeable: early in training the honest losses are still close together,
so every method rejects fewer high-loss clients and retention is higher. For
\methodname{} the pooled retained mass over the first 300 rounds is
99.77/98.48/100.00/90.06\% on CIFAR-10/MNIST/EMNIST-Balanced/AGNews against
95.23/94.27/99.65/91.40\% over the full horizon. The ordering between methods
is unaffected. The clean-side attacks
(Gaussian, Magnitude, and on most datasets Min-Max and Min-Sum) never remove a
high-loss honest client. The cost is concentrated in attacks whose Byzantine
vectors are placed near the honest cone: Sign Flip on MNIST, ALIE on CIFAR-10,
and ALIE on AGNews, where the cone stage removes the honest tail clients whose
class mixtures are least typical.

\input{appendix/high_loss_multiseed_per_attack.tex}

\subsection{Aggregation controls}
\label{app:aggregation}

\begin{table}[!htbp]
\centering\scriptsize
\caption{Post-screening aggregation controls, seed 42. Entries are global /
worst-10\% client accuracy (\%). Worst attack is the componentwise
minimum across the available attacked conditions. A dash indicates that at
least one required attack lacks a final window.}
\label{tab:aggregation-controls}
\begin{tabular}{llcc}
\toprule
Dataset & Aggregation & Clean & Worst attack\\
\midrule
CIFAR-10/ResNet-18 & server-loss q-FFL & 69.13/59.12 & \underline{63.38}/\underline{52.19}\\
 & mean-$u$ & \underline{79.25}/\underline{70.60} & 57.01/45.18\\
 & loss-normalized & \textbf{82.19}/\textbf{72.98} & \textbf{76.92}/\textbf{67.30}\\
\addlinespace
MNIST/MLP & server-loss q-FFL & \underline{95.54}/\underline{91.46} & \textbf{93.86}/\textbf{88.96}\\
 & mean-$u$ & 93.98/89.29 & \underline{92.81}/\underline{87.69}\\
 & recovered q-FFL & \textbf{95.56}/\textbf{91.47} & --\\
 & loss-normalized & 93.95/89.26 & 92.77/87.60\\
\addlinespace
EMNIST-Balanced/CNN & server-loss q-FFL & \underline{73.14}/\underline{64.03} & \underline{73.14}/\underline{63.56}\\
 & mean-$u$ & 60.96/50.97 & 9.14/2.83\\
 & recovered q-FFL & 69.65/60.30 & --\\
 & loss-normalized & \textbf{79.55}/\textbf{70.28} & \textbf{78.68}/\textbf{69.37}\\
\addlinespace
AGNews/GRU & server-loss q-FFL & 72.02/53.93 & \underline{63.86}/\underline{45.27}\\
 & mean-$u$ & \underline{82.01}/\underline{66.51} & 35.65/22.83\\
 & recovered q-FFL & 68.84/50.24 & --\\
 & loss-normalized & \textbf{83.79}/\textbf{68.08} & \textbf{66.29}/\textbf{45.70}\\
\bottomrule
\end{tabular}
\end{table}

\begin{table}[!htbp]
\centering\small
\caption{Loss-normalization exponent control, seed 42. Entries are global /
worst-10\% client accuracy (\%). CIFAR-10 uses the auxiliary five-layer CNN
in this controlled comparison. A dash for EMNIST means that all eight
conditions (Clean and seven attacks) met the registered collapse criterion.}
\label{tab:denominator-exponent}
\begin{tabular}{llcc}
\toprule
Dataset & Denominator exponent & Clean & Worst attack\\
\midrule
CIFAR-10/CNN & $q+1$ & \underline{39.80}/\underline{27.55} & \underline{39.70}/\underline{27.45}\\
CIFAR-10/CNN & $q$ & \textbf{46.53}/\textbf{34.15} & \textbf{46.35}/\textbf{33.93}\\
MNIST/MLP & $q+1$ & \underline{93.93}/\underline{89.23} & \underline{92.75}/\underline{87.64}\\
MNIST/MLP & $q$ & \textbf{93.95}/\textbf{89.26} & \textbf{92.77}/\textbf{87.60}\\
EMNIST/CNN & $q+1$ & -- & --\\
EMNIST/CNN & $q$ & \textbf{79.55}/\textbf{70.28} & \textbf{78.68}/\textbf{69.37}\\
AGNews/GRU & $q+1$ & \underline{82.75}/\underline{66.31} & \underline{65.23}/\underline{45.87}\\
AGNews/GRU & $q$ & \textbf{83.79}/\textbf{68.08} & \textbf{66.29}/\textbf{45.70}\\
\bottomrule
\end{tabular}
\end{table}

The eight EMNIST $q+1$ runs did not fail because of the scheduler, an
out-of-memory error, or a non-finite value.  They all stopped at round 123
under the registered chance-level rule.  For 47 classes, this rule declares
collapse after 25 consecutive evaluated rounds at or below
$1/47+0.005=2.628\%$, beginning no earlier than the 100th one-indexed round.
The recorded accuracies from zero-indexed rounds 99 through 123 all satisfy
this condition; at round 123 the clean run has 2.622\% global accuracy and
0\% worst-10\% client accuracy.  Hence round 123 is the first round at which
the 25-round patience can be exhausted.

The underlying failure is excessive step attenuation in this historical
denominator control.  Because $q=1$, using exponent $q+1$ squares the
reconstructed loss.  At round 0 of the clean EMNIST run, where the estimated
losses are close to $\log 47\approx3.85$, the $q+1$ denominator is 148.74,
compared with 38.57 for exponent $q$.  The corresponding aggregate-update
norm is $8.99\times10^{-4}$ rather than $3.47\times10^{-3}$, a reduction by
a factor of about 3.86.  Test loss decreases only from 3.853 to 3.846 by
round 123, so the model does not leave the chance-level region before the
registered stop.  The same outcome in the clean condition shows that this is
an optimization failure caused by the denominator choice, rather than an
attack-specific failure.  With exponent $q$, the matched clean run reaches
10.31\% accuracy at round 123 and later completes normally. This diagnostic
illustrates the optimization role of Component III: normalization should
suppress excessive post-screening steps without over-attenuating the update.
The $q+1$ denominator violates this balance on EMNIST, whereas the final
exponent $q$ preserves a substantially larger update and completes training
normally.

\subsection{Attack strength and budget}

\begin{table}[!htbp]
\centering\small
\caption{CIFAR-10/ResNet-18 sign-reversal strength, seed 42. Entries are
global / worst-10\% client accuracy (\%).}
\label{tab:sign-strength}
\begin{tabular}{lccc}
\toprule
Method & $\lambda=1$ & $\lambda=2$ & $\lambda=3$\\
\midrule
\methodname{} & \textbf{81.73}/\textbf{73.35} & \textbf{80.26}/\textbf{71.32} & \textbf{81.43}/\textbf{72.18}\\
server-loss q-FFL aggregation & 66.43/55.37 & 63.38/52.19 & 62.89/51.65\\
mean-$u$ aggregation & \underline{78.86}/\underline{69.87} & \underline{76.64}/\underline{68.34} & \underline{72.31}/\underline{61.88}\\
\bottomrule
\end{tabular}
\end{table}

\begin{table}[!htbp]
\centering\small
\caption{Declared-budget sensitivity for final \methodname{} on MNIST, seed 42, with four actual Byzantine submissions.}
\label{tab:budget-final}
\begin{tabular}{cc}
\toprule
Declared $f$ & Worst attack (global / worst-10\% accuracy, \%)\\
\midrule
2 & 81.91/74.58\\
4 & \underline{92.35}/\underline{87.01}\\
6 & \textbf{92.86}/\textbf{87.63}\\
\bottomrule
\end{tabular}
\end{table}

Under-declaring the budget is harmful under Gaussian perturbations. The slight advantage of $f=6$ is a single-seed endpoint observation rather than evidence that over-rejection is generally beneficial.

\subsection{Trusted data, heterogeneity, and identifiability}

\begin{table}[!htbp]
\centering
\small
\caption{Empirical validity of the trusted screening geometry at fixed
CIFAR-10/ResNet-18 checkpoints. Captured energy is the fraction of honest
update energy represented by the trusted class span; oracle overlap compares
the trusted span with an all-client-data class-span oracle.}
\label{tab:screen-space-main}
\begin{tabular}{ccccc}
\toprule
Checkpoint & Classifier energy & Classifier overlap & Full energy & Full overlap\\
\midrule
Round 299  & 0.892 & \textbf{0.978} & \textbf{0.249} & \textbf{0.703}\\
Round 599  & \underline{0.901} & 0.972 & \underline{0.161} & \underline{0.595}\\
Round 899  & \textbf{0.918} & 0.967 & 0.120 & 0.461\\
Round 1499 & 0.852 & \underline{0.975} & 0.051 & 0.337\\
\bottomrule
\end{tabular}
\end{table}

\begin{table}[!htbp]
\centering\small
\caption{Trusted augmentation repeats at fixed CIFAR-10/ResNet-18 checkpoints, classifier coordinates, seed 42.}
\label{tab:trusted-repeats}
\begin{tabular}{rcccc}
\toprule
Repeats & Projection energy & Negative coefficient mass & Loss MAE & Time (s)\\
\midrule
1 & 0.9223 & 14.5973 & 1.7817 & \textbf{0.408}\\
6 & \textbf{0.9237} & \textbf{5.1632} & \textbf{1.2567} & \underline{1.061}\\
12 & 0.9234 & 7.6181 & \underline{1.3797} & 1.793\\
24 & \underline{0.9236} & \underline{6.6128} & 1.4823 & 3.358\\
64 & \underline{0.9236} & 7.0401 & 1.4833 & 8.584\\
\bottomrule
\end{tabular}
\end{table}

\begin{table}[!htbp]
\centering\small
\caption{Trusted-size sensitivity for final \methodname{} on MNIST, seed 42.}
\label{tab:trusted-final}
\begin{tabular}{cc}
\toprule
Samples/class & Worst attack (global / worst-10\% accuracy, \%)\\
\midrule
1 & 90.30/84.44\\
20 & \underline{92.35}/\underline{87.01}\\
50 & \textbf{92.79}/\textbf{87.52}\\
\bottomrule
\end{tabular}
\end{table}

\begin{table}[!htbp]
\centering\small
\caption{Label-skew sensitivity for final \methodname{} on MNIST, seed 42.}
\label{tab:heterogeneity-final}
\begin{tabular}{cc}
\toprule
Dirichlet $\alpha$ & Worst attack (global / worst-10\% accuracy, \%)\\
\midrule
0.1 & \underline{92.70}/86.93\\
0.5 & \textbf{92.86}/\textbf{87.98}\\
1.0 & 92.35/\underline{87.01}\\
\bottomrule
\end{tabular}
\end{table}

\begin{table}[!htbp]
\centering\small
\caption{Numerical identifiability of the trusted basis over completed rounds and conditions. Values are median [5th, 95th percentile].}
\label{tab:basis-identifiability}
\begin{tabular}{lccc}
\toprule
Dataset & Numerical rank & $\sigma_{\min}$ & Condition number\\
\midrule
CIFAR-10/ResNet-18 & 10 [10, 10] & 0.0823 [0.0649, 0.0971] & 6.78 [5.77, 9.22]\\
MNIST & 10 [10, 10] & 0.0590 [0.0529, 0.0785] & 8.64 [7.60, 9.99]\\
EMNIST-Balanced & 47 [47, 47] & 0.0968 [0.0145, 0.1951] & 14.26 [7.18, 34.02]\\
AGNews & 4 [4, 4] & 0.0286 [0.0157, 0.0456] & 4.09 [1.73, 7.09]\\
\bottomrule
\end{tabular}
\end{table}

\subsection{Computation and communication}
\label{app:cost}

\begin{table}[!htbp]
\centering\small
\caption{Measured server-side time per attacked round with 14 submissions. All listed methods communicate one model-shaped vector per client.}
\label{tab:cost}
\begin{tabular}{lccc}
\toprule
Method & Server time/round & Client vectors & Extra client statistics\\
\midrule
q-FFL-GD & \textbf{1.71} ms & 1 & none\\
H-nobs & \underline{3.05} ms & 1 & none\\
BOBA & 478.18 ms & 1 & none\\
\methodname{} & 1387.10 ms & 1 & none\\
\bottomrule
\end{tabular}
\end{table}

\methodname{} preserves the one-vector communication protocol but is approximately $2.9\times$ slower than BOBA in this measurement. The timing uses PyTorch on NVIDIA A100 resources and includes trusted-reference construction.

\subsection{Over-declared budgets and always-on screening}
\label{app:always-on}

The main benchmark declares $f_t=0$ in clean conditions and $f_t=4$ when
four Byzantine submissions are appended. To test deployment with an
over-declared budget, we also run clean training with $f_t=4$ for five seeds.

\begin{table}[!htbp]
\centering\small
\caption{Clean accuracy (\%) with a matched budget ($f_t=0$) and an
over-declared always-on budget ($f_t=4$), five seeds.}
\label{tab:always-on-clean}
\begin{tabular}{lcc}
\toprule
Dataset/model & Matched $f_t=0$ & Always-on $f_t=4$\\
\midrule
CIFAR-10/ResNet-18 & 82.25 & 79.68\\
MNIST/MLP & 93.83 & 93.78\\
EMNIST-Balanced/CNN & 79.12 & 80.06\\
\bottomrule
\end{tabular}
\end{table}

Accuracy changes little, but the fairness cost is substantial. Because the
fixed-budget rule rejects exactly four of ten honest submissions, its honest
FPR is 40\%; retained fair mass falls to 59--62\%, and the highest-loss 20\%
of clients retain only 53--79\% of their mass. A median-plus-three-MAD
threshold alternative reduces clean FPR to 2.8--5.7\% and retains
94.4--98.3\% of fair mass, but trades away attack recall, which drops from
88.0--96.5\% under the fixed budget to 70.7--78.7\%. Thus an accurate fixed
budget favors attack protection, whereas adaptive rejection better protects
fairness when the attack count is unknown.

\paragraph{Declared budget above the actual attack count.}
The matched benchmark gives every fixed-budget rule the true Byzantine upper
bound. Table~\ref{tab:budget-mismatch} keeps the declared budget at $f=4$
while only 1, 2, or 3 of the 14 submissions are Byzantine (MNIST/MLP, seed
42). Robustness is unaffected: accuracy stays within 0.7 points of the matched
run for every attack, and Gaussian, Min-Max, Min-Sum, and Magnitude are still
removed with recall 1.00. The cost is entirely on the fairness side and is
monotone in the over-declaration: with one actual attacker the screens must
still reject four submissions, so the honest false-rejection rate rises from
0.00--0.17 to 0.30--0.35 and the retained fair mass of the highest-loss 20\%
of honest clients falls from 0.84--1.00 to 0.69--0.75; with three attackers
the values are 0.10--0.20 and 0.84--0.96. Sign Flip and IPM recall also
degrades as the budget is over-declared (0.58/0.62 at four attackers,
0.77/0.78 at one), because the surplus budget is spent on honest clients
whose cone scores overlap those attacks. Together with the clean $f=4$ run
(FPR 0.40, retained mass 0.61) this quantifies the price of not knowing the
attack count: accuracy is preserved, fairness mass is not.

\input{appendix/budget_mismatch_table.tex}

\subsection{Covariate-shift stress tests}
\label{app:domain-shift}

We stress-test CIFAR-10/ResNet-18 with client-specific photometric changes,
rotations, and CIFAR-10-C corruptions at severities 3 and 5. These experiments
use a common 300-round diagnostic window. Shift lowers clean accuracy for all
methods, but \methodname{}'s worst-attack accuracy remains within one point of
its own shifted-clean value in every condition; for example, severity-3
corruption gives 51.75\% clean and 51.03\% under the worst attack. The trusted basis
is built from unshifted clean examples, yet shifted honest updates retain cone
medians of 0.060--0.069, honest FPR of 1.4--2.0\%, total fair mass of 95--96\%,
and high-loss fair mass of 98.5--99.5\%.

Varying the trusted set from 5 to 50 examples per class changes the four-attack
worst-attack accuracy by at most 1.5 points in both the clean and corrupted domains. Honest
FPR falls from 3.2\% to 0.9\%, while retained fair mass rises from 97.5\% to
99.2\%. These results show empirical tolerance to the tested covariate shifts;
they do not establish guarantees for arbitrary violations of shared
class-conditionals.

\subsection{Adaptive cone-preserving attacks}
\label{app:adaptive-attacks}

The adaptive attacker constructs a nonnegative trusted-cone vector with a
coefficient mass chosen to remain plausible under Stage I. Its cone score can
therefore be lower than the honest median, directly realizing the observable
boundary of Theorem~\ref{thm:scale_cone_separation}.

\begin{table}[!htbp]
\centering\small
\caption{CIFAR-10 diagnostic accuracy (\%) under adaptive cone-preserving attacks.
The 300-round values describe the mid-training window and are not terminal
1500-round conclusions.}
\label{tab:adaptive-cifar}
\begin{tabular}{lccccc}
\toprule
Window & Clean & Surrogate & Oracle & Slack & Mixture\\
\midrule
Rounds 200--299 & 54.1 & 20.9 & 13.6 & 12.7 & 19.3\\
Rounds 1490--1499 & 82.4 & 65.6 & -- & -- & --\\
\bottomrule
\end{tabular}
\end{table}

The 300-round diagnostic shows severe degradation, but the full-horizon
surrogate runs partially recover to 65.6\% versus 82.4\% clean. We therefore
interpret the available evidence as a severe transient degradation with a
persistent terminal gap, not as universal permanent collapse. Short-horizon
Krum and FLTrust values are not used because those baselines have not
converged by round 300 on several dataset/model pairs. Cone-copy attacks,
which exactly reproduce a high-loss honest vector, remain within one to two
points of clean accuracy; harmful adaptive attacks require exploiting the
allowed cone and radial degrees of freedom rather than merely duplicating an
honest update.

\subsection{Other observable-boundary experiments}

A cone-copy attacker chooses a scale-plausible coefficient $a_b\succeq0$ and
submits $Ga_b$. Such a vector has no intrinsic cone-distance certificate of
malicious origin. On CIFAR-10/ResNet-18, seed 42, the cone-copy condition
completes with 78.32\%/68.54\% global / worst-10\% accuracy for
\methodname{} and 73.16\%/63.55\% for the unscreened q-FFL-GD control.

A parameter-selection attacker preserves the selected classifier coordinates
(it submits the honest classifier mean there) while reversing and amplifying
the unselected coordinates. On CIFAR-10/ResNet-18 both the classifier-selected
\methodname{} run and its full-parameter control triggered the collapse
criterion before the registered final window. The same attack on the
classifier-screened MNIST/MLP and EMNIST/CNN benchmarks (seed 42, $f=4$)
collapses \methodname{} at rounds 539 and 889: in the screened coordinates the
Byzantine submissions are indistinguishable from honest ones (honest cone
residual median 0.05 and 0.02, Byzantine recall 0.29 and 0.37, honest
false-rejection rate 0.28 and 0.25), so the screens spend their budget on
honest clients while the unselected coordinates carry the attack. This is the
sharpest empirical statement of the boundary in Section~\ref{sec:method}:
\methodname{} is robust only to deviations that are observable in the screening
coordinates; attacks confined to unobserved coordinates therefore fall outside
the guarantees of the selected-coordinate interface.

%% file: appendix/component_ablation_appendix.tex
\begin{table}[t]
\centering
\caption{\textbf{Component ablation of \methodname{}.} CIFAR-10/ResNet-18, seed 42, 1000 rounds. Global / worst-10\% client accuracy (\%) over the final ten rounds; the worst-attack column is the componentwise minimum over the seven attacks. Clean uses $f=0$; ``--'' marks a run with no terminal window.}
\label{tab:factorial-cifar}
\scriptsize
\setlength{\tabcolsep}{2.2pt}
\renewcommand{\arraystretch}{0.88}
\resizebox{\textwidth}{!}{\begin{tabular}{lccc|ccccccccc}
\toprule
Variant & Scale & Cone & Loss norm. & Clean & Magnitude & Sign & IPM$_{\lambda=3}$ & Gauss. & ALIE & Min-Max & Min-Sum & Worst attack\\
\midrule
\textbf{\methodname{}} & \checkmark & \checkmark & \checkmark & \textbf{79.03}/\textbf{70.41} & \textbf{78.97}/\textbf{70.38} & \underline{77.50}/\underline{68.37} & \textbf{77.83}/\textbf{68.91} & \textbf{78.82}/\textbf{69.57} & \underline{78.50}/\underline{69.28} & \textbf{79.28}/\textbf{70.73} & \underline{78.71}/\underline{69.93} & \textbf{77.50}/\textbf{68.37}\\
w/o scale & -- & \checkmark & \checkmark & \textbf{79.03}/\textbf{70.41} & 52.38/40.25 & \textbf{78.20}/\textbf{69.08} & \underline{76.08}/\underline{66.91} & \textbf{78.82}/\textbf{69.57} & \textbf{78.97}/\textbf{70.16} & \underline{79.00}/\underline{70.29} & \textbf{79.03}/\textbf{70.41} & 52.38/40.25\\
w/o cone & \checkmark & -- & \checkmark & \textbf{79.03}/\textbf{70.41} & \underline{78.67}/\underline{70.09} & 69.95/59.40 & -- & \textbf{78.82}/\textbf{69.57} & 55.16/42.83 & 36.71/24.49 & 38.94/27.18 & --\\
w/o scale and cone & -- & -- & \checkmark & \textbf{79.03}/\textbf{70.41} & 63.23/51.72 & 70.16/59.92 & -- & 26.11/13.69 & 63.20/50.88 & 34.07/22.76 & 35.66/23.99 & --\\
w/o loss norm. & \checkmark & \checkmark & -- & \underline{78.04}/\underline{69.52} & 70.06/59.88 & 73.33/63.97 & 74.01/64.70 & \underline{77.89}/\underline{68.61} & 76.98/67.90 & 76.04/67.28 & 76.02/67.30 & \underline{70.06}/\underline{59.88}\\
cone only & -- & \checkmark & -- & \underline{78.04}/\underline{69.52} & 10.89/0.05 & 71.59/62.06 & 75.30/65.97 & 77.65/69.14 & 76.87/68.40 & 78.04/69.52 & 78.04/69.52 & 10.89/0.05\\
scale only & \checkmark & -- & -- & \underline{78.04}/\underline{69.52} & 73.14/64.22 & 71.66/61.40 & -- & 77.74/69.17 & 46.45/34.25 & 13.87/2.92 & -- & --\\
no component & -- & -- & -- & \underline{78.04}/\underline{69.52} & 16.91/5.99 & 72.56/63.11 & -- & -- & 48.22/36.59 & 15.41/5.06 & 15.04/3.21 & --\\
\bottomrule
\end{tabular}}
\end{table}
\begin{table}[t]
\centering
\caption{\textbf{Component ablation of \methodname{}.} MNIST/MLP, seed 42, 1000 rounds. Global / worst-10\% client accuracy (\%) over the final ten rounds; the worst-attack column is the componentwise minimum over the seven attacks. Clean uses $f=0$; ``--'' marks a run with no terminal window.}
\label{tab:factorial-mnist}
\scriptsize
\setlength{\tabcolsep}{2.2pt}
\renewcommand{\arraystretch}{0.88}
\resizebox{\textwidth}{!}{\begin{tabular}{lccc|ccccccccc}
\toprule
Variant & Scale & Cone & Loss norm. & Clean & Magnitude & Sign & IPM$_{\lambda=3}$ & Gauss. & ALIE & Min-Max & Min-Sum & Worst attack\\
\midrule
\textbf{\methodname{}} & \checkmark & \checkmark & \checkmark & \underline{93.95}/\underline{89.26} & 93.94/89.23 & 93.20/88.09 & \textbf{92.35}/\textbf{87.01} & \underline{93.95}/\underline{89.26} & 93.94/89.14 & 93.94/89.23 & 93.96/89.26 & \textbf{92.35}/\textbf{87.01}\\
w/o scale & -- & \checkmark & \checkmark & \underline{93.95}/\underline{89.26} & \underline{94.40}/\underline{89.91} & 93.22/88.23 & 91.83/86.29 & \underline{93.95}/\underline{89.26} & 93.95/89.27 & 93.95/89.26 & 93.95/89.26 & 91.83/86.29\\
w/o cone & \checkmark & -- & \checkmark & \underline{93.95}/\underline{89.26} & 93.96/89.27 & 92.43/87.07 & -- & \underline{93.95}/\underline{89.26} & 93.97/89.38 & 89.06/83.15 & 55.42/41.64 & --\\
w/o scale and cone & -- & -- & \checkmark & \underline{93.95}/\underline{89.26} & \textbf{95.14}/\textbf{89.63} & 92.39/87.11 & -- & 52.87/39.71 & 93.96/89.21 & 89.78/84.07 & 79.52/69.42 & --\\
w/o loss norm. & \checkmark & \checkmark & -- & \textbf{93.97}/\textbf{89.29} & 93.98/89.26 & \underline{93.24}/\underline{88.15} & \underline{92.34}/\underline{87.06} & \textbf{93.97}/\textbf{89.29} & 93.96/89.14 & \textbf{93.98}/\textbf{89.29} & \textbf{93.98}/\textbf{89.29} & \underline{92.34}/\underline{87.06}\\
cone only & -- & \checkmark & -- & \textbf{93.97}/\textbf{89.29} & -- & \textbf{93.27}/\textbf{88.25} & 92.09/86.73 & \textbf{93.97}/\textbf{89.29} & \textbf{93.98}/\textbf{89.29} & \underline{93.97}/\underline{89.29} & \underline{93.97}/\underline{89.29} & --\\
scale only & \checkmark & -- & -- & \textbf{93.97}/\textbf{89.29} & 93.98/89.27 & 92.45/87.13 & 10.33/0.00 & \textbf{93.97}/\textbf{89.29} & \underline{93.97}/\underline{89.37} & 87.78/81.05 & 78.96/69.45 & 10.33/0.00\\
no component & -- & -- & -- & \textbf{93.97}/\textbf{89.29} & -- & 92.40/87.15 & -- & -- & 93.96/89.23 & 51.72/34.59 & 53.41/40.21 & --\\
\bottomrule
\end{tabular}}
\end{table}
\begin{table}[t]
\centering
\caption{\textbf{Component ablation of \methodname{}.} EMNIST-Balanced/CNN, seed 42, 1000 rounds. Global / worst-10\% client accuracy (\%) over the final ten rounds; the worst-attack column is the componentwise minimum over the seven attacks. Clean uses $f=0$; ``--'' marks a run with no terminal window.}
\label{tab:factorial-emnist}
\scriptsize
\setlength{\tabcolsep}{2.2pt}
\renewcommand{\arraystretch}{0.88}
\resizebox{\textwidth}{!}{\begin{tabular}{lccc|ccccccccc}
\toprule
Variant & Scale & Cone & Loss norm. & Clean & Magnitude & Sign & IPM$_{\lambda=3}$ & Gauss. & ALIE & Min-Max & Min-Sum & Worst attack\\
\midrule
\textbf{\methodname{}} & \checkmark & \checkmark & \checkmark & \underline{79.59}/\underline{70.34} & 79.59/70.34 & \textbf{79.31}/\textbf{70.05} & \textbf{79.51}/\textbf{70.33} & \textbf{79.30}/\textbf{69.89} & \textbf{79.70}/\textbf{70.65} & \underline{79.59}/\underline{70.34} & \underline{79.59}/\underline{70.34} & \textbf{79.30}/\textbf{69.89}\\
w/o scale & -- & \checkmark & \checkmark & \underline{79.59}/\underline{70.34} & \textbf{82.69}/\textbf{74.52} & \textbf{79.31}/\textbf{70.05} & \underline{79.21}/\underline{70.12} & \textbf{79.30}/\textbf{69.89} & \textbf{79.70}/\textbf{70.65} & \underline{79.59}/\underline{70.34} & \underline{79.59}/\underline{70.34} & \underline{79.21}/\underline{69.89}\\
w/o cone & \checkmark & -- & \checkmark & \underline{79.59}/\underline{70.34} & 79.59/70.34 & 71.50/62.16 & -- & \textbf{79.30}/\textbf{69.89} & 77.99/68.66 & 58.36/46.70 & 59.18/47.88 & --\\
w/o scale and cone & -- & -- & \checkmark & \underline{79.59}/\underline{70.34} & \underline{82.47}/\underline{73.37} & 71.50/62.16 & -- & -- & 77.99/68.66 & 58.36/46.70 & 59.18/47.88 & --\\
w/o loss norm. & \checkmark & \checkmark & -- & \textbf{80.34}/\textbf{71.73} & 80.34/71.73 & -- & -- & \underline{20.01}/\underline{11.20} & 58.01/48.11 & \textbf{80.34}/\textbf{71.73} & \textbf{80.34}/\textbf{71.73} & --\\
cone only & -- & \checkmark & -- & \textbf{80.34}/\textbf{71.73} & -- & -- & -- & \underline{20.01}/\underline{11.20} & 58.01/48.11 & \textbf{80.34}/\textbf{71.73} & \textbf{80.34}/\textbf{71.73} & --\\
scale only & \checkmark & -- & -- & \textbf{80.34}/\textbf{71.73} & 80.34/71.73 & \underline{76.71}/\underline{67.50} & -- & \underline{20.01}/\underline{11.20} & \underline{78.95}/\underline{70.21} & 50.48/38.00 & 52.31/40.19 & --\\
no component & -- & -- & -- & \textbf{80.34}/\textbf{71.73} & 78.21/68.79 & \underline{76.71}/\underline{67.50} & -- & -- & \underline{78.95}/\underline{70.21} & 50.48/38.00 & 52.31/40.19 & --\\
\bottomrule
\end{tabular}}
\end{table}
\begin{table}[t]
\centering
\caption{\textbf{Component ablation of \methodname{}.} AGNews/GRU, seed 42, 2000 rounds. Global / worst-10\% client accuracy (\%) over the final ten rounds; the worst-attack column is the componentwise minimum over the seven attacks. Clean uses $f=0$; ``--'' marks a run with no terminal window.}
\label{tab:factorial-agnews}
\scriptsize
\setlength{\tabcolsep}{2.2pt}
\renewcommand{\arraystretch}{0.88}
\resizebox{\textwidth}{!}{\begin{tabular}{lccc|ccccccccc}
\toprule
Variant & Scale & Cone & Loss norm. & Clean & Magnitude & Sign & IPM$_{\lambda=3}$ & Gauss. & ALIE & Min-Max & Min-Sum & Worst attack\\
\midrule
\textbf{\methodname{}} & \checkmark & \checkmark & \checkmark & \textbf{83.66}/\textbf{68.09} & 83.65/68.09 & 79.45/62.86 & 81.56/65.49 & \underline{83.35}/\underline{67.62} & \textbf{81.31}/\textbf{68.09} & \underline{83.14}/\underline{67.13} & 82.94/66.88 & \underline{79.45}/\underline{62.86}\\
w/o scale & -- & \checkmark & \checkmark & \underline{83.64}/\underline{68.03} & 86.53/73.99 & \textbf{83.64}/\textbf{68.03} & \textbf{83.64}/\textbf{68.03} & \textbf{83.65}/\textbf{68.06} & 80.44/66.99 & \textbf{83.64}/\textbf{68.03} & \underline{83.70}/\underline{68.16} & \textbf{80.44}/\textbf{66.99}\\
w/o cone & \checkmark & -- & \checkmark & \textbf{83.66}/\textbf{68.09} & 83.66/68.09 & -- & -- & 28.34/17.87 & 80.53/67.12 & 40.94/16.66 & 55.80/37.26 & --\\
w/o scale and cone & -- & -- & \checkmark & \textbf{83.66}/\textbf{68.09} & \underline{86.90}/\underline{75.05} & -- & -- & 28.41/16.32 & \underline{81.26}/\underline{67.35} & 43.55/21.00 & 56.08/36.75 & --\\
w/o loss norm. & \checkmark & \checkmark & -- & 81.93/66.28 & 81.93/66.28 & 62.69/46.97 & 68.08/50.74 & 35.92/25.22 & 71.58/58.13 & 76.38/62.02 & \textbf{84.18}/\textbf{68.95} & 35.92/25.22\\
cone only & -- & \checkmark & -- & 81.93/66.28 & 67.41/46.86 & \underline{81.93}/\underline{66.28} & \underline{81.93}/\underline{66.28} & 81.93/66.33 & 77.01/63.68 & 81.93/66.28 & 79.36/62.37 & 67.41/46.86\\
scale only & \checkmark & -- & -- & 81.93/66.28 & 81.93/66.28 & 65.85/46.21 & -- & 24.38/14.31 & 76.93/62.97 & 25.72/3.36 & 48.17/23.84 & --\\
no component & -- & -- & -- & 81.92/66.33 & \textbf{87.14}/\textbf{76.64} & 61.64/43.98 & -- & 25.16/14.26 & 78.51/64.55 & -- & 49.19/25.77 & --\\
\bottomrule
\end{tabular}}
\end{table}

%% file: appendix/simplex_vs_cone_table.tex
\begin{table}[!htbp]
\centering\scriptsize\setlength{\tabcolsep}{3pt}
\caption{\textbf{Unit-sum simplex versus nonnegative cone on identical inputs.} For each dataset the \methodname{} run's seed-43 checkpoint at round 600 is loaded, the next ten rounds' honest clients are replayed with the production local step (10 clients per round, 100 submissions), and the production server basis $G$ (20 trusted examples per class) is built once per round. Each honest update $\Delta_i$ is submitted as $u_i=F_i^q\Delta_i$ with $q=0$ (plain update) and $q=1$ (the paper's fairness-scaled vector, $F_i$ = start-of-round loss). $D^{\rm cone}=\min_{a\ge0}\|Su_i-Ga\|^2/\|Su_i\|^2$ and $D^{\rm simplex}=\min_{p\ge0,\mathbf 1^\top p=1}\|Su_i-Gp\|^2/\|Su_i\|^2$ (energy ratios) are computed in the same screened coordinates $S$ with the aggregator's own solvers; $\|\hat a_i\|_1$ is the mass of the cone fit. Medians over the 100 submissions; the gap is $D^{\rm simplex}-D^{\rm cone}$ and is positive in every case. No optimizer, trajectory, attack or aggregation rule is involved.}
\label{tab:simplex-vs-cone}
\resizebox{\textwidth}{!}{%
\begin{tabular}{llcccccccc}
\toprule
Dataset & Screened coords. & $q$ & $F_i$ med.\ [p10, p90] & $\|\hat a_i\|_1$ med. & $D^{\rm cone}$ med.\ / p90 & $D^{\rm simplex}$ med.\ / p90 & gap med.\ / p90 & $\Pr[D^{\rm simplex}>D^{\rm cone}]$\\
\midrule
CIFAR-10 / ResNet-18 & classifier (5,130) & 0 & 0.92 [0.81, 1.18] & 0.65 & 0.080 / 0.111 & 0.102 / 0.175 & +0.024 / +0.066 & 1.00\\
 &  & 1 & 0.92 [0.81, 1.18] & 0.59 & 0.080 / 0.111 & 0.111 / 0.223 & +0.035 / +0.113 & 1.00\\
 & full (11,173,962) & 0 & 0.92 [0.81, 1.18] & 0.28 & 0.823 / 0.890 & 1.242 / 1.660 & +0.427 / +0.799 & 1.00\\
 &  & 1 & 0.92 [0.81, 1.18] & 0.26 & 0.823 / 0.890 & 1.307 / 2.079 & +0.502 / +1.267 & 1.00\\
\midrule
MNIST / MLP & classifier (2,010) & 0 & 0.24 [0.20, 0.30] & 0.88 & 0.178 / 0.297 & 0.185 / 0.306 & +0.008 / +0.047 & 1.00\\
 &  & 1 & 0.24 [0.20, 0.30] & 0.21 & 0.178 / 0.297 & 1.077 / 3.247 & +0.899 / +3.057 & 1.00\\
 & full (159,010) & 0 & 0.24 [0.20, 0.30] & 0.57 & 0.411 / 0.575 & 0.484 / 0.751 & +0.066 / +0.269 & 1.00\\
 &  & 1 & 0.24 [0.20, 0.30] & 0.14 & 0.411 / 0.575 & 3.233 / 8.688 & +2.759 / +8.160 & 1.00\\
\midrule
EMNIST / CNN & classifier (68,655) & 0 & 0.77 [0.70, 0.83] & 0.84 & 0.162 / 0.211 & 0.165 / 0.218 & +0.006 / +0.019 & 1.00\\
 &  & 1 & 0.77 [0.70, 0.83] & 0.64 & 0.162 / 0.211 & 0.188 / 0.281 & +0.029 / +0.069 & 1.00\\
 & full (124,975) & 0 & 0.77 [0.70, 0.83] & 0.84 & 0.161 / 0.212 & 0.169 / 0.224 & +0.005 / +0.016 & 1.00\\
 &  & 1 & 0.77 [0.70, 0.83] & 0.64 & 0.161 / 0.212 & 0.190 / 0.275 & +0.030 / +0.069 & 1.00\\
\midrule
AGNews / GRU & full (14,282,672) & 0 & 1.10 [1.01, 1.17] & 0.74 & 0.339 / 0.395 & 0.403 / 0.464 & +0.061 / +0.078 & 1.00\\
 &  & 1 & 1.10 [1.01, 1.17] & 0.82 & 0.339 / 0.395 & 0.366 / 0.433 & +0.026 / +0.047 & 1.00\\
\bottomrule
\end{tabular}}
\end{table}

%% file: appendix/local_steps_table.tex
\begin{table}[!htbp]
\centering\scriptsize\setlength{\tabcolsep}{3pt}
\caption{Local-step sensitivity (MNIST/MLP, seed 42, $q=1$, 1000 rounds). Clients take $E$ full-batch local steps and upload $F_i(w_t)^q(w_i-w_t)$ with the start-of-round loss; server references use the same $E$ steps. Accuracy is global / worst-10\% over rounds 990--999; the worst attack is the minimum over the seven attacks. Honest cone residual and the Pearson correlation between $L_i=\|\hat a_i\|_1$ and $F_i^q$ are pooled over all rounds of the clean run; recall and honest FPR are pooled over the seven attacked runs.}
\label{tab:local-steps}
\resizebox{\textwidth}{!}{\begin{tabular}{lcccccccccccc}
\toprule
$E$ & Clean & Sign & IPM$_{\lambda=3}$ & Gauss. & ALIE & Min-Max & Min-Sum & Magnitude & Worst & honest $D^{\rm cone}$ med/p90 & $r(L_i,F_i^q)$ & recall / FPR\\
\midrule
1 & 93.95/89.26 & 93.20/88.09 & 92.35/87.01 & 93.95/89.26 & 93.94/89.14 & 93.94/89.23 & 93.96/89.26 & 93.96/89.27 & 92.35/87.01 & 0.139/0.238 & 0.944 & 0.88 / 0.05\\
2 & 94.98/90.37 & 92.85/87.57 & 91.33/85.96 & 94.98/90.37 & 94.95/90.22 & 95.01/90.41 & 95.01/90.45 & 94.98/90.37 & 91.33/85.96 & 0.246/0.391 & 0.981 & 0.81 / 0.08\\
5 & 96.06/91.42 & 94.07/89.02 & 91.45/85.78 & 96.06/91.42 & 96.03/91.49 & 96.04/91.47 & 96.05/91.52 & 96.06/91.42 & 91.45/85.78 & 0.358/0.530 & 0.976 & 0.79 / 0.08\\
\bottomrule
\end{tabular}}
\end{table}

%% file: appendix/high_loss_multiseed_per_attack.tex
\begin{table}[!htbp]
\centering\scriptsize
\setlength{\tabcolsep}{3pt}
\caption{\textbf{Per-attack high-loss retention, five seeds.} Retained fair mass (\%) of the highest-loss 20\% of honest clients, mean$_{\pm\text{std}}$ over seeds 40--44, pooled over rounds $r$ with $(r+1)\bmod 10=0$. The last column is the retained fair mass of all honest clients (pooled over attacks).}
\label{tab:high-loss-per-attack-5seed}
\resizebox{\textwidth}{!}{\begin{tabular}{lccccccccc}
\toprule
Dataset/model & Sign & IPM$_{\lambda=3}$ & Gauss. & ALIE & Min-Max & Min-Sum & Magnitude & Pooled & All honest\\
\midrule
CIFAR-10/ResNet-18 & $89.09_{\scriptscriptstyle\pm7.27}$ & $99.17_{\scriptscriptstyle\pm0.78}$ & $100.00_{\scriptscriptstyle\pm0.00}$ & $81.43_{\scriptscriptstyle\pm5.53}$ & $98.22_{\scriptscriptstyle\pm1.91}$ & $99.86_{\scriptscriptstyle\pm0.13}$ & $100.00_{\scriptscriptstyle\pm0.00}$ & $95.23_{\scriptscriptstyle\pm2.32}$ & $97.63_{\scriptscriptstyle\pm0.58}$\\
MNIST/MLP & $82.64_{\scriptscriptstyle\pm5.71}$ & $89.40_{\scriptscriptstyle\pm2.91}$ & $100.00_{\scriptscriptstyle\pm0.00}$ & $92.17_{\scriptscriptstyle\pm5.93}$ & $98.46_{\scriptscriptstyle\pm3.24}$ & $98.45_{\scriptscriptstyle\pm3.47}$ & $100.00_{\scriptscriptstyle\pm0.00}$ & $94.27_{\scriptscriptstyle\pm2.87}$ & $96.01_{\scriptscriptstyle\pm1.07}$\\
EMNIST-Balanced/CNN & $99.18_{\scriptscriptstyle\pm0.47}$ & $99.26_{\scriptscriptstyle\pm0.51}$ & $100.00_{\scriptscriptstyle\pm0.00}$ & $99.13_{\scriptscriptstyle\pm0.38}$ & $100.00_{\scriptscriptstyle\pm0.00}$ & $100.00_{\scriptscriptstyle\pm0.00}$ & $100.00_{\scriptscriptstyle\pm0.00}$ & $99.65_{\scriptscriptstyle\pm0.17}$ & $99.21_{\scriptscriptstyle\pm0.34}$\\
AGNews/GRU & $90.28_{\scriptscriptstyle\pm10.73}$ & $97.79_{\scriptscriptstyle\pm2.34}$ & $100.00_{\scriptscriptstyle\pm0.00}$ & $57.33_{\scriptscriptstyle\pm6.03}$ & $98.49_{\scriptscriptstyle\pm2.00}$ & $98.58_{\scriptscriptstyle\pm1.82}$ & $100.00_{\scriptscriptstyle\pm0.00}$ & $91.40_{\scriptscriptstyle\pm1.92}$ & $94.54_{\scriptscriptstyle\pm0.66}$\\
\bottomrule
\end{tabular}}
\end{table}

%% file: appendix/budget_mismatch_table.tex
\begin{table}[!htbp]
\centering\scriptsize\setlength{\tabcolsep}{3pt}
\caption{Declared budget $f=4$ with fewer actual Byzantine submissions (MNIST/MLP, seed 42, $q=1$, 1000 rounds). Global / worst-10\% accuracy over rounds 990--999; below each block, Byzantine recall, honest false-rejection rate, and the retained fair mass of the highest-loss 20\% of honest clients, pooled over all rounds. Actual $=0$ is the over-declared clean run; actual $=4$ is the matched main-table run.}
\label{tab:budget-mismatch}
\resizebox{\textwidth}{!}{\begin{tabular}{lccccccc}
\toprule
Actual Byzantine & Sign & IPM$_{\lambda=3}$ & Gauss. & ALIE & Min-Max & Min-Sum & Magnitude\\
\midrule
1 of 14: accuracy & 93.76/88.86 & 93.61/88.76 & 93.95/89.32 & 93.92/89.32 & 93.95/89.30 & 93.95/89.29 & 93.93/89.32\\
\quad recall / honest FPR & 0.77 / 0.32 & 0.78 / 0.32 & 1.00 / 0.30 & 0.49 / 0.35 & 1.00 / 0.30 & 1.00 / 0.30 & 1.00 / 0.30\\
\quad retained top-20\% fair mass & 0.71 & 0.73 & 0.75 & 0.69 & 0.75 & 0.75 & 0.75\\
\midrule
2 of 14: accuracy & 93.58/88.67 & 93.28/88.37 & 93.97/89.21 & 93.92/89.28 & 93.97/89.26 & 93.97/89.22 & 93.97/89.21\\
\quad recall / honest FPR & 0.74 / 0.25 & 0.75 / 0.25 & 1.00 / 0.20 & 0.45 / 0.31 & 1.00 / 0.20 & 1.00 / 0.20 & 1.00 / 0.20\\
\quad retained top-20\% fair mass & 0.80 & 0.79 & 0.85 & 0.73 & 0.84 & 0.85 & 0.85\\
\midrule
3 of 14: accuracy & 93.40/88.40 & 92.86/87.68 & 93.98/89.34 & 93.90/89.01 & 93.91/89.12 & 93.97/89.33 & 93.98/89.34\\
\quad recall / honest FPR & 0.68 / 0.20 & 0.68 / 0.20 & 1.00 / 0.10 & 0.99 / 0.10 & 1.00 / 0.10 & 1.00 / 0.10 & 1.00 / 0.10\\
\quad retained top-20\% fair mass & 0.84 & 0.88 & 0.96 & 0.90 & 0.95 & 0.96 & 0.96\\
\midrule
4 of 14: accuracy & 93.20/88.09 & 92.35/87.01 & 93.95/89.26 & 93.94/89.14 & 93.94/89.23 & 93.96/89.26 & 93.96/89.27\\
\quad recall / honest FPR & 0.58 / 0.17 & 0.62 / 0.15 & 1.00 / 0.00 & 0.95 / 0.02 & 1.00 / 0.00 & 1.00 / 0.00 & 0.99 / 0.00\\
\quad retained top-20\% fair mass & 0.84 & 0.90 & 1.00 & 0.93 & 0.99 & 1.00 & 1.00\\
\midrule
0 of 10 (clean, $f=4$): accuracy 93.89/89.18, honest FPR 0.40, retained top-20\% fair mass 0.61\\
\bottomrule
\end{tabular}}
\end{table}

%% file: appendix/theory.tex

\section{Additional Theoretical Analysis and Proofs}
\label{appendix:scboba_theory}

This appendix provides the full statements and proofs for the theoretical
results in Sections~\ref{sec:geometry} and~\ref{sec:theory}, under the
modeling assumptions introduced in Section~\ref{sec:setting}. We restate only
the notation and additional quantitative bounds needed by the proofs.

\subsection{Proof notation}
\label{appendix:scboba_theory:assumptions}

We fix one communication round and suppress the round index. For compactness,
$u_i\in\mathbb R^{d_s}$ and $\Delta_i\in\mathbb R^{d_s}$ in this appendix
denote the screening-coordinate representations (that is, $u_i$ denotes
$v_i$ from the main text). The
server constructs a trusted class basis
$
G=[g_1,\ldots,g_C]\in\mathbb R^{d_s\times C}.
$
For any nonzero vector $u$, define its NNLS coefficient, cone projection, and
normalized cone distance by
$
\widehat a(u)
\in
\arg\min_{a\succeq0}\|u-Ga\|_2^2,
\qquad
z(u):=G\widehat a(u),
\qquad
D_G(u):=\frac{\|u-z(u)\|_2^2}{\|u\|_2^2}.
$
For client $i$, we write $\widehat a_i=\widehat a(v_i)$,
$z_i=z(v_i)$, and $L_i=\|\widehat a_i\|_1$.
All class losses and class-conditioned updates are evaluated at the same
broadcast model, so both are expectations under the same mixture $p_i$.
The trusted loss vector $\widehat\ell$ is used only after screening. For
notational economy, full-model submissions are distinguished only when needed.

\begin{assumption}[Bounded estimation error and nondegenerate trusted geometry]
\label{assump:estimation_geometry}
For every honest client $i$,
\[
|F_i-\overline F_i|\le\epsilon_F,\qquad
\|\Delta_i-\overline\Delta_i\|_2\le\epsilon_x.
\]
The trusted estimates satisfy
\[
\|G-G^\star\|_2\le\epsilon_G,\qquad
\|\widehat\ell-\ell\|_\infty\le\epsilon_\ell,\qquad
\sigma_{\min}(G)\ge\sigma>0.
\]
\end{assumption}

\subsection{Fairness-warped geometry and stable recovery}
\label{appendix:scboba_theory:geometry_recovery}

\begin{theorem}[Full finite-sample coefficient-recovery guarantee]
\label{thm:fair_warped_geometry_full}
Under Assumption~\ref{assump:estimation_geometry}, the finite-sample model in
Eq.~\eqref{eq:finite_cone_model} satisfies
$\|e_i\|_2\le F_i^q(\epsilon_x+\epsilon_G)$, and NNLS recovery obeys
\[
\|\widehat a_i-a_i\|_2
\le\frac{2\|e_i\|_2}{\sigma_{\min}(G)},\qquad
\big|\,\|\widehat a_i\|_1-F_i^q\,\big|
\le\frac{2\sqrt C\,\|e_i\|_2}{\sigma_{\min}(G)}.
\]
\end{theorem}

\begin{proof}[Proof of Theorem~\ref{thm:fair_warped_geometry_full}]
We prove the geometric and recovery claims together.

\paragraph{Population geometry.}
By Assumption~\ref{assump:label_skew} and linearity of expectation, an honest
client with class proportions $p_i$ has population loss and population
one-step update
$
\overline F_i=\ell^\top p_i,
\qquad
\overline\Delta_i=G^\star p_i.
$
Therefore its population fairness-scaled update is
\begin{equation}
\overline u_i
=(\ell^\top p_i)^qG^\star p_i
=G^\star a_i^\star,
\qquad
a_i^\star:=(\ell^\top p_i)^q p_i\succeq0.
\label{eq:app_population_warp}
\end{equation}
Because $\|p_i\|_1=1$,
$
\|a_i^\star\|_1=(\ell^\top p_i)^q.
$
Thus the honest population family is a radially warped image of the
class-mixture simplex: fairness changes the radial coefficient scale while
preserving the nonnegative class-mixture structure.

For completeness, the same identities imply
$
\mathbf 1^\top a_i^\star=(\ell^\top p_i)^q,
\qquad
\ell^\top a_i^\star=(\ell^\top p_i)^{q+1},
$
and hence
$
(\mathbf 1^\top a_i^\star)^{q+1}
=(\ell^\top a_i^\star)^q.
$
At $q=0$, this reduces to the usual simplex constraint
$\mathbf 1^\top a_i^\star=1$.

Every population point in Eq.~\eqref{eq:app_population_warp} belongs to
$\operatorname{cone}(G^\star)$. Conversely, for any $a\succeq0$, $a\neq0$,
let $s=\mathbf 1^\top a$ and $p=a/s$. Since $\ell^\top p>0$,
$
G^\star a
=
\frac{s}{(\ell^\top p)^q}
G^\star[(\ell^\top p)^q p].
$
Hence the conic hull of the fairness-warped honest population family is
exactly $\{G^\star a:a\succeq0\}$. This equality concerns the conic hull,
not the family itself, which generally occupies only a subset of the cone
and retains the fairness-dependent radial constraint above.

\paragraph{Finite-sample cone model.}
For an honest client, set
$
s_i:=F_i^q,
\qquad
a_i:=s_i p_i\succeq0.
$
Using $u_i=s_i\Delta_i$, add and subtract $s_iG^\star p_i$ to obtain
\begin{equation}
u_i=Ga_i+e_i,
\qquad
e_i
:=
s_i(\Delta_i-G^\star p_i)
+s_i(G^\star-G)p_i.
\label{eq:app_perturbed_cone}
\end{equation}
Since $\|p_i\|_2\le\|p_i\|_1=1$, Assumption~\ref{assump:estimation_geometry}
gives
$
\|e_i\|_2
\le
s_i(\epsilon_x+\epsilon_G).
$

\paragraph{NNLS coefficient and scale recovery.}
The true coefficient $a_i$ is feasible for the NNLS problem. Therefore, by
optimality of $\widehat a_i$,
$
\|u_i-G\widehat a_i\|_2
\le
\|u_i-Ga_i\|_2
=
\|e_i\|_2.
$
Together with $u_i=Ga_i+e_i$,
$
\|G(\widehat a_i-a_i)\|_2
\le
\|G\widehat a_i-u_i\|_2+\|e_i\|_2
\le
2\|e_i\|_2.
$
The smallest-singular-value bound then yields
$
\|\widehat a_i-a_i\|_2
\le
\frac{2\|e_i\|_2}{\sigma_{\min}(G)}
\le
\frac{2s_i(\epsilon_x+\epsilon_G)}{\sigma}.
$
Define
$
\delta_i
:=
\frac{2\sqrt C\,s_i(\epsilon_x+\epsilon_G)}{\sigma}.
$
Since $\|v\|_1\le\sqrt C\|v\|_2$ and $\|a_i\|_1=s_i$,
$
|L_i-s_i|
\le
\|\widehat a_i-a_i\|_1
\le
\delta_i.
$
This is the scale-recovery claim used by Component I.

\paragraph{Composition and loss recovery.}
When $\delta_i<s_i$, $L_i\ge s_i-\delta_i>0$ and
$\widehat p_i:=\widehat a_i/L_i$ satisfies
$
\begin{aligned}
\|\widehat p_i-p_i\|_1
&\le
\frac{\|\widehat a_i-a_i\|_1}{L_i}
+
\frac{|L_i-s_i|}{L_i}
\\
&\le
\frac{2\delta_i}{s_i-\delta_i}.
\end{aligned}
$
Consequently, with $\widehat F_i:=\widehat\ell^\top\widehat p_i$,
$
\begin{aligned}
|\widehat F_i-F_i|
&\le
|\widehat\ell^\top(\widehat p_i-p_i)|
+
|(\widehat\ell-\ell)^\top p_i|
+
|\ell^\top p_i-F_i|
\\
&\le
(\ell_{\max}+\epsilon_\ell)
\frac{2\delta_i}{s_i-\delta_i}
+
\epsilon_\ell+
\epsilon_F.
\end{aligned}
$

Thus the same coefficient recovery that supplies the Stage-I scale also yields
a stable class-composition estimate, and therefore a stable trusted-loss
reconstruction whenever the coefficient error is small relative to the honest
fairness scale. This completes the proof.
\end{proof}

\subsection{Two-stage screening guarantee}
\label{appendix:scboba_theory:two_stage}

\begin{theorem}[Full two-stage oracle-recovery guarantee]
\label{thm:scale_cone_separation_full}
Assume that $G$ has full column rank, honest clients form a strict majority,
and $f_t=|\cB_t|$. If $\log\kappa>2r_H$, then no honest submission enters the
Stage-I candidate set. Moreover, any Byzantine submission $b$ satisfying
\[
\log L_b-m_H>\log\kappa+r_H
\]
is rejected by Stage~I; in particular, for a positively scaled copy
$v_b=\lambda v_h$ this condition is guaranteed whenever
\[
\log\lambda>\log\kappa+2r_H,
\qquad
r_H\le q\log\frac{\ell_{\max}}{\ell_{\min}}
+2\epsilon_{\log L}.
\]
For a remaining Byzantine submission $v_b=Gc_b+r_b$ with
$r_b\perp\operatorname{span}(G)$,
\[
D_G(v_b)
\ge
\frac{\|r_b\|_2^2+\sigma_{\min}^2(G)\|(c_b)_-\|_2^2}
{\|v_b\|_2^2}.
\]
If this lower bound exceeds the largest honest cone score, the submission
ranks above every honest client in Stage~II. Consequently, if every Byzantine
submission satisfies either the Stage-I or the Stage-II separation condition,
then
\[
\widehat{\cH}_t=\cH_t.
\]
\end{theorem}

\begin{proof}[Proof of Theorem~\ref{thm:scale_cone_separation_full}]
We analyze the radial and structural stages in the same proof because they are
designed to detect complementary deviations.

\paragraph{Stage I: positive homogeneity and the contaminated median.}
For any $\lambda>0$, the change of variables $a=\lambda b$ gives
\begin{equation}
\widehat a(\lambda u)=\lambda\widehat a(u),
\qquad
L(\lambda u)=\lambda L(u).
\label{eq:app_nnls_homogeneity}
\end{equation}
The cone projection scales in the same way, $z(\lambda u)=\lambda z(u)$, so
$
D_G(\lambda u)=D_G(u).
$
Thus a positive radial copy changes the coefficient scale but not the normalized
cone score.

Let $x_h=\log L_h$ for honest clients, let
$
m_H=\operatorname{median}_{h\in\mathcal H}x_h,
\qquad
r_H=\max_{h\in\mathcal H}|x_h-m_H|,
$
and let $m_S=\log L_{\rm med}$ be a median over all received submissions.
Every honest score lies in $[m_H-r_H,m_H+r_H]$. Because honest submissions form
a strict majority, any contaminated median must also lie in this interval.
Therefore, for every honest client,
$
x_h-m_S\le2r_H.
$
If $\log\kappa>2r_H$, no honest client can satisfy
$L_h>\kappa L_{\rm med}$.

Conversely, for any Byzantine submission $b$ satisfying
$
x_b-m_H>\log\kappa+r_H,
$
we have, since $m_S\le m_H+r_H$,
$
x_b-m_S\ge x_b-m_H-r_H>\log\kappa.
$
Hence $b$ enters the Stage-I candidate set. Since no honest submission enters
that set and $f_t=|\cB_t|$, every such candidate is rejected.

Now consider a positive radial copy $u_b=\lambda u_h$. By
Eq.~\eqref{eq:app_nnls_homogeneity},
$x_b=x_h+\log\lambda$. Since $x_h\ge m_H-r_H$ and
$m_S\le m_H+r_H$,
$
x_b-m_S
\ge
\log\lambda-2r_H.
$
Hence
$
\log\lambda>\log\kappa+2r_H
$
places $b$ in the Stage-I candidate set. The same condition implies
$x_b>m_H+r_H\ge\max_{j\in\mathcal H}x_j$, so such an attacker has larger
recovered scale than every honest client and cannot be displaced by an honest
candidate in the top-$f$ truncation.

The fairness strength $q$ enters this condition through the honest radial
spread. In the exact population model,
$
L_h=(\ell^\top p_h)^q,
$
so
$
q\log\ell_{\min}
\le
\log L_h
\le
q\log\ell_{\max},
\qquad
r_H
\le
q\log\frac{\ell_{\max}}{\ell_{\min}}.
$
If every recovered log scale differs from its population value by at most
$\epsilon_{\log L}$, the same argument gives
$
r_H
\le
q\log\frac{\ell_{\max}}{\ell_{\min}}
+2\epsilon_{\log L}.
$
Thus stronger fairness weighting legitimately widens the honest radial range
and correspondingly increases the separation margin required by a purely
radial detector.

\paragraph{Stage II: distance to the trusted nonnegative cone.}
For any nonzero received vector $u$, write its orthogonal decomposition relative
to $\operatorname{span}(G)$ as
$
u=Gc+r,
\qquad
c=G^\dagger u,
\qquad
r\perp\operatorname{span}(G).
$
Here $\operatorname{span}(G)$ contains all linear combinations of the
columns of $G$, including signed combinations, and
$G^\dagger=(G^\top G)^{-1}G^\top$ is the Moore--Penrose pseudoinverse of the
full-column-rank basis.
For every $a\succeq0$, orthogonality gives
\begin{equation}
\|u-Ga\|_2^2
=
\|r\|_2^2+
\|G(c-a)\|_2^2.
\label{eq:app_cone_pythagoras}
\end{equation}
Let $c_-:=\max\{-c,0\}$ coordinate-wise. The Euclidean distance from $c$ to
the nonnegative orthant is $\|c_-\|_2$. Applying the singular-value bounds to
Eq.~\eqref{eq:app_cone_pythagoras} yields
\begin{equation}
D_G(u)
\ge
\frac{
\|r\|_2^2+
\sigma_{\min}^2(G)\|c_-\|_2^2
}{\|u\|_2^2}.
\label{eq:app_cone_structural_lower}
\end{equation}
The first term measures energy outside the trusted class span; the second
measures a within-span violation of nonnegative class-mixture structure.

The NNLS projection also gives the geometric condition used in the method
interpretation. Let $\rho=u-z(u)$. The Karush--Kuhn--Tucker (KKT)
optimality conditions are
$
G^\top\rho\preceq0,
\qquad
\widehat a_c(u)\,g_c^\top\rho=0
\quad(c=1,\ldots,C).
$
Hence $\rho$ is orthogonal to every active generator
$g_c$ with $\widehat a_c(u)>0$, and therefore to the active face containing the
projection $z(u)$. It need not be orthogonal to the entire span of $G$.
In particular, summing the complementary products gives
$
\langle u-z(u),z(u)\rangle
=\widehat a(u)^\top G^\top\rho=0.
$
Thus the residual is orthogonal to the fitted update, not to every vector
in the cone.

For an honest client, Eq.~\eqref{eq:app_perturbed_cone} provides a feasible
nonnegative coefficient $a_i$, and therefore
$
D_G(u_i)
\le
\frac{\|e_i\|_2^2}{\|u_i\|_2^2}.
$
Define
$
\varepsilon_C
:=
\max_{h\in\mathcal H}D_G(u_h).
$
For any remaining Byzantine submission $u_b=Gc_b+r_b$, if
\begin{equation}
\frac{
\|r_b\|_2^2+
\sigma_{\min}^2(G)\|(c_b)_-\|_2^2
}{\|u_b\|_2^2}
>
\varepsilon_C,
\label{eq:app_structural_separation}
\end{equation}
then Eq.~\eqref{eq:app_cone_structural_lower} implies
$D_G(u_b)>D_G(u_h)$ for every honest client $h$. Thus every Byzantine vector
satisfying the stated structural margin ranks ahead of all honest clients in
Stage II. If Stage I has used rejection slots only on Byzantine clients and all
remaining Byzantine clients satisfy Eq.~\eqref{eq:app_structural_separation},
the remaining top-score budget removes them before any honest client. This
completes the proof.
\end{proof}

\paragraph{Post-screening step-size control.}
The final aggregation does not alter the direction of the retained mean
submission. Writing
$
\Delta_t^{\rm SC}=\alpha_t\bar u_t,
\qquad
\alpha_t=
\frac{|\widehat{\mathcal H}_t|}
{\sum_{i\in\widehat{\mathcal H}_t}d_i},
$
and assuming $\widetilde F_i\le\widetilde F_{\max}$ gives
$
\frac{1}{\max\{1,\widetilde F_{\max}^q\}}
\le
\alpha_t
\le
1.
$
Therefore the post-screening rule prevents additional amplification while
retaining a positive multiplicative lower bound on the step size.

\[
\bar u_t=\frac{1}{|\widehat\cH_t|}\sum_{i\in\widehat\cH_t}u_i,\qquad
\alpha_t=\frac{|\widehat\cH_t|}
{\sum_{i\in\widehat\cH_t}\max\{1,\widetilde F_i^q\}},\qquad
w_{t+1}=w_t+\alpha_t\bar u_t.
\label{eq:aggregation_scale}
\]

\subsection{Conditional one-step descent}
\label{appendix:scboba_theory:conditional_descent}

\begin{proposition}[Conditional one-step descent after screening]
\label{prop:conditional_descent}
Let $\bar u_t$ and $\alpha_t>0$ be as defined above.
Suppose $\mathcal J_q$ is $L_J$-smooth along the update segment, with $L_J>0$,
and the retained mean satisfies, for some $\gamma_t>0$,
$
-\langle\nabla\mathcal J_q(w_t),\bar u_t\rangle
\ge\gamma_t\|\nabla\mathcal J_q(w_t)\|_2\|\bar u_t\|_2.
$
Then $w_{t+1}=w_t+\alpha_t\bar u_t$ satisfies
$
\mathcal J_q(w_{t+1})
\le\mathcal J_q(w_t)
-\alpha_t\gamma_t\|\nabla\mathcal J_q(w_t)\|_2\|\bar u_t\|_2
+\frac{L_J}{2}\alpha_t^2\|\bar u_t\|_2^2.
$
Consequently, $\mathcal J_q(w_{t+1})<\mathcal J_q(w_t)$ if $\bar u_t\ne0$
and
$
\alpha_t\|\bar u_t\|_2
<\frac{2\gamma_t}{L_J}\|\nabla\mathcal J_q(w_t)\|_2.
$
\end{proposition}

\begin{proof}
Here $\bar u_t$ is the original full-model retained mean, not the selected representation used in the
screening proofs. By $L_J$-smoothness along the update segment,
$
\mathcal J_q(w_t+\alpha_t\bar u_t)
\le\mathcal J_q(w_t)
+\alpha_t\langle\nabla\mathcal J_q(w_t),\bar u_t\rangle
+\frac{L_J}{2}\alpha_t^2\|\bar u_t\|_2^2.
$
Since $\alpha_t>0$, substituting the assumed alignment inequality gives the
claimed bound. Its change from $\mathcal J_q(w_t)$ is at most
$
-\alpha_t\|\bar u_t\|_2
\left(\gamma_t\|\nabla\mathcal J_q(w_t)\|_2
-\frac{L_J}{2}\alpha_t\|\bar u_t\|_2\right).
$
For $\bar u_t\ne0$, the prefactor $\alpha_t\|\bar u_t\|_2$ is positive;
the stated step-size condition makes the parenthesis positive, yielding
strict descent. If $\bar u_t=0$, the model and objective remain unchanged.
\end{proof}

\subsection{Attack implications and observable boundary}

\paragraph{Representative attack implications.}
The two parts of Theorem~\ref{thm:scale_cone_separation} immediately explain the
representative attack geometries used in the experiments. A positive magnitude
copy $u_b=\lambda u_h$ satisfies
$D_G(u_b)=D_G(u_h)$ but shifts the recovered log scale by exactly
$\log\lambda$, so it is a Stage-I rather than Stage-II deviation. For an ideal
sign reversal $u_b=-\lambda Ga$ with $a\succeq0$, the vector remains inside
$\operatorname{span}(G)$ but has negative cone coordinates; consequently
$
D_G(u_b)
\ge
\frac{\sigma_{\min}^2(G)\|a\|_2^2}{\|Ga\|_2^2}.
$
An unconstrained subspace distance would be zero for the same vector, which is
why the nonnegativity constraint matters. Finally, for an off-span perturbation
$u=Ga+r$ with $a\succeq0$ and $r\perp\operatorname{span}(G)$,
$
D_G(u)=\frac{\|r\|_2^2}{\|u\|_2^2}.
$
These are direct specializations of the general two-stage theorem rather than
separate theoretical claims.

\paragraph{Observable boundary.}
The nonnegative cone is a relaxation of the fairness-warped honest family;
cone consistency tests structural compatibility but does not certify honesty.
An exact cone-preserving submission with a plausible coefficient scale satisfies
$
u_b=Ga_b,\quad a_b\succeq0,\quad
\|a_b\|_1\le\kappa L_{\rm med}
\quad\Longrightarrow\quad
D_G(u_b)=0,\quad b\notin\mathcal R_{\rm scale}.
$
Likewise, parameter-selection equivalence implies
$
v_b=v_h
\quad\Longrightarrow\quad
(\widehat a_b,L_b,D_G(u_b),\widetilde F_b)
=
(\widehat a_h,L_h,D_G(u_h),\widetilde F_h).
$
Thus \methodname{} provides no score-based certificate unless an attack creates a
separated scale or cone deviation in the selected screening representation.